%% file: paper.tex
\def\year{2021}\relax
\documentclass[a4]{article}
\usepackage[a4paper,left=0.5in,right=0.5in,top=0.5in,bottom=0.5in,%
footskip=.25in]{geometry}
\usepackage{titling}
\usepackage{helvet}

\usepackage[hyphens]{url}
\usepackage{graphicx}
\usepackage{caption}
\usepackage[superscript]{cite}
\usepackage{enumitem}%
\usepackage[justification=centering]{caption}

\usepackage{stfloats}
\usepackage[hang,flushmargin, norule]{footmisc}
\usepackage{marvosym}

\usepackage[table,x11names,dvipsnames]{xcolor}

\usepackage{titlesec}
\usepackage{multicol}
\usepackage{cite}
\usepackage{amsmath,amssymb,amsfonts}
\usepackage{algcompatible}
\usepackage{algorithm,algpseudocode}
\usepackage{graphicx}
\usepackage{booktabs, multirow, soul}
\usepackage[flushleft]{threeparttable}
\usepackage{xspace, mathtools, comment}
\usepackage{caption}
\usepackage{subcaption}
\usepackage[resetlabels,labeled]{multibib}
\usepackage{etoolbox}

\usepackage{multicol}
\usepackage{lipsum}

\usepackage{xr}
\titlespacing*{\section}{0pt}{1\baselineskip}{0.1\baselineskip}
\titlespacing*{\subsection}{0pt}{1\baselineskip}{0.1\baselineskip}
\titlespacing*{\subsubsection}{0pt}{1\baselineskip}{0.1\baselineskip}

\titleformat*{\section}{\normalfont\sffamily\large\bfseries}
\titleformat*{\subsection}{\normalfont\sffamily\normalsize\bfseries}
\titleformat*{\subsubsection}{\normalfont\sffamily\small\bfseries}

\algnewcommand{\LineComment}[1]{\Statex \(\triangleright\) #1}
\algnewcommand{\ShortLineComment}[1]{\Statex \hspace{1.8em}\(\triangleright\) #1}
\algnewcommand{\ShortShortLineComment}[1]{\Statex \hspace{3.1em}\(\triangleright\) #1}

\DeclareCaptionLabelSeparator{bar}{\space\textbar\space}

\makeatletter
\renewcommand\@biblabel[1]{#1.}
\makeatother

\makeatletter
\renewenvironment{table*}%
{\renewcommand\familydefault\sfdefault
	\@float{table}}
{\end@float}
\makeatother

\makeatletter
\renewcommand{\scriptsize}{\@setfontsize\scriptsize{7}{7}}
\renewcommand{\footnotesize}{\@setfontsize\footnotesize{8}{9}}
\makeatother

\patchcmd{\thebibliography}
{\settowidth}
{\setlength{\parsep}{0pt}\setlength{\itemsep}{0pt plus 0pt}\settowidth}
{}{}
\apptocmd{\thebibliography}
{\small}
{}{}

\renewcommand{\footnoterule}{%
	\kern -3pt
	\hrule width 0.5\textwidth height 0.3pt
	\kern 2pt
}

\makeatletter
\newcommand\footnoteref[1]{\protected@xdef\@thefnmark{\ref{#1}}\@footnotemark}
\makeatother

\title{Generating 3D Binding Molecules Using Shape-Conditioned Diffusion Models with Guidance}
\date{\vspace{-5ex}}

\author{
	Ziqi Chen\textsuperscript{\rm 1}, 
	Bo Peng\textsuperscript{\rm 1}, 
	Tianhua Zhai\textsuperscript{\rm 2},
	Daniel Adu-Ampratwum\textsuperscript{\rm 3},
	Xia Ning\textsuperscript{\rm 1,3,4,5 \Letter}
}
\newcommand{\Address}{
	\textsuperscript{\rm 1}Computer Science and Engineering, The Ohio State University, Columbus, OH 43210.
	\textsuperscript{\rm 2}Perelman School of Medicine, University of Pennsylvania, Philadelphia, PA 19104.
	\textsuperscript{\rm 3}Medicinal Chemistry and Pharmacognosy, The Ohio State University, Columbus, OH 43210.
	\textsuperscript{\rm 4}Biomedical Informatics, The Ohio State University, Columbus, OH 43210.
	\textsuperscript{\rm 5}Translational Data Analytics Institute, The Ohio State University, Columbus, OH 43210.
	\textsuperscript{\Letter}ning.104@osu.edu
}
\newcommand{\myabstract}[2][1]{%
	\renewcommand\maketitlehookd{
		\Address
		\vspace{10pt}
		\mbox{}\medskip\par
		\centering
		\begin{minipage}{#1\textwidth}
			\textbf{
			{\fontfamily{phv}\selectfont
				#2
				}
			}
		\end{minipage}
	}
}

\input{define}

\usepackage{bibentry}

\begin{document}

\myabstract{Drug development is a critical but notoriously resource- and time-consuming process.
In this manuscript, we develop a novel generative artificial intelligence (genAI) method \method to facilitate drug development.
\method generates 3D binding molecules based on the shapes of known ligands.
\method encapsulates geometric details of ligand shapes within pre-trained, expressive shape embeddings and then generates new binding molecules through a diffusion model.  
\method further modifies the generated 3D structures iteratively via shape guidance to better resemble the ligand shapes.
It also tailors the generated molecules toward optimal binding affinities under the guidance of protein pockets.
Here, we show that \method outperforms the state-of-the-art methods on benchmark datasets. 
When generating binding molecules resembling ligand shapes, \method with shape guidance achieves a success rate 61.4\%, substantially outperforming the best baseline (11.2\%), meanwhile producing molecules with novel molecular graph structures.
\method with pocket guidance also outperforms the best baseline in binding affinities by 13.2\%, and even by 17.7\% when combined with shape guidance.
Case studies for two critical drug targets demonstrate very favorable physicochemical and pharmacokinetic properties of 
the generated molecules, thus,  the potential of \method in developing promising drug candidates.}

\maketitle

\vspace{-10pt}
\section*{Introduction}

Drug development is a critical but notoriously resource- and time-consuming process~\cite{Sun2022}.
It typically takes 10-15 years and \$1 to \$1.6 billion to fully develop a successful drug~\cite{Wouters2020}.
To expedite the process and improve cost efficiency,  tremendous research efforts have been dedicated to developing computational methods to facilitate drug development~\cite{Yu2016}.
Existing computational methods to design potential drug candidates
could be categorized into ligand-based drug design (LBDD)~\cite{Acharya2011} and structure-based drug design (SBDD)~\cite{Anderson2003}, which search over molecule libraries to 
identify those resembling known ligands or binding to known binding sites of protein targets, respectively. 
Though promising, the opportunistic trial-and-error paradigm underpinning LBDD and SBDD 
is often confined by the limited scale of molecule libraries and cannot ensure optimal precision design~\cite{Gimeno2019}. 
Thus, the outcomes are highly subjective to the knowledge and experience of the domain experts conducting the experiments, which also limits the scalability and automation of rapid drug design 
for new protein targets. 
Recently, generative artificial intelligence (genAI) methods, such as variational autoencoders~\cite{kingma2013auto},  diffusion~\cite{song2021denoising}, and ChatGPT~\cite{OpenAI2023}, 
have emerged as groundbreaking computational tools for 
many applications~\cite{Yu2024}, including drug design~\cite{jin18jtvae,schneuing2022structure,liu2024conversational}.
Instead of searching for drug candidates, genAI methods could directly generate molecules 
satisfying prescribed properties (e.g., lipophilicity, druglikeness), 
through learning the underlying chemical knowledge carried by vast molecule datasets, 
and making autonomous decisions in constructing new molecules~\cite{jin18jtvae,schneuing2022structure} (e.g., molecular graphs, 3D structures). 
The powerful generative capabilities of genAI demonstrate significant promise in fundamentally transforming the traditional drug development process into a more focused, accurate, swift, and sustainable alternative.

In this manuscript, we introduce \method, a novel genAI method to generate molecules in 3D that 
effectively bind to 
a protein target and have realistic structures (e.g., correct bond angles and bond lengths).
Motivated by LBDD, \method generates novel binding molecules based on the shapes of known ligands, following the principle that molecules with similar shapes tend to have similar binding activities~\cite{Bostroem2006, Acharya2011}.
\method encapsulates the geometric details of ligand shapes within pre-trained, expressive shape embeddings, and 
generates new binding molecules, including their atom types and atom positions 
through diffusion~\cite{song2021denoising}. 
During the iterative diffusion process, \method leverages a novel molecule graph representation learning approach and integrates ligand shape embeddings in generating and refining the atom types and atom positions, and thus, a new molecule and its 3D 
structures. 
To better resemble the known ligand shapes, \method further modifies the generated 3D structures iteratively under the guidance 
of the ligand shapes. 
Inspired by SBDD, in addition to ligands, \method can also leverage the geometric information of protein binding pockets
and tailor the generated molecules toward optimal binding affinities under the guidance of binding pockets.

Our comprehensive experiments demonstrate that \method achieves superior performance in generating molecules with highly similar shapes to ligands, compared to state-of-the-art shape-conditioned molecule generation (SMG) methods.
Notably, \method achieves a 28.4\% success rate in generating molecules that closely resemble ligand shapes and have novel graph structures, substantially outperforming the 11.2\% success rate of the best SMG method.
Moreover, incorporating shape guidance further boosts the performance of \method to a remarkable 61.4\% success rate, while generating realistic 3D molecules.
This highlights the effectiveness of \method's pre-trained shape embeddings to capture geometric details of ligand shapes and the ability of its customized diffusion model in generating realistic and novel binding molecules.
In addition, by utilizing geometric information from protein binding pockets, \method with pocket guidance outperforms the best pocket-conditioned molecule generation (PMG) method by 13.2\% improvement in binding affinities of generated molecules.
When both pocket and shape guidance are incorporated, the improvement reaches 17.7\%.
Case studies with extensive \emph{in silico} analyses for two important drug targets, cyclin-dependent kinase 6 (CDK6) 
that is highly associated with multiple cancers such as lymphoma and leukemia, 
and neprilysin (NEP) that is highly associated with Alzheimer's disease, 
demonstrate that \method effectively generates drug-like molecules specifically for these targets. 
The two studied generated molecules for CDK6 show binding affinities (Vina scores)~\cite{Eberhardt2021} of -6.817 kcal/mol and -6.970 kcal/mol, better than that of the known CDK6 ligand (0.736 kcal/mol); 
and the studied generated molecule for NEP also achieves a superior Vina score of -11.953 kcal/mol compared to the known 
NEP ligand (-9.399 kcal/mol).
These molecules also have favorable drug-like properties, with high QED values~\cite{Bickerton2012} close to or above 0.8, low toxicity scores ranging from 0.000 to 0.236, and compliance with Lipinski's rule of five~\cite{Lipinski1997}.
Notably, their profiles for absorption, distribution, metabolism, excretion, and toxicity (ADMET) are comparable to those of FDA-approved drugs.
These results further highlight the potential of \method in advancing drug development.
%

\section*{Related Work}

A variety of deep generative models have been developed to generate molecules using various molecule representations, 
including generating SMILES string representations~\cite{GmezBombarelli2018}, or 2D molecular graph representations~\cite{jin18jtvae,Chen2021modof}.
{However, these representations fall short in capturing the 3D structures of molecules, which are critical for understanding their biological activities and certain properties.}
Recent efforts have been dedicated to the generation of 3D molecules. 
For example, Hoogeboom \etal~\cite{hoogeboom22diff} developed an equivariant diffusion model 
in which an equivariant network is employed to
jointly predict both the positions and features of all atoms.
In 3D molecule generation, two types of methods have been developed.
Motivated by LBDD, the first type of methods, referred to as shape-conditioned molecule generation (SMG) methods, generates molecules with similar shapes to condition molecules (e.g., ligands).
The second type of methods, referred to as pocket-conditioned molecule generation (PMG) methods, is motivated by SBDD and generates binding molecules to a target protein pocket.
%

\subsection*{SMG Methods}

%
%
%
Previous SMG methods~\cite{long2022zero, adams2023equivariant} generally leverage shapes as conditions and use generative models such as variational autoencoders (VAE)~\cite{kingma2013auto} to generate potentially binding molecules.  
Among SMG methods, Adams and Coley~\cite{adams2023equivariant} developed \squid, which consists of a fragment-based generative model and a rotatable-bond scoring model.
The former generates molecules using VAE and sequentially decodes fragments based on the shapes of condition molecules (e.g., ligands), while the latter adjusts the angles of rotatable bonds between fragments to adapt to the condition shapes.
%
%
Long \etal~\cite{long2022zero} developed an encoder-decoder framework, referred to as $\mathsf{Shape2Mol}$, which first encodes 3D shapes of molecules into latent embeddings and then generates fragments sequentially based on these embeddings to build molecules.
{In our preliminary work~\cite{Chen2023ShapeMol}, we also demonstrated the potential of diffusion models for generating binding molecules conditioned on shapes. 
By improving the shape-conditioned molecule prediction module (Section ``Shape-conditioned Molecule Prediction''), we have significantly enhanced the performance of \method.}

%
It is worth noting that \method is fundamentally different from \squid.
%
%
\squid, as a fragment-based method, generates molecules by sequentially adding fragments. 
When predicting the next fragments, however, \squid fails to consider the effects of their various poses, and thus, could lead to inaccurate fragment predictions.
Due to the sequential nature, the prediction errors will be cumulated and could substantially degrade the generation performance.
Different from \squid, \method generates molecules by directly arranging atoms in the 3D space using diffusion models.
This design explicitly considers the influence of varying 3D atom positions in the generation process, leading to effective generations. 
In addition, by using only fragments in a predefined library, \squid could struggle to generate diverse molecules, 
while \method ensures superior diversity by allowing for the generation of any fragments.
%
%
%
%
\method also captures the flexibility of bonding geometries
in real 3D molecules by generating molecules with flexible bond lengths and angles.
However, \squid can only generate molecules with fixed bond lengths and angles, leading to the discrepancy in 3D structures between the 
generated molecules and real molecules. 

\method 
is also different from $\mathsf{Shape2Mol}$. 
\method is specifically designed to be equivariant under any rotations and translations of the shape condition, allowing for better sampling efficiency~\cite{Jonas20a}.
Conversely, $\mathsf{Shape2Mol}$ is not equivariant, and thus, suffers from limited training efficiency. 
In addition, different from \method, $\mathsf{Shape2Mol}$ is a fragment-based approach and could suffer from the same issues as discussed above for \squid. 
%

%
%
%
%
%
%
%
%
 
\subsection*{PMG Methods}


For PMG, previous work~\cite{luo2021sbdd,peng22pocket2mol,guan2023targetdiff,guan2023decompdiff} has been focused on directly utilizing protein pockets as a condition and generating molecules binding towards these pockets.
These methods can be grouped into three categories: VAE-based, autoregressive model-based, and diffusion model-based.
Among VAE-based methods, Ragoza \etal~\cite{ragoza2022chemsci} developed a conditional VAE model to generate atomic density grids based on the density grids of protein pockets.
The generated atomic density grids are then converted to molecules.
Several autoregressive models~\cite{luo2021sbdd,peng22pocket2mol,liu2022} also have been developed to generate binding molecules by sequentially adding atoms into the 3D space conditioned on 
protein pocket atoms. 
Particularly, Luo \etal~\cite{luo2021sbdd} developed an autoregressive model \AR to estimate the probability density of atoms' occurrences in the 3D space conditioned on protein pockets.
\AR sequentially adds atoms based on these estimations to construct molecules. 
Peng \etal~\cite{peng22pocket2mol} improved \AR into \pockettwomol by incorporating a more efficient atom sampling strategy.
\pockettwomol determines the positions of newly added atoms by predicting their relative positions to previously added atoms.
Diffusion models are also very popular in PMG.
Guan \etal~\cite{guan2023targetdiff} developed a conditional diffusion model \targetdiff that generates molecules based on protein pockets by sequentially denoising both continuous atom coordinates and categorical atom types in noisy molecules. 
Guan \etal~\cite{guan2023decompdiff} further improved \targetdiff into \decompdiff by utilizing data-dependent prior distributions over molecular arms and scaffolds.
These priors are derived from either known ligands or protein pockets.

%
Though promising, 
PMG methods require protein-ligand complex data for training.
However, such data is expensive and thus highly limited.
%
%
{The sparse ground-truth binding ligands}
%
confine 
these methods in exploring a wide range of molecules with desired properties.  
In contrast, \method can learn from rich molecule data, improving its ability to generate effective and novel binding molecules.

\section*{Materials}
%

{We evaluate the effectiveness of \method in both SMG and PMG.
For SMG, following the literature~\cite{adams2023equivariant}, we evaluate whether \method could generate realistic 3D molecules 
that have shapes similar to condition molecules;
for PMG, following the literature~\cite{luo2021sbdd,peng22pocket2mol,guan2023targetdiff}, we assess, given target protein pockets, whether \method could generate molecules with high binding affinities and realistic structures. }
%
%
Particularly, for SMG, we evaluate \method and its variant with shape guidance (detailed in Section ``\method with Shape Guidance''), referred to as \methodwithsguide, against the state-of-the-art SMG baselines in terms of shape similarity, diversity, and realism of generated molecules.
For PMG, we compare both \method and \methodwithsguide with pocket guidance (detailed in Section ``\method with Pocket Guidance''), referred to as \methodwithpguide and \methodwithsandpguide, to state-of-the-art PMG baselines to investigate if \method can effectively generate realistic molecules binding towards protein targets.
Note that different from PMG baselines trained on sparse protein-ligand complex data, \method is capable of leveraging large-scale molecule data for better generation. 
%
%
%
In the following sections, we will first present the SMG and PMG baselines (Section ``Baselines'').
Subsequently, we will present the data used in our experiments (Section ``Data''), the experimental setups (Section ``Experimental Setup'') and the evaluation metrics (Section ``Evaluation Metrics''). 
Details about hyper-parameters used in \method are available in Supplementary Section~\ref{supp:experiments:parameters}.

%
%
%
%
%
%
%

\subsection*{Baselines}

\paragraph{SMG Baselines}
To evaluate the effectiveness of \method in generating molecules with similar shapes to condition molecules, 
we compare \method and \methodwithsguide with the state-of-the-art SMG baseline \squid and a virtual screening method \dataset.
%
%
%
%
%
%
%
%
%
As introduced in the original paper~\cite{adams2023equivariant}, \squid uses a variable $\lambda$ to balance the interpolation and extrapolation in the latent space.
In our experiments, we include \squid with $\lambda=0.3$ and \squid with $\lambda=1.0$ following the literature~\cite{adams2023equivariant}. 
%
%
\dataset aims to screen through the training set to identify molecules with high shape similarities with the condition
molecule.
%
%
%
%
Note that we do not consider $\mathsf{Shape2Mol}$~\cite{long2022zero} as our baseline for two reasons.
First, the code they provided is closely tied to a private infrastructure~\footnote{https://github.com/longlongman/DESERT/tree/830562e13a0089e9bb3d77956ab70e606316ae78}, making it {highly nontrivial to adapt their code to our infrastructure}. 
%
Moreover, $\mathsf{Shape2Mol}$ requires {prohibitively} 
{intensive} computing resources. 
According to their paper, $\mathsf{Shape2Mol}$ is trained on 32 Tesla V100 GPUs for 2 weeks. 
%
%

%

\paragraph{PMG Baselines} 
To evaluate the effectiveness of \method in generating molecules binding towards target protein pockets,
%
we compare \methodwithpguide and \methodwithsandpguide with four state-of-the-art PMG baselines, including  \AR~\cite{luo2021sbdd}, \pockettwomol~\cite{peng22pocket2mol}, \targetdiff~\cite{guan2023targetdiff}, and \decompdiff~\cite{guan2023decompdiff}. 
%
%
%
For \decompdiff, we exclude \decompdiff with protein pocket priors from the comparison, and only include \decompdiff with known ligand priors.
This is due to the substantially lower performance of \decompdiff with protein pocket priors in generating molecules with desirable drug-likeness compared to other methods.
More details about \decompdiff with protein pocket priors will be discussed in Supplementary Section~\ref{supp:app:decompdiff}.

\subsection*{Data}

\input{tables/dataset}

\paragraph{Data for SMG} 
%
%
%
Following \squid~\cite{adams2023equivariant}, we use molecules in the MOSES dataset~\cite{mose2020}, with their 3D conformers calculated by RDKit~\cite{rdkit}.
We use the same training and testing split as in \squid.
%
%
{Please note that {\squid} further modifies the generated conformers into artificial ones, by adjusting acyclic bond distances to their empirical means 
and fixing acyclic bond angles using heuristic rules.}
%
%
Unlike \squid, we do not make any additional adjustments to the calculated 3D conformers, as \method is designed with sufficient flexibility to accept any 3D conformers as input. 
%
Limited by the predefined fragment library, \squid also removes molecules with fragments not present in its fragment library. 
In contrast, we keep all the molecules, as {\method} is not based on fragments. 
As a result, our training set includes 1,593,653 molecules. 
%
%
%
%
The same set of 1,000 
molecules as in \squid is used {for testing.}
For hyper-parameter tuning, we randomly sample 1,000 molecules from the training set for validation.
Table~\ref{tbl:data} presents the data statistics for SMG.


\paragraph{Data for PMG} 
Following the previous work~\cite{luo2021sbdd, peng22pocket2mol, guan2023targetdiff, guan2023decompdiff}, we use the CrossDocked2020 benchmark dataset~\cite{Francoeur2020} with protein-ligand complex data to evaluate \method. 
%
%
During evaluation, for \method, we directly utilize the model trained on the MOSES without fine-tuning on the complex data. 
For all the PMG baselines, we use the model checkpoints released by the authors.
All PMG baselines are trained on the training set of CrossDocked2020 as presented in their original paper~\cite{luo2021sbdd, peng22pocket2mol, guan2023targetdiff, guan2023decompdiff}, which includes {11,915 unique ligands, 15,207 unique proteins}, and 100,000 protein-ligand complexes in total.
Note that the PMG baselines are designed specifically for protein-ligand complex data, and cannot accept {molecule data as input. }
Thus, we do not tune PMG baselines on the MOSES training set.
We use the same test dataset as in the previous work~\cite{luo2021sbdd, peng22pocket2mol, guan2023targetdiff, guan2023decompdiff}, which includes 100 protein-ligand complexes with novel proteins.
{Note that the MOSES dataset focuses on molecules with number of atoms ranging from 8 to 27.
In this evaluation, we do not consider complexes with out-of-distribution ligands (i.e., ligands with more than 27 atoms).
Thus, we exclude 28 complexes from the test set of CrossDocked2020.
Table~\ref{tbl:data} presents the data statistics for PMG.
\subsection*{Experimental Setup}

\paragraph{Evaluation of \method in SMG} 
%
%
%
%
%
%
%
%
%
%
Following \squid~\cite{adams2023equivariant}, we apply \method, \methodwithsguide and all SMG baselines to generate 50 molecules per test molecule for evaluation.  
%
%
For {\dataset}, following \squid, we randomly sample 500 training molecules for each test molecule.
We then identify and select the top-50 molecules from the 500 molecules that have the highest shape similarities to the test molecule.
%
%

\paragraph{Evaluation of \method in PMG} 
As discussed above, we directly utilize the \method model trained on the MOSES dataset for the evaluation against PMG baseline methods.
%
%
%
Following previous PMG baselines~\cite{luo2021sbdd, peng22pocket2mol, guan2023targetdiff, guan2023decompdiff}, we use \methodwithpguide, and \methodwithsandpguide to generate 100 molecules for each test protein-ligand complex.
{For \methodwithpguide and \methodwithsandpguide}, we use \SE to encode the shapes of ligands in test protein-ligand complexes into shape embeddings.
{Then, we use \methodwithpguide and \methodwithsandpguide to generate molecules conditioned on these embeddings. 
For baselines, 
we directly use molecules generated {from} 
\AR, \pockettwomol and \targetdiff, as provided by \targetdiff~\footnote{https://github.com/guanjq/targetdiff}, to calculate evaluation metrics.}
For \decompdiff, we use the model checkpoints released by the authors to generate 100 molecules for each test protein-ligand complex.
%

\subsection*{Evaluation Metrics}

\paragraph{Metrics for SMG} 
%
To evaluate the performance of \method and SMG baselines in generating molecules with similar shapes to condition molecules, we use shape similarity $\shapesim$ and molecular graph similarity 
$\graphsim$ as evaluation metrics. 
%
%
Higher $\shapesim$ and lower \graphsim suggests that generated molecules could have similar binding activities and substantially different molecular graphs compared to condition molecules (e.g., ligands).
%
%
%
We calculate the shape similarity $\shapesim$ via the overlapped volumes between two aligned molecules following the literature~\cite{adams2023equivariant}.
Each generated molecule is aligned with the condition molecule by the ROCS tool~\cite{Hawkins2006}.
%
For the molecular graph similarity $\graphsim$, we use the Tanimoto similarity, calculated by RDKit~\cite{rdkit}, over Morgan fingerprints between the generated and condition molecule. 
Based on \shapesim and \graphsim, we calculate the following three metrics using the set of 50 generated molecules per condition molecule, and report the average of these metrics across all condition molecules in the test set:
%
(1) \desire calculates the percentage of molecules in each set with \shapesim$>$0.8 and \graphsim smaller than a threshold $\delta_g$, referred to as desirable molecules; 
(2) \diversity\ measures the diversity among desirable molecules within each set, calculated as 1 minus the average pairwise graph similarity; 
(3)  \novel calculates the percentage of desirable molecules in each set that cannot be found in the MOSES dataset. 
{Following Bostroem \etal~\cite{Bostroem2006}, we select 0.8 as the threshold of \shapesim for desirable molecules.
This threshold is chosen to ensure that the selected desirable molecules have highly similar shapes, and thus, similar binding activities to condition molecules.
During evaluation, we use test molecules as condition molecules for the generation.}

%

\paragraph{Metrics for PMG} 
We evaluate the performance of \method and PMG baselines in generating molecules binding towards protein targets. 
Following previous work~\cite{guan2023targetdiff, guan2023decompdiff}, we evaluate the binding affinities, drug-likeness, and diversity of generated molecules.
For binding affinity, we use Vina Scores (Vina S) calculated by AutoDock Vina~\cite{Eberhardt2021} as an evaluation metric. 
As suggested in the literature~\cite{guan2023targetdiff, guan2023decompdiff}, we also consider the optimized poses from the generated 3D molecules in evaluation.
Specifically, we use Vina Minimization (Vina M) and Vina Dock (Vina D) calculated from AutoDock Vina as evaluation metrics.
Vina M and Vina D optimize the poses by local energy minimization and global search optimization, respectively~\cite{Eberhardt2021}. 
%
%
%
%
%
%
%
For drug-likeness, we evaluate whether the generated molecules are drug-like using QED scores~\cite{Bickerton2012} and synthesizable using synthesizability scores (SA)~\cite{Ertl2009}.
We also calculate the diversity as defined in the previous paragraph among generated molecules. 
%
Following previous work~\cite{guan2023targetdiff, guan2023decompdiff}, we report the average and median of the above metrics across all test complexes. 
%

\paragraph{Metrics for Evaluation of Molecule Quality} 
We evaluate the quality of generated molecules based on their realism for both SMG and PMG.
We use a comprehensive set of metrics to evaluate the stability, 
3D structures, and 2D structures of generated molecules. 
For stability, following Hoogeboom \etal~\cite{hoogeboom22diff}, we calculate atom and molecule stability of generated molecules.
Atom stability measures the proportion of atoms that have the right valency, while molecule stability measures the proportion of generated molecules that all the atoms are stable. 
For 3D structures and 2D structures, we use the same metrics as in Peng \etal~\cite{peng2023moldiff}
Particularly,
%
%
for 3D structures, we use root mean square deviations (RMSDs) and Jensen-Shannon (JS) divergences of bond lengths, bond angles and dihedral angles to evaluate the quality of 3D molecule structures.
RMSDs measure the discrepancies between the generated 3D structures of molecules and their optimal structures identified by RDKit toolkit~\cite{rdkit} via energy minimization. 
%
%
%
In addition, the JS divergences of bond lengths, bond angles and dihedral angles measure the divergences between the generated molecules and the real molecules (i.e., training molecules) regarding the 3D structures.
Smaller divergence values indicate that the generated molecules have these properties more similar to those of training molecules and thus more realistic.
%
To evaluate the quality of 2D molecule structures, we primarily assess if the bonds and rings within the generated molecules are similar to those in real molecules.
Particularly, for bonds, we evaluate the JS divergences in terms of bond counts per atom and bond types (single, double, triple, and aromatic).
For rings, we compare both the counts of all rings and the counts of rings of varying sizes (n-sized rings) in the generated molecules to those in real molecules using JS divergences.
%
%
Furthermore, we measure if the generated molecules capture the frequent rings in real molecules. 
To be specific, we calculate the number of overlapping rings observed in the top-10 frequent ring types of both generated and real molecules.
Note that we consider different molecules to calculate JS divergences when comparing against SMG and PMG baselines.
 For SMG, we use the training molecules in the MOSES dataset.
For PMG, we use the ligands in the training protein-ligand complexes of CrossDocked2020.
%
%

\section*{Experimental Results}



\subsection*{Overall Comparison on Generating Desirable Molecules in SMG}
\label{sec:results:overall_desirable}


\input{tables/overall_results_desirable}

%
We evaluate \method, \methodwithsguide and state-of-the-art SMG baselines in generating desirable molecules.
%
%
Following Bostroem \etal~\cite{Bostroem2006}, we define desirable molecules as those satisfying $\delta_g$ and with shape similarities larger than 0.8 {(detailed in Section ``Evaluation Metrics'')}.
These molecules have highly similar shape with the condition molecules (e.g., ligands), and thus, could also have desirable binding activities~\cite{Acharya2011}. 
%
In this analysis, for each method, we calculate the possibility of generating desirable molecules (i.e., \desire), the diversity {and the novelty} among these molecules (i.e., \diversity and \novel) under different graph similarity constraints (i.e., $\delta_g$=0.3, 0.5, 0.7, and 1.0). 
As shown in Table~\ref{tbl:overall_desirable},
\method and \methodwithsguide consistently outperform all the baseline methods 
%
in terms of {all the metrics}. 
%
%
For example, when $\delta_g$=0.3, at \desire, \methodwithsguide (61.4\%) demonstrates a substantial improvement of $448.2\%$ compared to the best baseline \squid ($\lambda$=1.0) (11.2\%).
%
%
In terms of \diversity, \method (0.762) also substantially outperforms the best baseline \dataset (0.736) by 2.3\%. 
{At \novel, both \method and \methodwithsguide ensure that nearly all the generated desirable molecules are novel (99.8\% for \method and 99.9\% for \methodwithsguide), substantially outperforming the best baseline \squid with $\lambda$=1.0 (96.9\%) by 3.1\% and 3.0\%, respectively.}
%
When $\delta_g$=0.5, 0.7, or 1.0, a similar trend is observed.
%
Specifically, when $\delta_g$=0.5, at \desire, \methodwithsguide (70.9\%) establishes a notable improvement of 225.2\% compared to the best baseline method \squid with $\lambda$=0.3 (21.8\%).
{At \diversity and \novel, \method and \methodwithsguide also demonstrate top performance among all the methods. }
When $\delta_g$=0.7, at \desire, \methodwithsguide (71.0\%) also achieves a remarkable improvement of 158.2\% over the best baseline method \squid with $\lambda$=0.3 (27.5\%).
%
%
%
%
%
%
The superior performance of \method and \methodwithsguide in \desire, \diversity, and \novel,  particularly at small $\delta_g$, indicates their strong capacity in generating {novel molecules that have desirable shapes and distinct graph structures compared to the condition molecules, }
thereby facilitating the process of drug development.
Additional results about the comparison of shape similarity and graph similarity and the comparison of validity and novelty are available in Supplementary Section~\ref{supp:app:results:overall_shape} and \ref{supp:app:results:valid_novel}, respectively. 

%
%
%

It is worth noting that, as shown in Table~\ref{tbl:overall_desirable}, \method and \methodwithsguide consistently outperform \squid with $\lambda$=0.3 and 1.0 in terms of all the metrics. 
A key distinction between \method and \squid is that \squid generates molecules by sequentially predicting fragments. 
However, during fragment prediction, their poses are not fully considered, leading to suboptimal prediction accuracy and limited generation performance.
%
%
%
%
%
On the other hand, by directly arranging atoms, \method and \methodwithsguide explicitly consider the 3D atom positions when generating molecules, and thus, achieve remarkable improvement over \squid as shown in Table~\ref{tbl:overall_desirable}. 
%

{Different from \method and \squid which directly generate desirable molecules, \dataset screens over randomly sampled training molecules to identify molecules of interest. 
However, it cannot ensure optimal precision design, 
resulting in the suboptimal performance of \dataset at \desire. 
In addition, due to the reliance on existing molecules, \dataset cannot discover novel molecules. 
In contrast, \method can effectively generate novel molecules with desirable shapes, making it a promising tool for discovering novel drug candidates.
}

Comparing \methodwithsguide and \method, Table~\ref{tbl:overall_desirable} shows that incorporating shape guidance into \method substantially boosts its effectiveness in generating desirable molecules.
For example, when $\delta_g$=0.3, at \#d\%, \methodwithsguide (61.4\%) substantially outperforms \method (28.4\%) by 116.2\%.
when $\delta_g$=0.5, 0.7, and 1.0, \methodwithsguide also achieve a considerable improvement  of 119.5\%, 119.1\% and 119.1\%, respectively, compared to \method.
In the meantime, \methodwithsguide retains very similar performance with \method in terms of the diversity and novelty of generated desirable molecules.
These results signify that shape guidance effectively improves the ability of \method in generating molecules that have similar shapes to condition molecules without degrading the novelty and diversity among generated molecules.

\subsection*{Quality Comparison between Desirable Molecules Generated by \method and \squid}
\label{sec:results:quality_desirable}


\input{tables/overall_results_quality_desired}


We also evaluate the quality of desirable molecules 
%
generated from \method, \methodwithsguide, and baseline methods in terms of stability, 3D structures, and 2D structures. 
%
%
Table~\ref{tbl:overall_results_quality_desired} presents the performance comparison 
in the quality of
desirable molecules generated by different methods when the graph similarity constraint $\delta_g$ is 0.3.
Details about the comparison under different $\delta_g$ (e.g., 0.5, 0.7, and 1.0) are available in Supplementary Section~\ref{supp:app:results:quality_desirable}. 
%
%
{Note that, in this analysis, we focus on desirable molecules that could have high utility in drug development.}
We also exclude the search algorithm \dataset and consider only generative models, such as \method and \squid, in this analysis. 

As shown in Table~\ref{tbl:overall_results_quality_desired}, \method generates molecules with 
comparable quality
to baselines in terms of stability, 3D structures, and 2D structures. 
For example, in stability, Table~\ref{tbl:overall_results_quality_desired} shows that \method and \methodwithsguide 
either achieve comparable performance or fall slightly behind 
\squid ($\lambda$=0.3) and \squid ($\lambda$=1.0) in atom stability and molecule stability.
Particularly, \method achieves similar performance with 
\squid ($\lambda$=0.3) and \squid ($\lambda$=1.0) in atom stability (0.993 for \method vs 0.996 for \squid with $\lambda$ of 0.3 and 1.0).
%
%
%
%
In terms of molecule stability, \method underperforms {\squid} ($\lambda$=0.3) by 6.5\%.
However, \method still demonstrates strong effectiveness in generating stable molecules, with 89.1\% of generated molecules being stable. 

Table~\ref{tbl:overall_results_quality_desired} also shows that \method and \methodwithsguide generate molecules with more realistic 3D structures compared to \squid.
%
%
Particularly, for RMSD, 
\method and \methodwithsguide outperform the best baseline \squid ($\lambda$=1.0)  
by 0.8\% and 2.2\%,
respectively.
In addition, they also establish a notable improvement of 4.6\% and 6.3\% 
over the best baseline \squid ($\lambda$=0.3) in JS. bond lengths.
In terms of JS. bond angles and JS. dihedral angles, \method and \methodwithsguide outperform the best baseline \squid ($\lambda$=0.3) by 30.9\% and 25.7\%, and by  15.6\% and 14.6\%, respectively.
%
%
%
As discussed in Section ``Related Work'', \squid fixes the bond lengths and angles within the generated molecules, leading to the discrepancy in 3D structures between the generated molecules and real molecules.
Conversely, \method and \methodwithsguide use a data-driven manner to infer distances and angles between atoms.
This design enables  \method and \methodwithsguide to achieve superior performance in generating molecules with realistic 3D structures. 

Table~\ref{tbl:overall_results_quality_desired} also presents that \method and \methodwithsguide achieve comparable performance with  \squid in generating realistic 2D molecule structures. 
Particularly, for JS. \#bonds per atom, \method and \methodwithsguide substantially outperform the best baseline \squid ($\lambda$=0.3) by 77.8\% and 73.9\%, respectively. 
In terms of JS. basic bond types, \method and \methodwithsguide underperform \squid ($\lambda$=0.3) considerably.
\method and \methodwithsguide also achieve substantially better performance (0.042 and 0.048) than \squid with $\lambda$=0.3 (0.309) in JS. \#rings.
\method and \methodwithsguide slightly underperform \squid in terms of JS. \#n-sized rings and the number of intersecting rings. 
%
%
%
%
These results signify that \method and \methodwithsguide, though not explicitly leverage fragments 
%
%
as \squid does, can still generate molecules with realistic 2D structures.
%

\subsubsection*{Analysis on Shape and Graph Similarities}
\label{sec:results:analysis_similarity}

\begin{figure}[!ht]
	\centering
	\begin{subfigure}[b]{.24\linewidth}
		\centering
		\includegraphics[width=\linewidth]{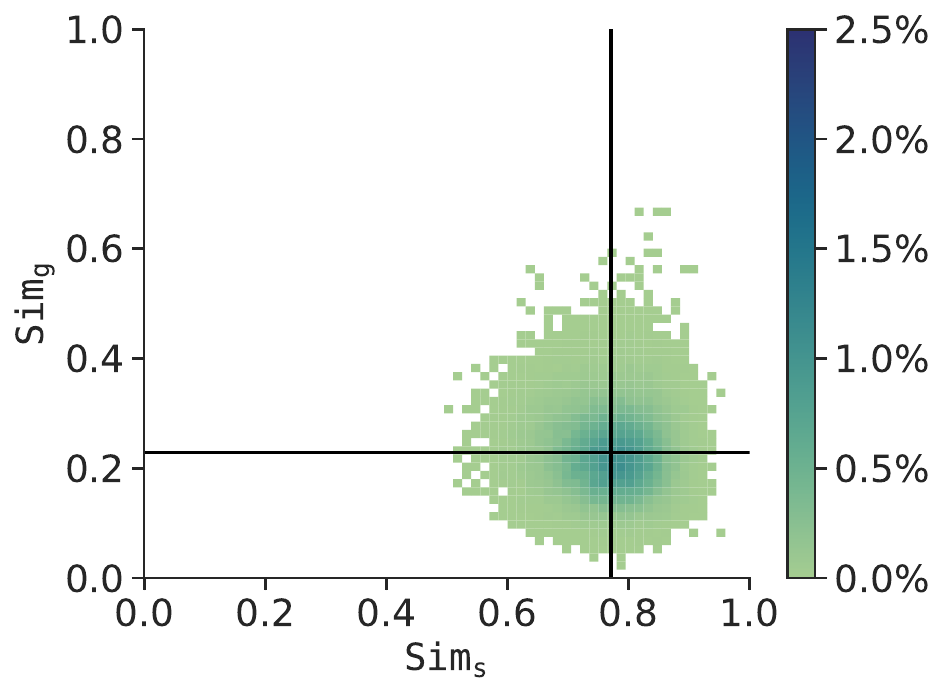}
		\caption{\method}
		\label{fig:sim_comparison：method}
	\end{subfigure}
	\begin{subfigure}[b]{.24\linewidth}
		\centering
		\includegraphics[width=\linewidth]{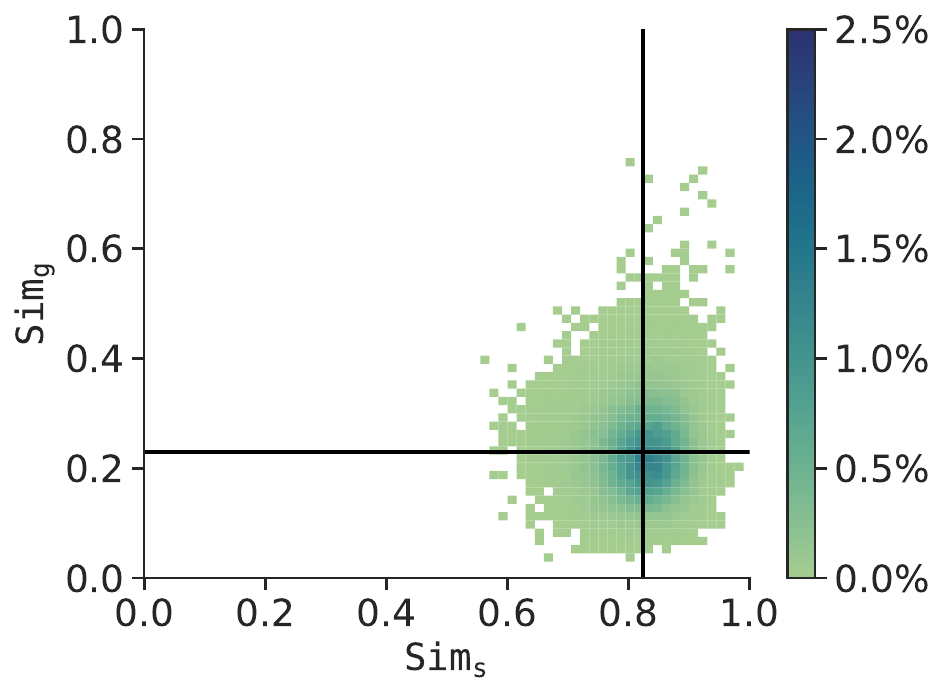}
		\caption{\methodwithsguide}
		\label{fig:sim_comparison：methodwithsguide}
	\end{subfigure}
	\begin{subfigure}[b]{.24\linewidth}
		\centering
		\includegraphics[width=\linewidth]{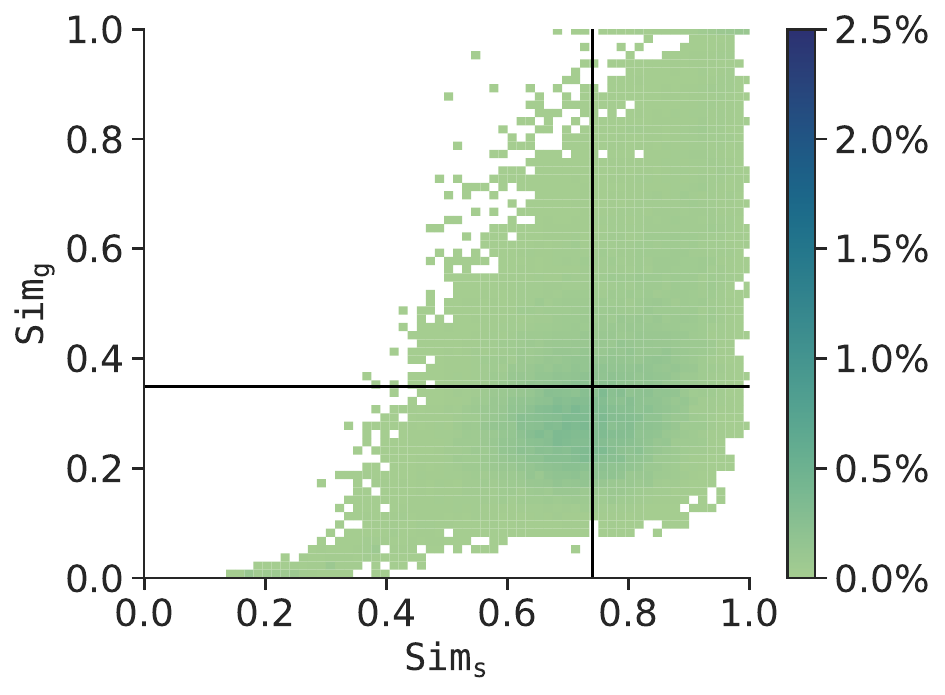}
		\caption{\squid ($\lambda$=0.3)}
		\label{fig:sim_comparison：squid03}
	\end{subfigure}
	\begin{subfigure}[b]{.24\linewidth}
		\centering
		\includegraphics[width=\linewidth]{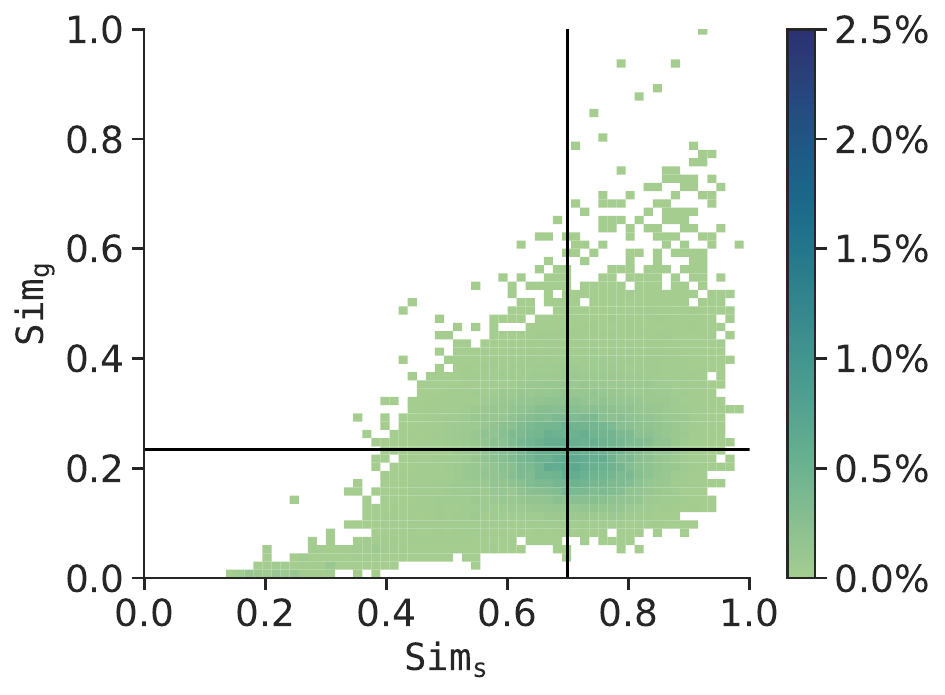}
		\caption{\squid ($\lambda$=1.0)}
		\label{fig:sim_comparison：squid10}
	\end{subfigure}
	%
	\vspace{-5pt}
	\captionsetup{justification=centering}
	\caption{\textbf{Heatmaps of Similarities Calculated from Molecules Generated by 
	\squid and \method.}}
	\label{fig:sim_comparison}
\end{figure}

We analyze the distributions of shape and graph similarities between condition molecules and all molecules generated from \method and \squid.
%
%
We conduct this analysis to
(1) assess the capacity of \method and \squid in generating molecules with similar shapes to condition molecules (e.g., ligands); and
(2) compare the strategies in \method and \squid to further improve shape similarities. 
%
In this analysis, for each condition molecule, we use all 50 molecules generated by \method, \methodwithsguide, \squid with $\lambda$=0.3 and \squid with $\lambda$=1.0.  
%
As shown in Fig.~\ref{fig:sim_comparison}, for each method, we visualize a heatmap with the x-axis representing shape similarities (\shapesim) and the y-axis representing graph similarities (\graphsim).
Each grid in this heatmap shows the percentage of molecules that have shape and graph similarities within specific ranges, and the grid color represents the scale of the percentage (e.g., a darker color indicates a higher percentage).
In each heatmap, the vertical black line marks the average of shape similarities, and the horizontal black line marks the average of graph similarities.
%


%
%

%

%

Fig.~\ref{fig:sim_comparison} demonstrates the exceptional performance of \method and \methodwithsguide in generating molecules with high shape similarity to condition molecules.
Particularly, in terms of shape similarity, Fig.~\ref{fig:sim_comparison}(a) and~\ref{fig:sim_comparison}(b) show that \method and \methodwithsguide have 99.5\% and 100.0\% of generated molecules with \shapesim$>$ 0.6. 
In contrast, \squid with $\lambda$=0.3 and 1.0 generate 89.9\% and 86.3\% of molecules with \shapesim$>$ 0.6, respectively. %
It is worth noting that \squid could generate molecules that are dramatically different from condition molecules in terms of shapes and have \shapesim$<$ 0.2, while all the generated molecules from \method and \methodwithsguide have considerably similar shapes to the conditions.
%
%

{Both \method and \squid develop specific strategies to enhance the shape similarity.
Particularly, \method incorporates shape guidance (i.e., \methodwithsguide) to iteratively modify the generated 3D structures to better resemble the known ligand shapes.
On the other hand, \squid leverages a balance variable $\lambda$ to control the interpolation level during generation. 
A lower $\lambda$ indicates stronger interpolation, and thus, better shape similarity but worse graph similarity.
According to Fig.~\ref{fig:sim_comparison}, the shape guidance in \methodwithsguide effectively boosts the shape similarities of generated molecules without degrading their graph similarities.
Specifically, Fig.~\ref{fig:sim_comparison}(a) and~\ref{fig:sim_comparison}(b) show that compared to \method, \methodwithsguide achieves a higher average shape similarity (0.824 for \methodwithsguide vs 0.771 for \method) and comparable average graph similarity (0.230 for \methodwithsguide and 0.229 for \method).
However, we observe a different trend for \squid in Fig.~\ref{fig:sim_comparison}(c) and~\ref{fig:sim_comparison}(d): by decreasing $\lambda$ from 1.0 to 0.3, there exists a trade-off between shape similarity and graph similarity.
To be specific, \squid ($\lambda$=0.3) achieves superior average shape similarity (0.740) while inferior average graph similarity (0.349), compared to \squid ($\lambda$=1.0) (0.699 for average shape similarity and 0.235 for average graph similarity). %
%
%
%
%
%
%
These results suggest that compared to adjusting the interpolation level ($\lambda$) as in \squid, including shape guidance could more effectively enhance shape similarities of generated molecules without compromising graph similarities. }

\subsection*{{Case Study for SMG}}


Fig.~\ref{fig:example_shape2} presents three generated molecules from \dataset, \squid with $\lambda$=0.3 and \methodwithsguide given the same condition molecule.
Each molecule has the highest shape similarity among the 50 candidates generated by each method.
As shown in Fig.~\ref{fig:example_shape2}, the molecule generated by \methodwithsguide has higher shape similarity (0.883) with the condition molecule than those 
from the baseline methods (0.768 for \dataset and 0.759 for \squid with $\lambda$=0.3).
Particularly, the molecule from \methodwithsguide has the {surface shape} 
(represented as blue shade in Fig.~\ref{fig:ours}) most 
similar to that of the condition molecule. 
On the contrary, the molecules generated from \dataset and \squid with $\lambda$=0.3 show noticeable misalignments when compared to the condition molecule. 
This comparison demonstrates the superior ability of \methodwithsguide to generate molecules with highly similar 3D shapes to the condition molecule.
In terms of graph similarities, all these generated molecules have low graph similarities with the condition molecule. 
The ability to generate molecules that have similar 3D shapes yet different molecular graphs demonstrates the potential high utility of \methodwithsguide in facilitating the drug development.

\begin{figure}[!h]
	\centering
	\begin{subfigure}[b]{.48\linewidth}
		\centering
		\begin{subfigure}[b]{.45\linewidth}
			\centering
			\includegraphics[width=0.75\linewidth]{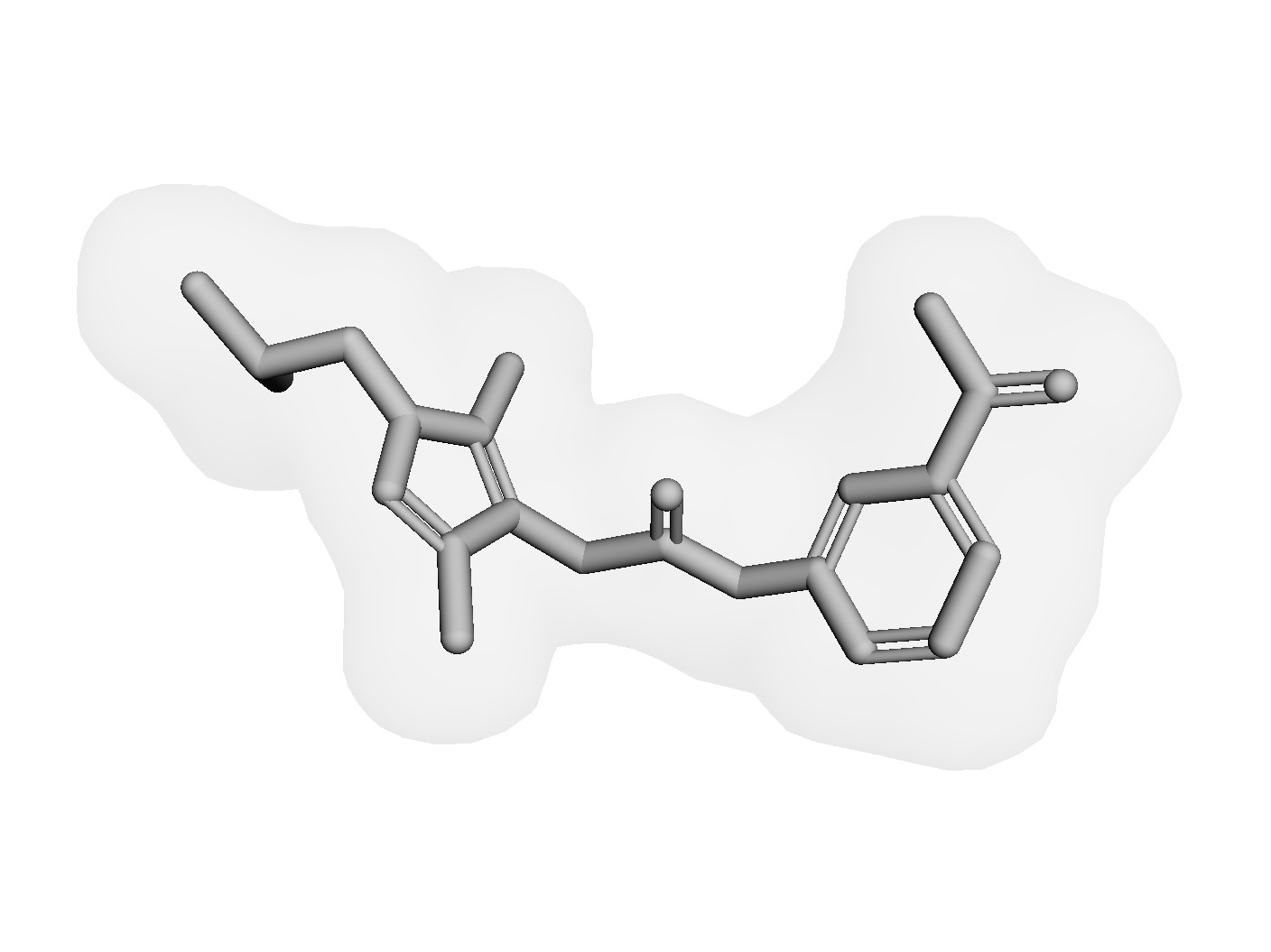}
		\end{subfigure}
		\begin{subfigure}[b]{.45\linewidth}
			\centering
			\includegraphics[width=.68\linewidth]{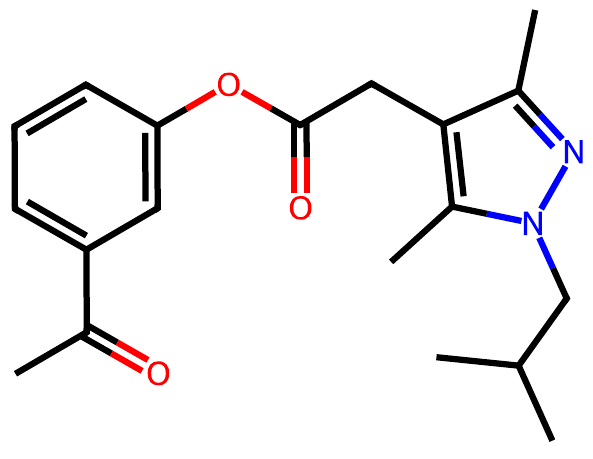}
		\end{subfigure}
		\captionsetup{justification=centering}
		\caption{Condition Molecule}
	\end{subfigure}
	\begin{subfigure}[b]{.48\linewidth}
		\centering
		\begin{subfigure}[b]{.32\linewidth}
			\centering
			\includegraphics[width=\linewidth]{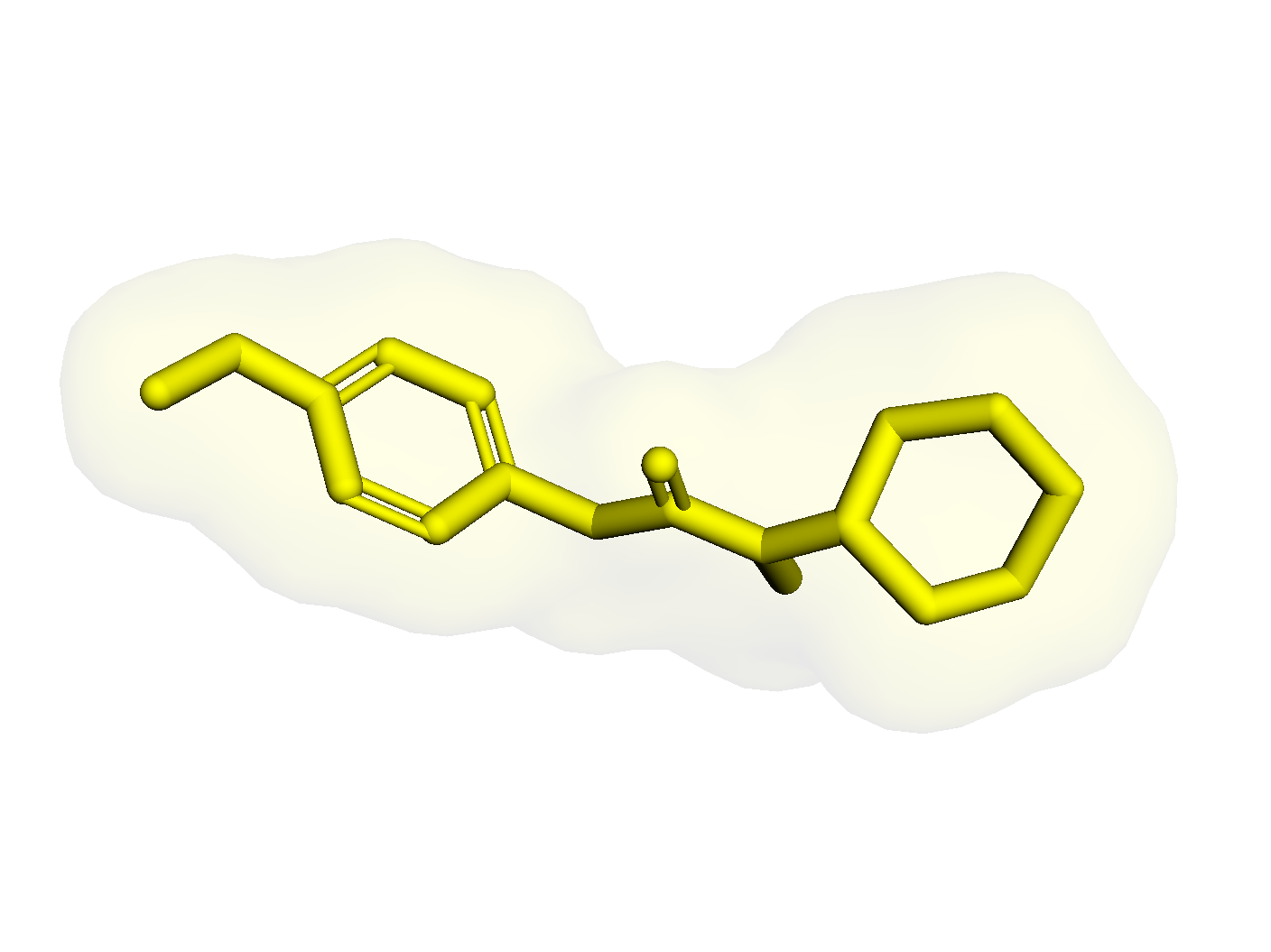}
		\end{subfigure}
		\begin{subfigure}[b]{.32\linewidth}
			\centering
			\includegraphics[width=\linewidth]{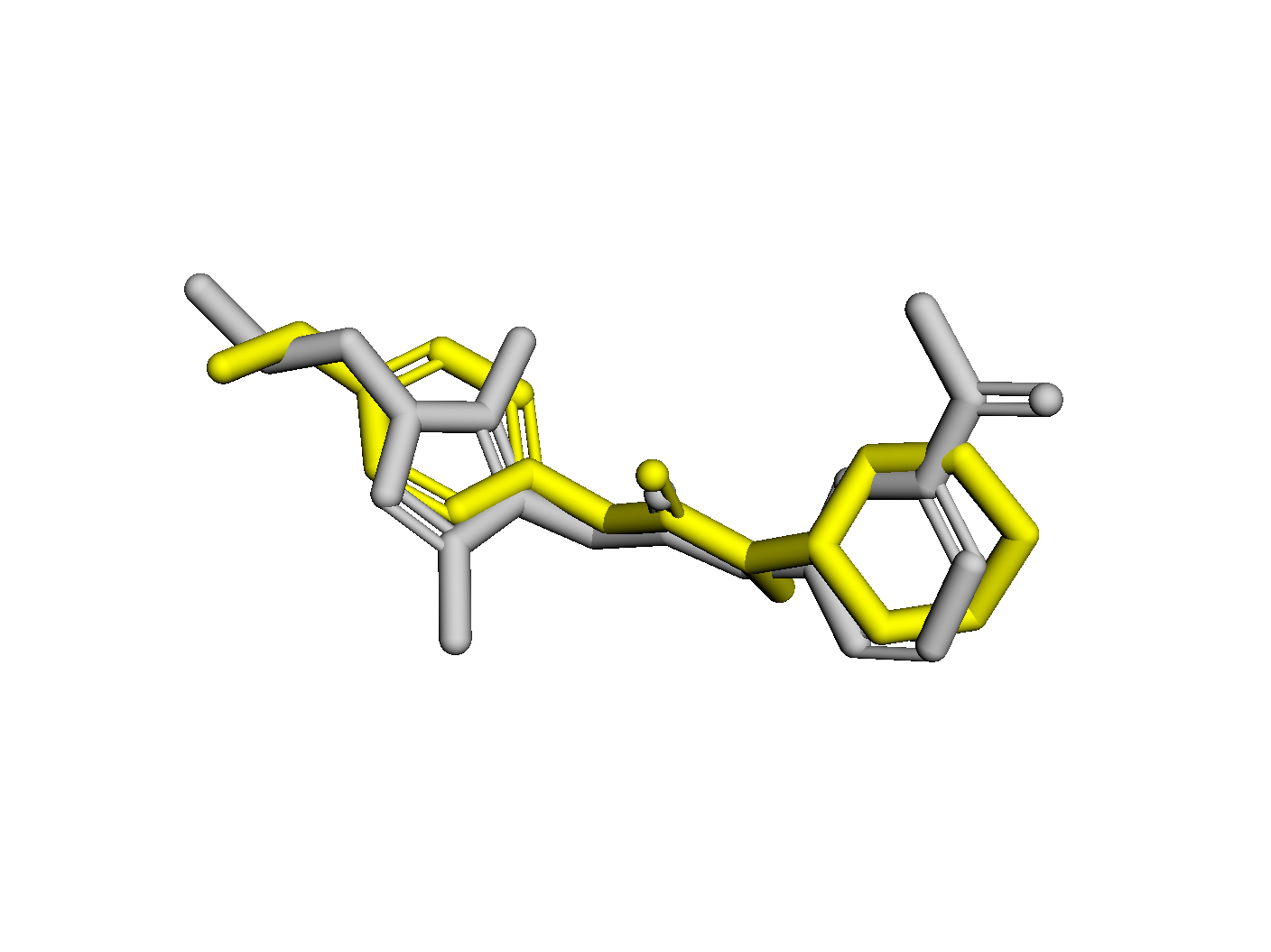}
		\end{subfigure}
		\begin{subfigure}[b]{.32\linewidth}
			\centering
			\includegraphics[width=\linewidth]{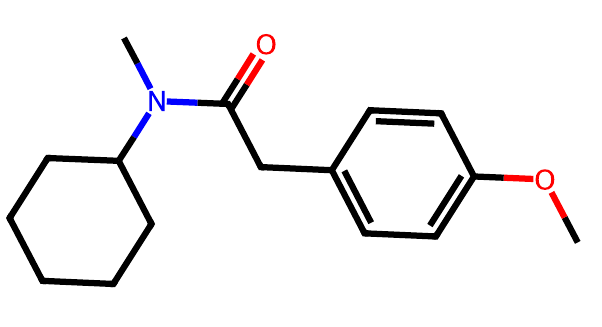}
		\end{subfigure}
		\captionsetup{justification=centering}
		\caption{Molecule from \dataset: \shapesim = 0.768, \graphsim = 0.237} 
	\end{subfigure}
	\\
	\begin{subfigure}[b]{.48\linewidth}
		\centering
		\begin{subfigure}[b]{.32\linewidth}
			\centering
			\includegraphics[width=\linewidth]{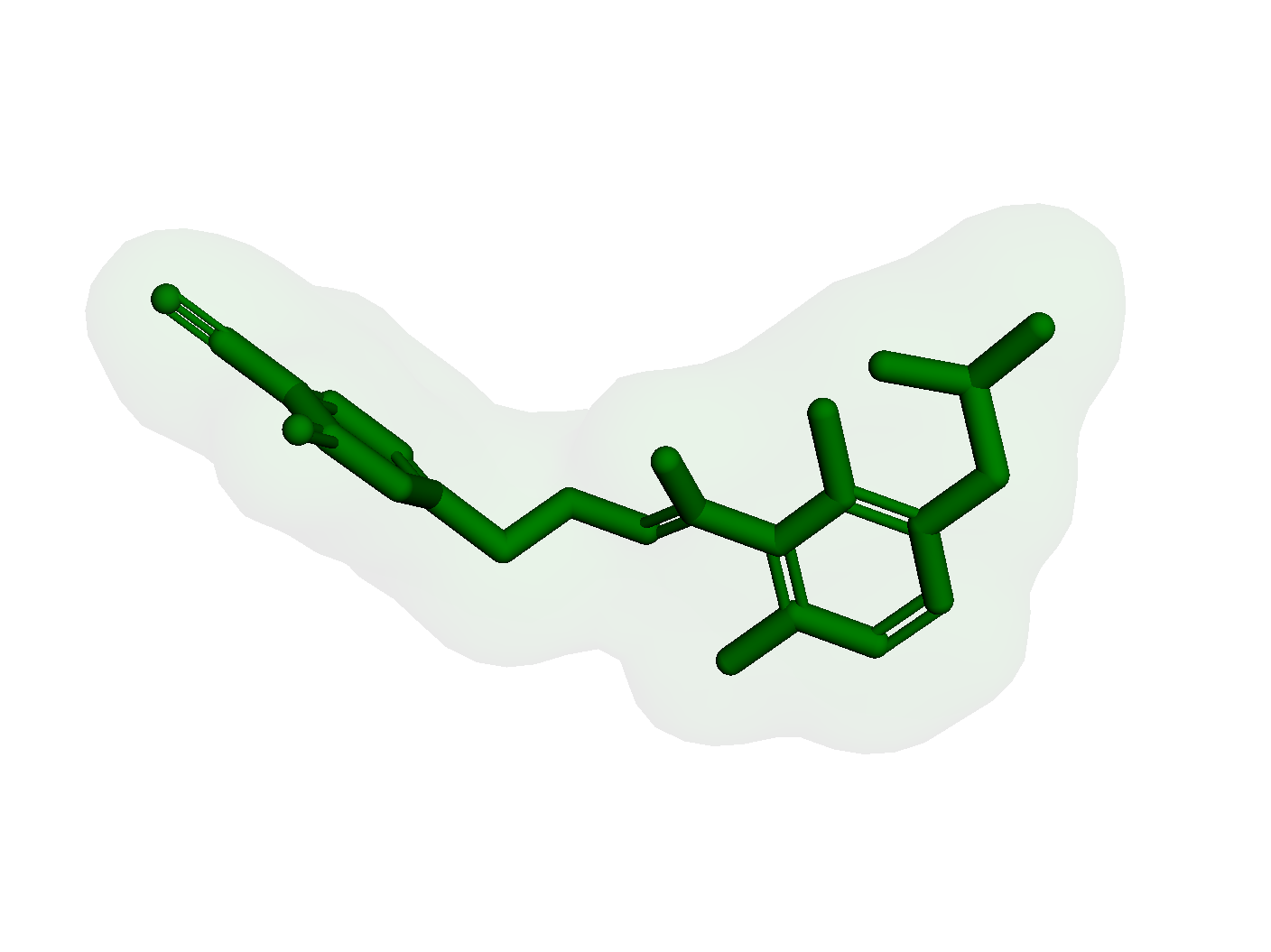}
		\end{subfigure}
		\begin{subfigure}[b]{.32\linewidth}
			\centering
			\includegraphics[width=\linewidth]{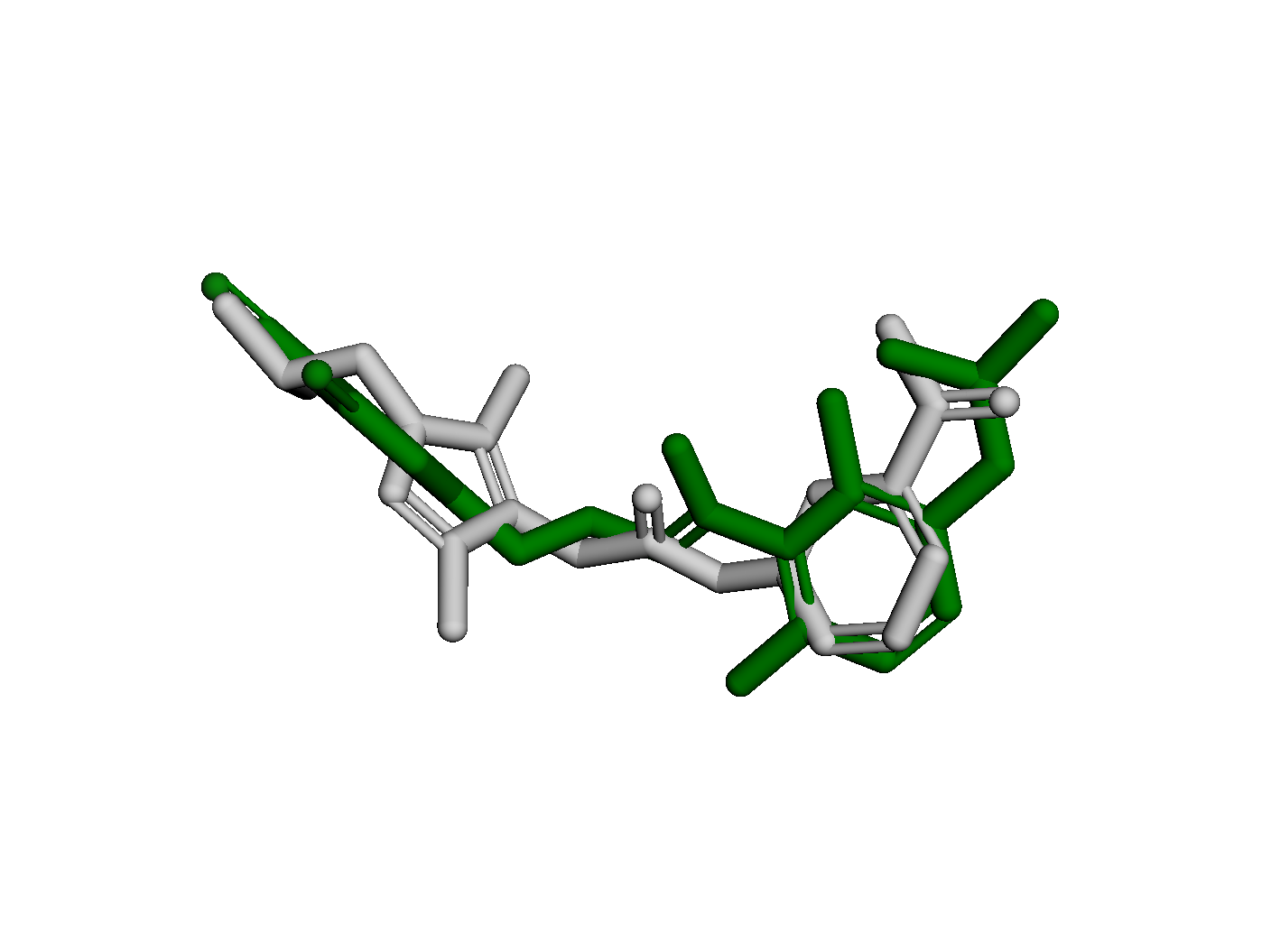}
		\end{subfigure}
		\begin{subfigure}[b]{.32\linewidth}
			\centering
			\includegraphics[width=\linewidth]{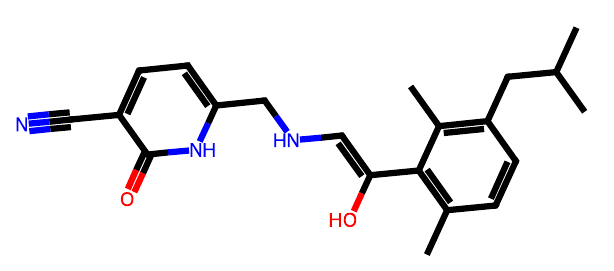}
		\end{subfigure}
		\captionsetup{justification=centering}
		\caption{Molecule from \squid with $\lambda$=0.3: \shapesim = 0.759, \graphsim = 0.280}
	\end{subfigure}
	\begin{subfigure}[b]{.48\linewidth}
		\centering
		\begin{subfigure}[b]{.32\linewidth}
			\centering
			\includegraphics[width=\linewidth]{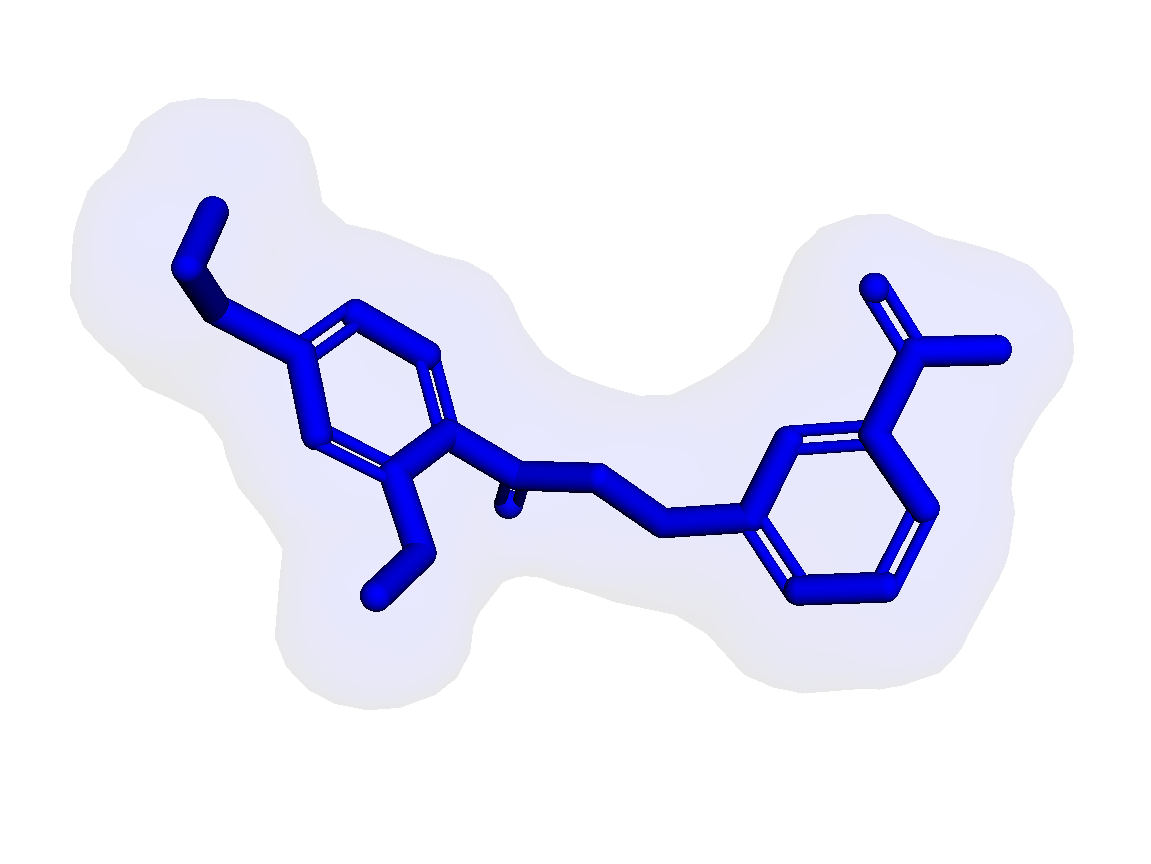}
		\end{subfigure}
		\begin{subfigure}[b]{.32\linewidth}
			\centering
			\includegraphics[width=\linewidth]{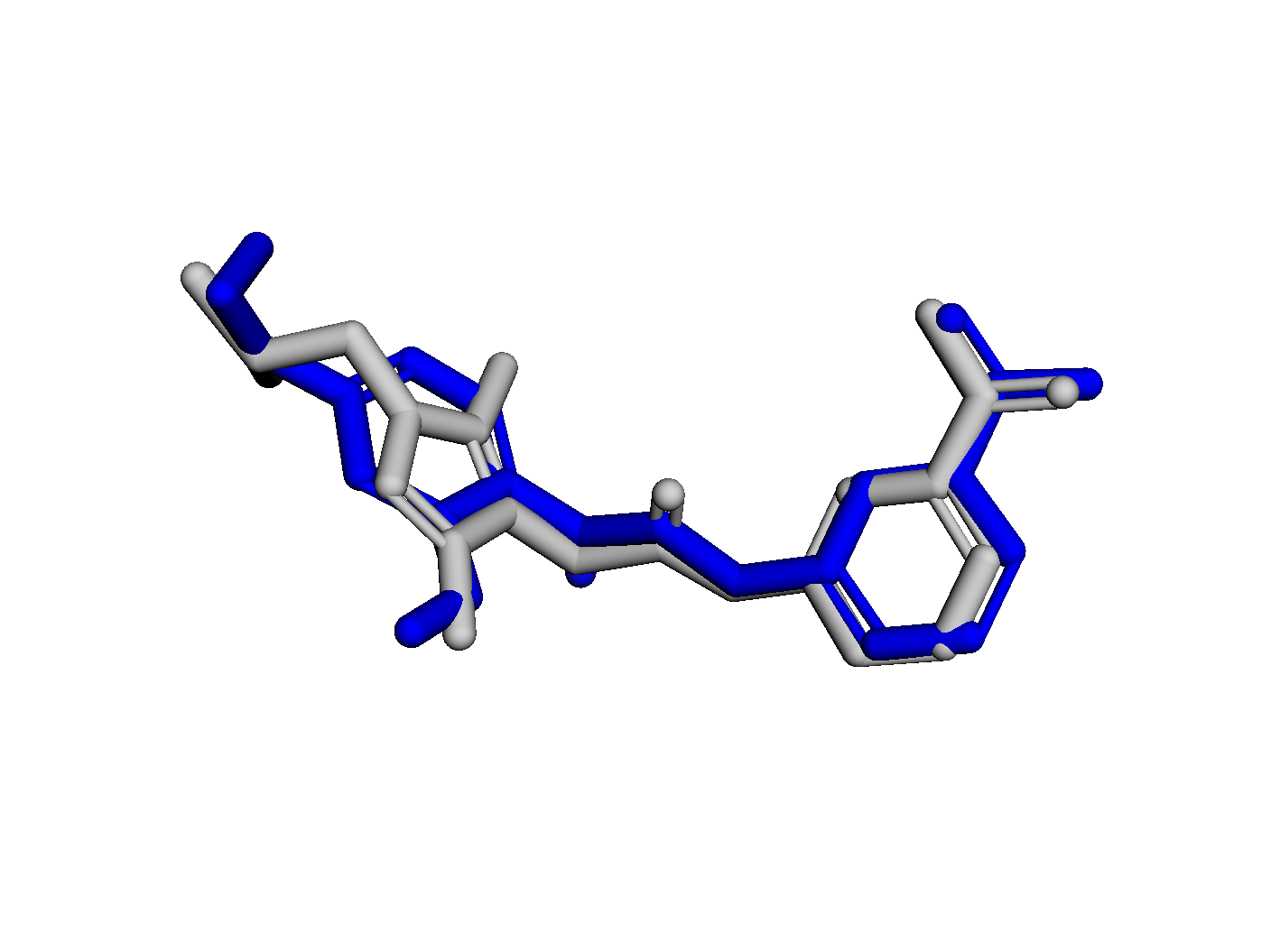}
		\end{subfigure}
		\begin{subfigure}[b]{.32\linewidth}
			\centering
			\includegraphics[width=\linewidth]{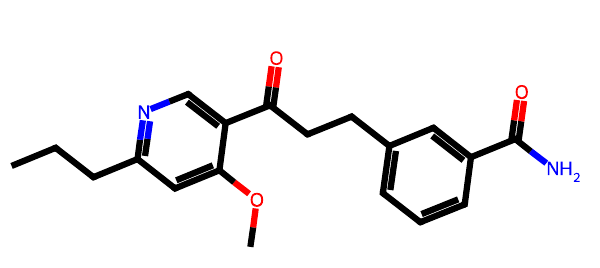}
		\end{subfigure}
		\captionsetup{justification=centering}
		\caption{Molecule from \methodwithsguide: \shapesim = 0.883, \graphsim = 0.317}
		\label{fig:ours}
	\end{subfigure}
	\caption{\textbf{Generated 3D Molecules from Different Methods. } Molecule 3D shapes are in shades; generated molecules are superpositioned with the condition molecule; and the molecular graphs of generated molecules are presented.}
	\label{fig:example_shape2}
\end{figure}

\subsection*{Overall Comparison for PMG}



\input{tables/overall_results_docking3}

From this section, we shift our focus from evaluating against SMG baselines to PMG baselines, methods that leverage protein pockets for binding molecule generation (see Section ``Related Work'' for details).
We evaluate the effectiveness of \method against state-of-the-art PMG baselines (see Section “Baselines” for details) in generating molecules binding towards specific protein pockets.
All the baselines require protein-ligand complex data for training and generate molecules by explicitly modeling their interactions with protein binding pockets.
Different from these baselines, \method does not require complex data and consumes molecules for training. 
%
%
{Note that protein-ligand complex data is expensive and thus highly limited.
In contrast, there exist several high-quality and large-scale molecule databases~\cite{zinc22,mose2020}. 
By consuming molecule data for training, \method could fully leverage the rich data for better generation.}
{
%
\method further enhances the binding molecule generation by incorporating pocket guidance as detailed in Section ``\method with Pocket Guidance''.}
%
%

We utilize two variants of \method for evaluation: \method with pocket guidance (\methodwithpguide) and \method with both pocket and shape guidance (\methodwithsandpguide).
%
%
In this section, we evaluate \methodwithpguide, \methodwithsandpguide and PMG baselines in both effectiveness and efficiency.
Following previous work~\cite{guan2023targetdiff, guan2023decompdiff}, in terms of the effectiveness, we evaluate the binding affinity, drug-likeness, and diversity 
of molecules generated from  \methodwithpguide, \methodwithsandpguide and all PMG baselines.
Please refer to Section ``Evaluation Metrics'' for a detailed description for the evaluation metrics.
Regarding efficiency, we report the inference time of all methods used to generate molecules.

We notice that \methodwithpguide and \methodwithsandpguide show remarkable efficiency over baselines by using pocket guidance instead of directly modeling these pockets.
Specifically, {\methodwithpguide and \methodwithsandpguide}
generate 100 molecules in 48 and 58 seconds on average, respectively, while the most efficient baseline \targetdiff takes 1,252 seconds.
The superior efficiency enables \methodwithpguide and \methodwithsandpguide to generate more than 10x molecules than baselines in the same time duration.
Therefore, following Long \etal~\cite{long2022zero}, for baselines, we apply them to generate 100 molecules for each test protein target.
For \method, we generate 1,000 molecules and select the top 100 molecules for comparison based on their Vina S, QED, and SA scores.
We report the performance of all methods in Table~\ref{tbl:overall_results_docking2}.
Additionally, for a more comprehensive comparison, we present the results of \methodwithpguide and \methodwithsandpguide when generating 100 molecules 
in the Supplementary Section~\ref{supp:app:results_PMG}.

%
As shown in Table~\ref{tbl:overall_results_docking2},  \methodwithpguide and \methodwithsandpguide 
achieve the second-best and best performance in terms of the binding affinities of generated molecules.
Particularly, they demonstrate the second-best (-5.53 kcal/mol) and best (-5.81 kcal/mol) average Vina S, with 17.7\% and 13.2\% improvement over the best baseline \AR (-5.06 kcal/mol).
%
%
Similarly, for vina scores obtained from locally minimized poses (i.e., Vina M), they also achieve the second-best (-6.37 kcal/mol) and best (-6.50 kcal/mol) performance, 
outperforming the best baseline \targetdiff (-6.20 kcal/mol) by 11.5\% and 9.5\%.
%
%
%
Moreover, for Vina D, a score calculated from poses optimized by global search,  \methodwithpguide and \methodwithsandpguide again yield the second-best (-7.19 kcal/mol) and third-best performance (-7.16 kcal/mol) and only slightly underperform the best baseline \targetdiff (-7.37 kcal/mol). 
%
%
These results demonstrate that even without explicitly training on protein-ligand complexes, 
\methodwithpguide and \methodwithsandpguide could still generate molecules with superior binding affinities towards protein targets in terms of Vina S, Vina M, and Vina D, compared to state-of-the-art PMG baselines.
Note that we consider Vina M and Vina D in evaluation, as it is a common practice in drug development to find more favorable binding poses of candidates through pose optimization~\cite{Ferreira2015}.

Table~\ref{tbl:overall_results_docking2} also shows that \methodwithpguide and \methodwithsandpguide are able to generate molecules with better binding affinities than condition molecules (i.e., known ligands).
Particularly, they achieve the best (79.92\%) and second-best (78.75\%) performance in terms of the average percentage of generated molecules with Vina D higher than those of known ligands (i.e., HA).
The superior performance in HA demonstrates the high utility of \methodwithpguide and \methodwithsandpguide in generating promising drug candidates with better binding affinities than known ligands.
%

Table~\ref{tbl:overall_results_docking2} further presents the superior performance of \methodwithpguide and \methodwithsandpguide in metrics related to \mbox{drug-likeness} {and diversity}. 
%
%
Particularly, {for drug-likeness}, they achieve the best (0.77) and second-best (0.76) QED scores, respectively, with 31.0\% and 29.3\% improvement over the best baseline \pockettwomol (0.58).
\methodwithpguide and \methodwithsandpguide also achieve the second (0.72) and third (0.70) SA scores, and only slightly fall behind 
the best baseline \pockettwomol (0.76).
In terms of the {diversity} 
among generated molecules, \methodwithpguide and \methodwithsandpguide underperform the baselines.
The inferior diversity can be attributed to the design of \methodwithpguide and \methodwithsandpguide that generates molecules with similar shapes to the ligands. 
This design allows \methodwithpguide and \methodwithsandpguide to generate molecules with desired drug-likeness while could slightly degrade the diversity among generated molecules.

%
%
%
%
%

When comparing \methodwithpguide and \methodwithsandpguide, Table~\ref{tbl:overall_results_docking2} shows that overall, \methodwithsandpguide with shape guidance can generate molecules with higher binding affinities compared to \methodwithpguide.
To be specific, for Vina S and Vina M, \methodwithsandpguide outperforms \methodwithpguide by 5.1\% and 2.0\%, respectively.
At Vina D, the performance of \methodwithsandpguide and \methodwithpguide is highly comparable.
These results indicate that even with pocket guidance, including additional shape guidance could further enhance the generation of binding molecules.
\subsection*{Quality Comparison for PMG}


%
%
In addition to binding affinities, drug-likeness, and diversity, we also evaluate the quality of molecules generated by \methodwithpguide, \methodwithsandpguide, and all the PMG baselines.
We assess the quality of these molecules across multiple dimensions, including stability, 3D structures, and 2D structures, using the same metrics as in Table~\ref{tbl:overall_results_quality_desired}.
To ensure a fair comparison, instead of using molecules from the MOSES dataset to calculate the JS divergence metrics as in Table~\ref{tbl:overall_results_quality_desired}, we use the known ligands from the baselines' training set (i.e., CrossDocked2020) to calculate JS divergences.  
We report the performance of all methods in terms of molecule quality in Table~\ref{tbl:overall_docking_results_quality_10}. 
%
%

%
Table~\ref{tbl:overall_docking_results_quality_10} shows that \methodwithpguide and  \methodwithsandpguide achieve higher or at least comparable performance with all baselines 
in most quality metrics.
Specifically, for stability, Table~\ref{tbl:overall_docking_results_quality_10} shows that \methodwithpguide and \methodwithsandpguide either achieve 
comparable performance or slightly fall behind the baselines in atom stability and molecule stability.
Particularly, \methodwithpguide achieves the second-best performance in atom stability and only slightly underperforms the best baseline \targetdiff (0.934 vs 0.949).
\methodwithpguide also achieves the best performance in molecule stability. 
%
\methodwithsandpguide underperforms \methodwithpguide in both atom stability and molecule stability but still outperforms \pockettwomol and \AR in atom stability and \pockettwomol, \targetdiff, and \decompdiff in molecule stability.
%
%
%
%
%
These results demonstrate the effectiveness of \methodwithpguide and  \methodwithsandpguide in generating binding molecules with high 
stability.

In terms of 3D structures, overall, both \methodwithpguide and  \methodwithsandpguide achieve similar performance 
compared to the baselines.
%
%
Particularly, \methodwithpguide and \methodwithsandpguide achieve the best (0.274) and second-best performance (0.278) in terms of JS. bond lengths.
For JS. dihedral angles, they also outperform the best baseline \decompdiff by 8.9\% and 8.4\%, respectively. 
We also note that, in terms of RMSD, \methodwithpguide and \methodwithsandpguide underperform the best baseline \pockettwomol, and achieve very comparable performance (0.663 for \methodwithpguide and 0.675 for \methodwithsandpguide) with the second-best baseline \AR (0.656).
For JS. bond angles, both \methodwithpguide and \methodwithsandpguide again underperform the best baseline \decompdiff, and achieve the second and fourth performance among all the methods.
%
%
%
%
The overall comparable performance of \methodwithpguide and  \methodwithsandpguide against the PMG methods in these metrics demonstrates their ability to generate molecules with realistic 3D structures.  
%
%
%
%
%

{For 2D structures, both \methodwithpguide and \methodwithsandpguide demonstrate comparable performance with the PMG baselines.
Specifically,  for JS. basic bond types, \methodwithpguide and \methodwithsandpguide achieve the second and third performance (0.061 and 0.080), and only slightly underperform the best baseline \pockettwomol (0.055).
For JS. \#rings, they also achieve the best and second performance among all the methods. 
Similarly, in terms of the number of intersecting rings, \methodwithsandpguide again achieves the best performance, while \methodwithpguide slightly underperforms \methodwithpguide by just one ring (6 vs 7).}
We also note that \methodwithpguide and \methodwithsandpguide underperform the best baseline \targetdiff in JS. \#bonds per atom.
For JS. \#n-sized rings, they also underperform the best baseline \pockettwomol and achieve the second and third performance.
%
%
These results highlight that compared to the state-of-the-art PMG methods, \methodwithpguide and \methodwithsandpguide enjoy similar performance in generating molecules with realistic 2D structures. 
%


\input{tables/overall_docking_results_quality1.0}

\subsection*{Case Studies for Targets}

\method can generate binding molecules that serve as promising drug candidates.
%
%
To demonstrate this ability, we highlight three molecules generated by \methodwithsandpguide for two crucial drug targets, cyclin-dependent kinase 6 (CDK6) and neprilysin (NEP).
CDK6 plays a critical role in cell proliferation by regulating cell cycle progression.
Inhibiting CDK6 can disrupt the abnormal cell cycles of cancer cells, making it a valuable therapeutic target for cancers~\cite{Tadesse2015}.  
NEP can help prevent amyloid plaque formation associated with Alzheimer's disease, making it an important target for therapies to potentially slow the disease's progression~\cite{ElAmouri2008}. 
We use an existing protein-ligand complex for CDK6 (PDBID:4AUA), and an existing protein-ligand complex 
for NEP (PDBID:1R1H), respectively, from Protein Data Bank (PDB)~\cite{Burley2022} and generate 1,000
molecules for each of the targets. 
Both complexes are included in our test set for PMG.
For each target, we prioritize the best molecule based on their Vina S, QED~\cite{Bickerton2012}, SA~\cite{Ertl2009}, toxicity scores calculated by ICM~\cite{Neves2012} and absorption, distribution, metabolism, excretion, and toxicity (ADMET) metrics calculated by admetSAR 2.0~\cite{Yang2018admetsar}.
%
Figure~\ref{fig:case_example:cdk6} and Figure~\ref{fig:case_example:cdk6:2}  
present the top drug candidates for CDK6, and  Figure~\ref{fig:case_example:nep} presents the top drug candidate for NEP. 
Note that the 3D structures of these molecules and their binding poses are generated by \methodwithsandpguide 
without any post-processing such as energy minimization or docking.
%
All the generated 3D structures are validated to be realistic by their close match, with an RMSD of less than 2\r{A}, to minimized structures from the Cartesian MMFF minimization algorithm~\cite{Halgren1996}.

%
\subsubsection*{Generated Molecules for CDK6}

\begin{figure}[!h]
	\centering
	\begin{minipage}{0.4\linewidth}
		\begin{subfigure}[b]{\linewidth}
			\centering
			\includegraphics[width=0.7\linewidth]{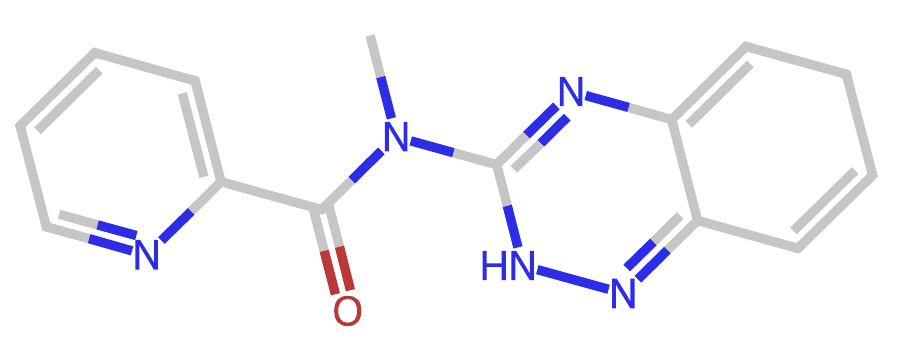}
			\caption{NL-001 for CDK6}
			\label{fig:case_example:cdk6:graph}
		\end{subfigure}
		\begin{subfigure}[b]{\linewidth}
			\centering
			\includegraphics[width=0.7\linewidth]{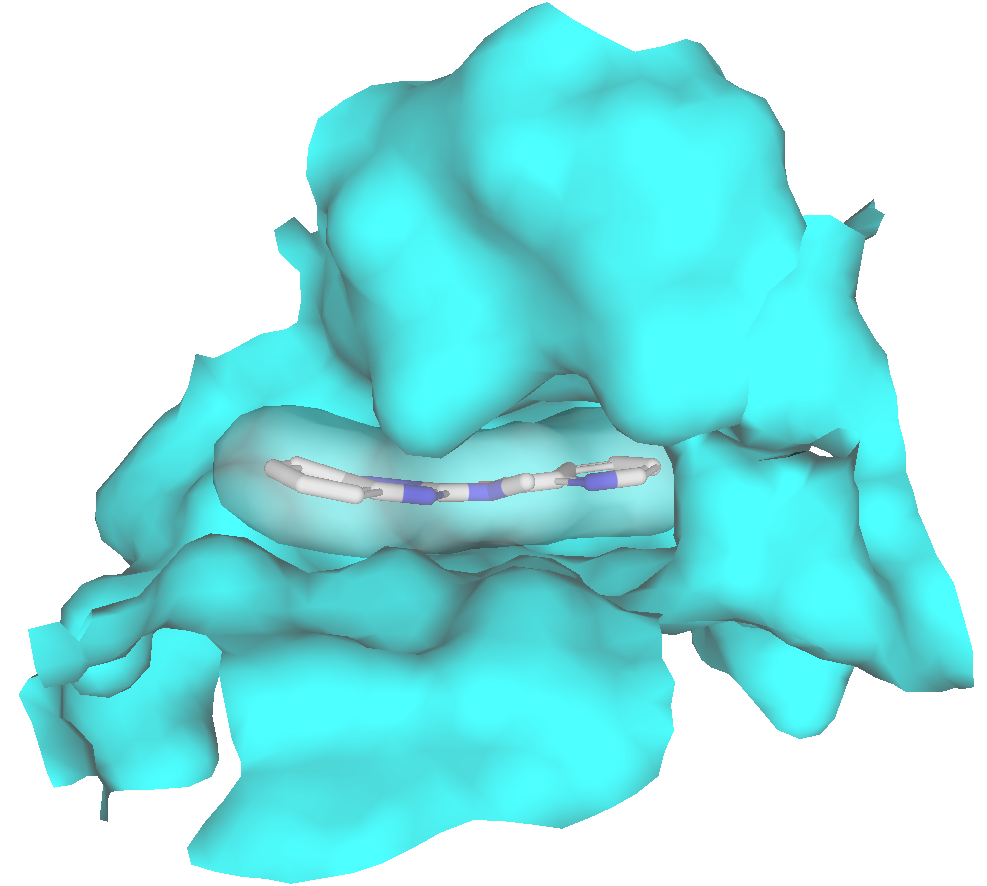}
			\caption{Surface representation of NL-001 and CDK6 interaction}
			\label{fig:case_example:cdk6:2d}
		\end{subfigure}	
	\end{minipage}
	\begin{minipage}{0.58\linewidth}	
		\begin{subfigure}[b]{\linewidth}
			\centering
			\includegraphics[width=1\linewidth]{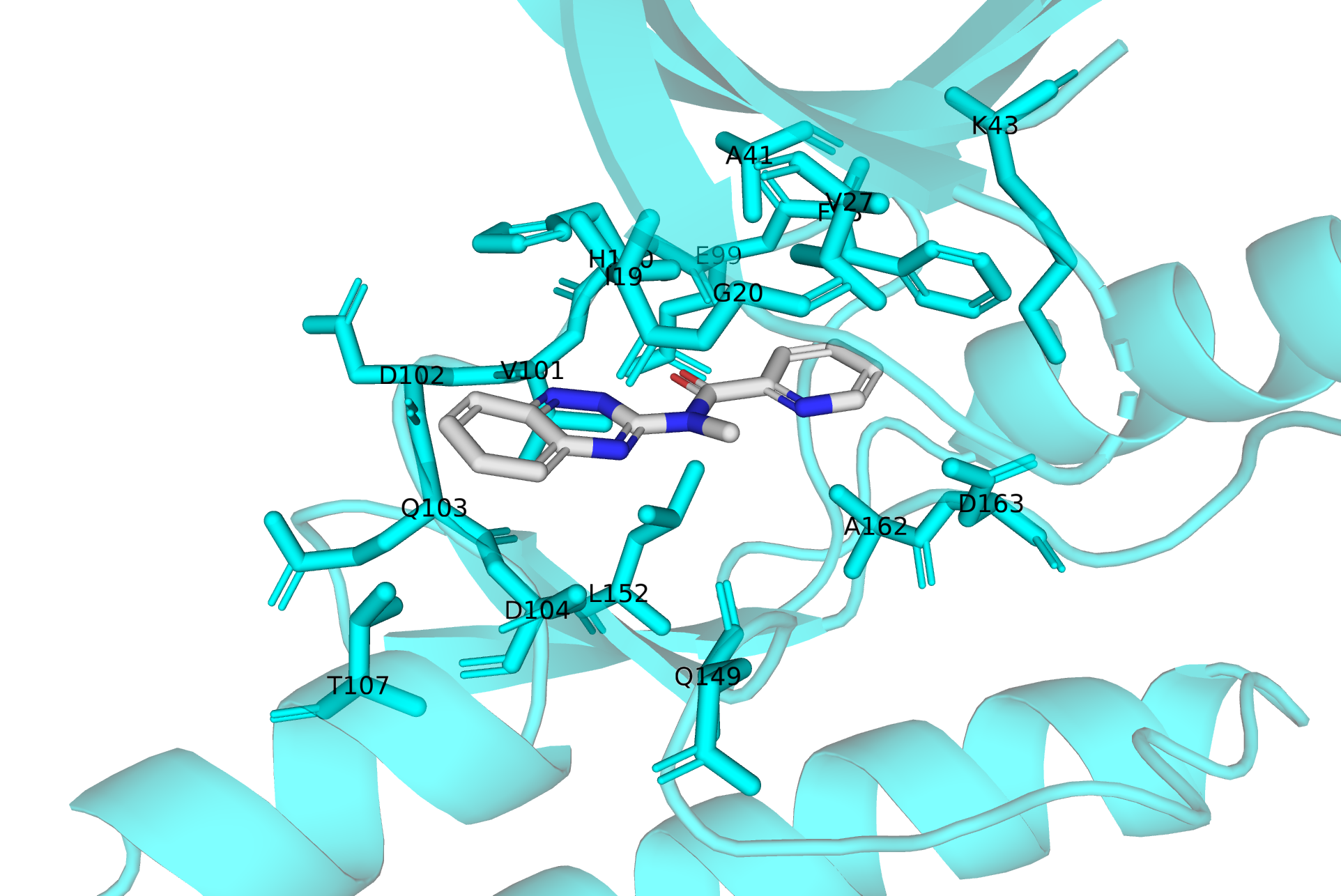}
			\caption{Cartoon representation for NL-001 and CDK6 interaction}
			\label{fig:case_example:cdk6:3d}
		\end{subfigure}
	\end{minipage}	
	\caption{Generated drug candidate NL-001 for CDK6}	
	\label{fig:case_example:cdk6}
\end{figure}	

Figure~\ref{fig:case_example:cdk6} shows a generated molecule,
referred to as NL-001, for CDK6 and its structures and binding 
interactions with the CDK6 binding pocket. Figure~\ref{fig:case_example:cdk6:graph} presents its molecular graph. 
%
%
As shown in Figure~\ref{fig:case_example:cdk6:3d}, molecule NL-001 fits well within the CDK binding pocket and forms hydrophobic interactions with the surrounding residues, such as T107, D104, L152, A162, etc. 
The interactions are further illustrated in Figure~\ref{fig:case_example:cdk6:2d}. 
This effective binding results in a better Vina S of -6.817 kcal/mol for the generated molecule, compared to the Vina S (0.736 kcal/mol) of the 4AU ligand in the complex 4AUA. 
Local minimization and docking refinement can further improve the Vina score of this molecule to -7.251 kcal/mol (Vina M) 
and -8.319 kcal/mol (Vina D), respectively, outperforming the 4AU ligand (-5.939 kcal/mol for Vina M and -7.592 kcal/mol for Vina D).

In addition to the binding activity, the molecule in Figure~\ref{fig:case_example:cdk6} also 
demonstrates favorable properties that are important for drug development, 
including drug-likeness, synthesizability, toxicity, and ADMET profiles.
This molecule meets the Lipinski rule of five criteria~\cite{Lipinski1997}, with a QED score of 0.834, higher than that of 4AU ligand (0.773). 
Its synthetic accessibility (SA) score of 0.720 suggests favorable synthesizability.
This molecule also has a low toxicity score (0.236). 
To fully evaluate its potential as a drug candidate, we compare its ADMET profile with those of three FDA-approved CDK6 inhibitors, including Abemaciclib~\cite{Patnaik2016}, Palbociclib~\cite{Lu2015}, and Ribociclib~\cite{Tripathy2017}. 
The results show that our molecule has comparable or even better ADMET properties in metrics crucial for cancer 
drug development, compared to those approved drugs.
For example, same as the approved drugs, our molecule is predicted to be negative for carcinogenicity~\cite{Benigni2010} and nephrotoxicity~\cite{Soo2018}.
Notably, our molecule has a higher score than all the approved drugs in plasma protein binding, indicating its capacityto be distributed throughout the body and reach the target site.
Details about the properties of the generated molecule NL-001 for CDK6 are available in Supplementary Section~\ref{supp:app:results:properties}.

\begin{figure}[!h]
	\centering
	\begin{minipage}{0.4\linewidth}	
		\begin{subfigure}[b]{\linewidth}
			\centering
			\includegraphics[width=0.7\linewidth]{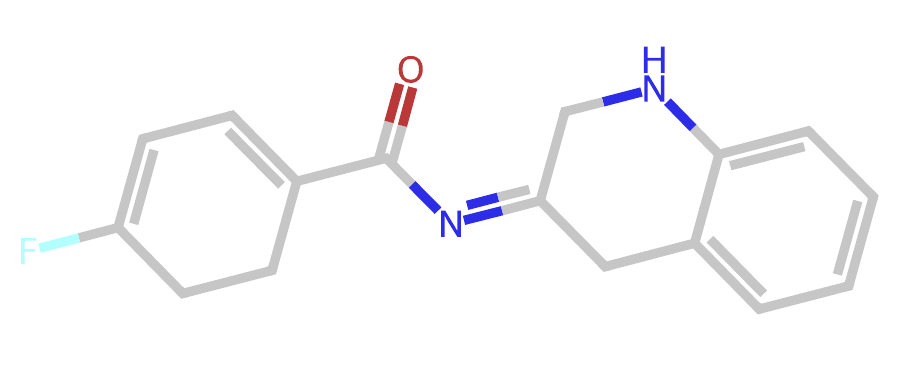}
			\caption{NL-002 for CDK6}
			\label{fig:case_example:cdk6:2:graph}
		\end{subfigure}
		\begin{subfigure}[b]{\linewidth}
			\centering
			\includegraphics[width=0.7\linewidth]{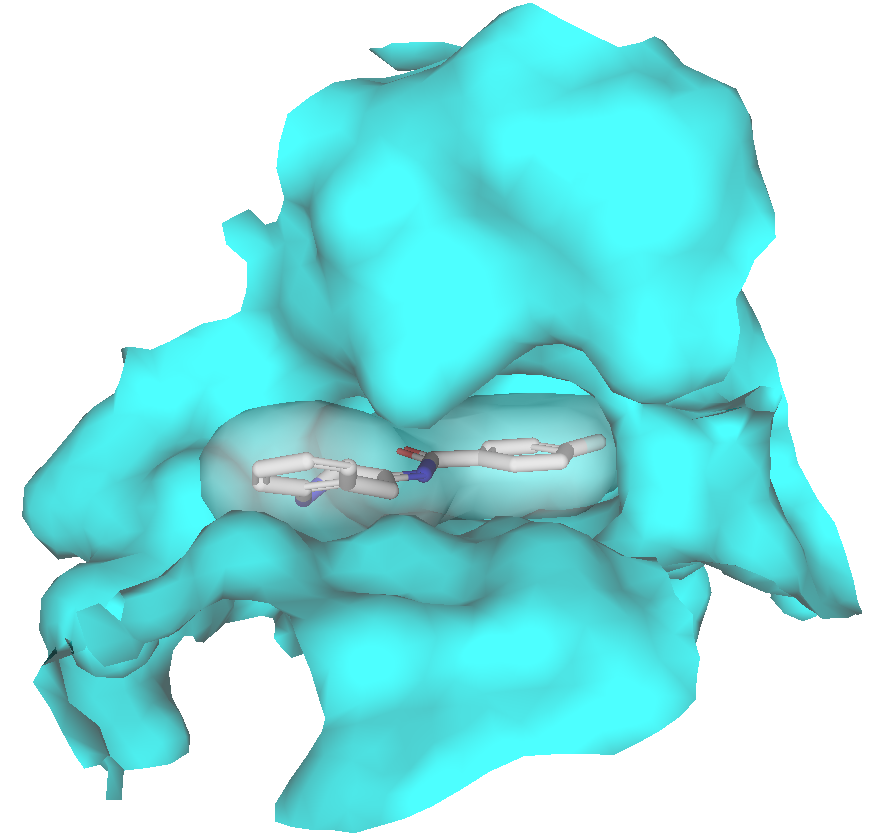}
			\caption{Surface representation of NL-002 and CDK6 interaction}
			\label{fig:case_example:cdk6:2d:2}
		\end{subfigure}
	\end{minipage}
	\begin{minipage}{0.58\linewidth}
	\vspace{10pt}
		\begin{subfigure}[b]{\linewidth}
			\centering
			\includegraphics[width=1\linewidth]{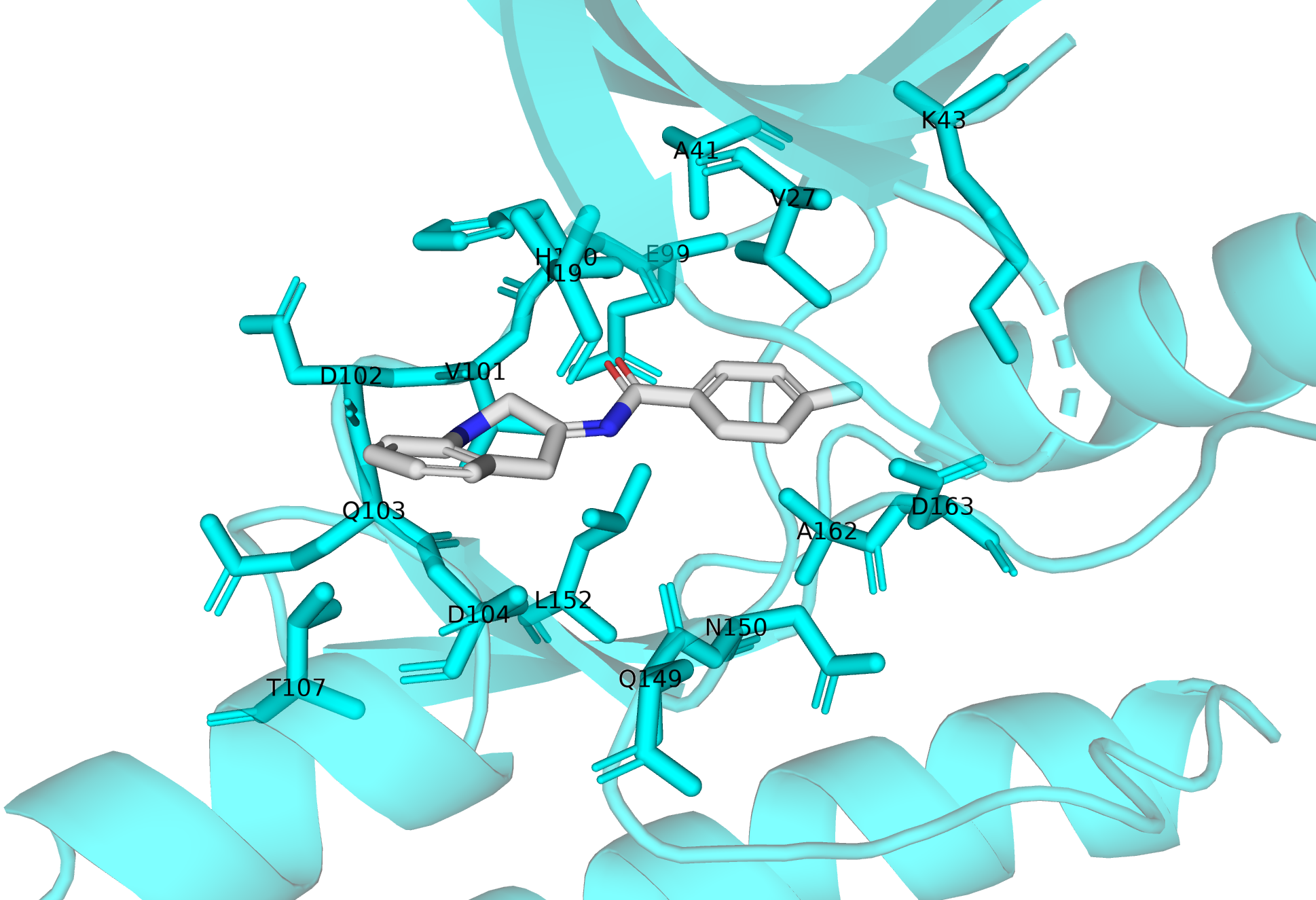}
			\caption{Cartoon representation for NL-002 and CDK6 interaction}
			\label{fig:case_example:cdk6:3d:2}
		\end{subfigure}
	\end{minipage}			
	\caption{Generated drug candidate NL-002 for CDK6}	
	\label{fig:case_example:cdk6:2}
\end{figure}	
Figure~\ref{fig:case_example:cdk6:2} presents another promising drug candidate for CDK6 generated by \method, referred
to as NL-002. It has very similar properties to NL-001, with a Vina S score of -6.970 kcal/mol, a Vina M score of -7.605 kcal/mol, and a Vina D score of -8.986 kcal/mol, 
showing its strong binding affinity to CDK6. 
Notably, NL-002 has a low toxicity score (0.000) and does not have any known toxicity-inducing functional groups detected~\cite{Neves2012}.
NL-002 also has a very similar ADMET profile as NL-001, suggesting it could be another strong drug candidate for CDK6. 
Details about the properties of the generated molecule NL-002 for CDK6 are available in Supplementary Section~\ref{supp:app:results:properties}.

\subsubsection*{Generated Molecule for NEP}

\begin{figure}[!h]	
	\centering
	\begin{minipage}{0.4\linewidth}
		\begin{subfigure}[b]{\linewidth}
			\centering
			\includegraphics[width=0.7\linewidth]{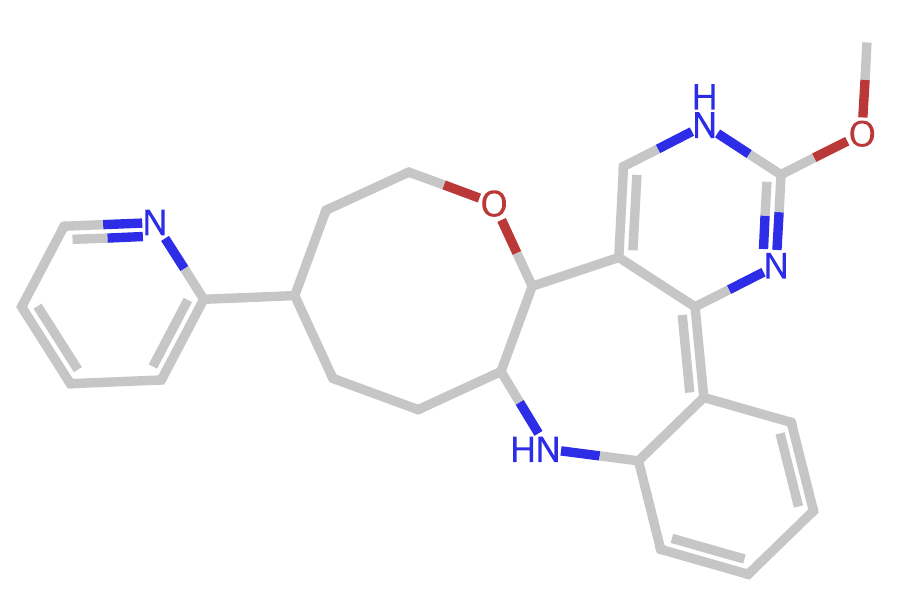}
			\caption{NL-003 for NEP}
			\label{fig:case_example:nep:graph}
		\end{subfigure}
		\begin{subfigure}[b]{\linewidth}
			\centering
			\includegraphics[width=0.7\linewidth]{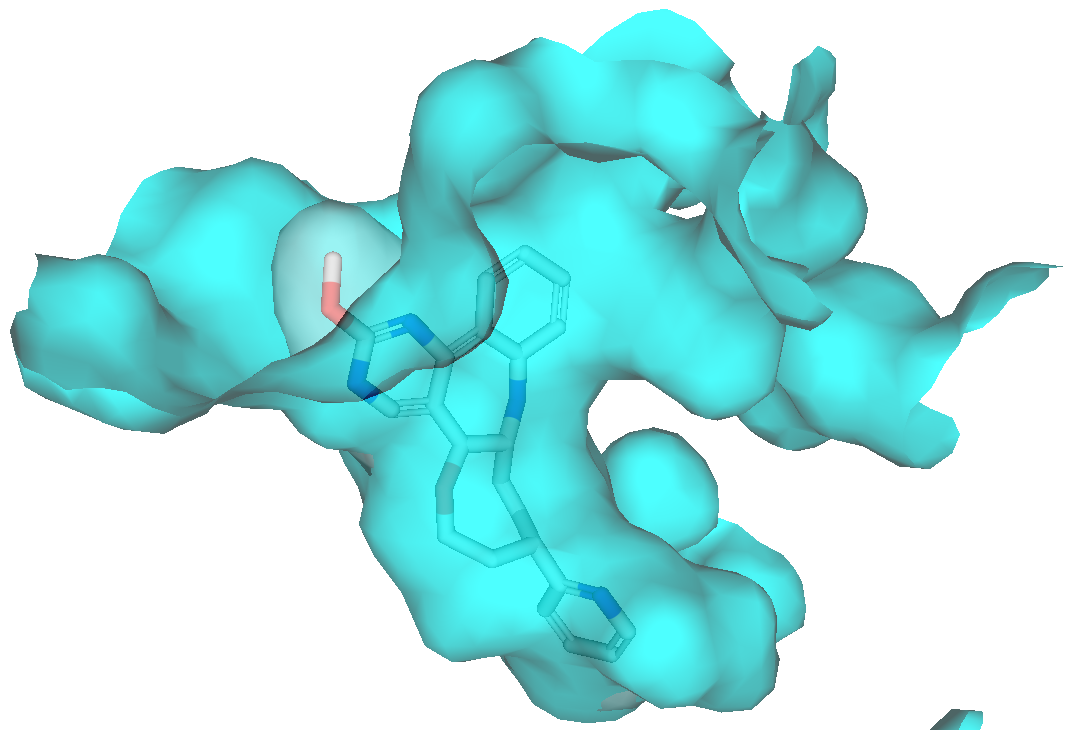}
			\caption{Surface representation for NL-003 and NEP interaction}
			\label{fig:case_example:nep:2d}
		\end{subfigure}
		\label{fig:case_example:nep}
	\end{minipage}
	\begin{minipage}{0.58\linewidth}
	\vspace{30pt}			
		\begin{subfigure}[b]{\linewidth}
			\centering
			\includegraphics[width=1\linewidth]{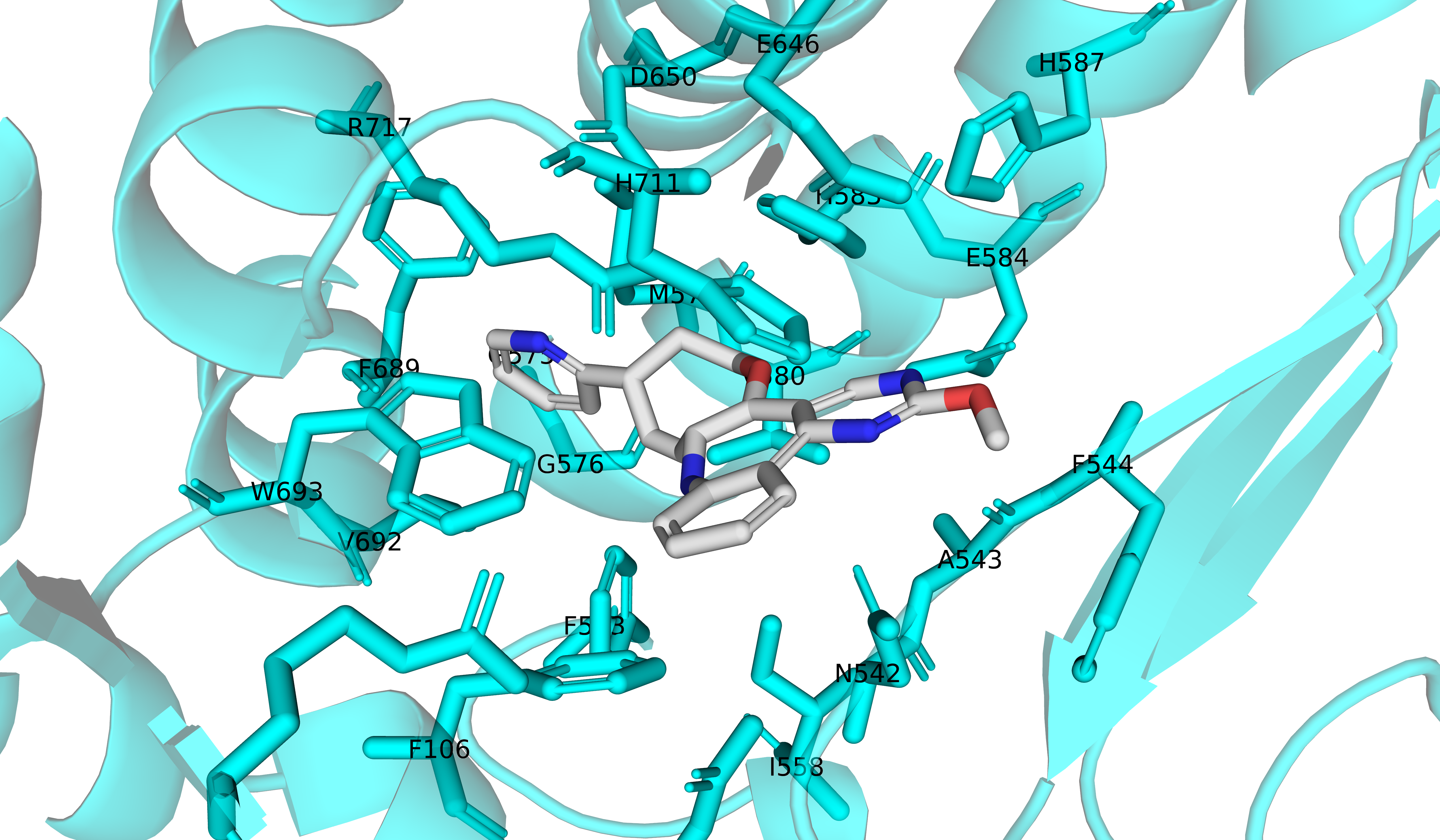}
			\caption{Cartoon representation for NL-003 and NEP interaction}
			\label{fig:case_example:nep:3d}
		\end{subfigure}
	\end{minipage}
	\caption{Generated drug candidate NL-003 for NEP}
	\label{fig:case_example:nep}
\end{figure}

Figure~\ref{fig:case_example:nep} shows the generated molecule,
referred to NL-003, for NEP. Figure~\ref{fig:case_example:nep:graph} presents its molecular graph. 
%
Figure~\ref{fig:case_example:nep:3d} shows how the molecule binds to the NEP ligand binding pocket through hydrogen bonds with residues W693 and E584 and hydrophobic interactions. 
Such interactions are further illustrated in Figure~\ref{fig:case_example:nep:2d}. 
This results in a lower Vina S (-11.953 kcal/mol) of this molecule than that of the BIR ligand in complex 1R1H (-9.399 kcal/mol). 
Through local minimization and docking refinement, this molecule yields lower Vina M (-12.165 kcal/mol) and Vina D (-12.308 kcal/mol) than the Vina M (-9.505 kcal/mol) and Vina D (-9.561 kcal/mol) of BIR ligand.

The molecule NL-003 in Figure~\ref{fig:case_example:nep} also has favorable properties in terms of drug-likeness, 
synthesizability, toxicity, and ADMET profiles.
Particularly, it meets Lipinski's rule of five and achieves a QED score of 0.772, which is substantially higher than that of BIR ligand (0.463).
It also demonstrates a favorable SA score of 0.570 for synthesizability.
%
This molecule is also predicted to be non-toxic and does not have any known toxicity-inducing functional groups detected~\cite{Neves2012}.
It also has a promising ADMET profile comparable to those of three approved drugs, Donepezil, Galantamine, and Rivastigmine, for Alzheimer’s disease~\cite{Hansen2008}, 
specifically in metrics crucial for Alzheimer’s disease drug development.
For example, this molecule is predicted to be permeable to the blood-brain barrier that is essential for treating Alzheimer's disease~\cite{Deane2007} and negative for carcinogenicity~\cite{Benigni2010}, same as the approved drugs.
Details about the properties of the generated molecule for NEP are available in Supplementary Section~\ref{supp:app:results:properties}.


\section*{Discussions and Conclusions}

\subsection*{Integrating Protein Targets for Binding Molecule Generation}

For PMG, our experiments show that \method with pocket guidance (\methodwithpguide and \methodwithsandpguide) can effectively generate molecules with high binding affinities toward protein targets.
As detailed in Section ``\method with Protein Pocket Guidance'', this pocket guidance enables \methodwithpguide and \methodwithsandpguide to consider the geometric information of protein binding pockets when generating binding molecules.
%
%
In addition to geometric information, we acknowledge that incorporating other information about protein pockets could further 
enable molecules generated in high quality. 
For example, the physicochemical properties of amino acid sequences within the binding pockets, 
such as polarity, electrostatics, and hydrophobicity, can affect the strength of interactions between 
proteins and molecules~\cite{Du2016}. 
Therefore, a generative model
considering these properties could produce molecules that better conform to what is expected based on pharmaceutical chemistry, 
shortening their pathways to be induced into downstream tasks of drug development. 
Identifying and integrating essential properties of protein binding pockets into molecule generation 
could be an interesting yet challenging future research direction.
%

\subsection*{Multi-Objective Molecule Generation}

When developing a molecule into a drug, in addition to its binding affinity to the protein target, many 
other properties also need to be considered, such as drug-likeness, synthesizability, toxicity, metabolism, and cell permeability~\cite{Lin2003}.
Similar to other molecule generation methods~\cite{luo2021sbdd,peng22pocket2mol,schneuing2022structure,guan2023targetdiff}, \method primarily emphasizes molecule binding affinities.
Although the case study in Section ``Case Studies for Targets'' indicates that its generated binding molecules may have favorable ADMET profiles, properties beyond binding affinities and shapes are not specifically optimized by design 
in the molecules out of \method.  
Consequently, the generated molecules may need to go through further optimization and refinement to gain other 
necessary properties in order to become viable drug candidates.  
Towards this end, a multi-objective genAI model that generates molecules exhibiting  
multiple properties simultaneously and satisfying multiple objectives (e.g., high drug-likeness, high synthesizability) 
could be greatly demanded, which calls for a significant future research endeavor, though out of the scope of this study. 

\subsection*{\emph{In vitro} Validation}

\emph{In vitro} experimental validation is indispensable for accessing \emph{in silico} generated molecules 
for further investigation into real-world therapeutic agents. 
Even when all the desired properties could be ideally incorporated into the generative process, which, by itself, is 
highly nontrivial, these properties of the 
generated molecules remain unclear until they are experimentally confirmed. 
Meanwhile, other unanticipated properties may emerge due to the 
unknown interactions between the molecules and the complex biological systems, 
which also requires rigorous \emph{in vitro} testing. 
%
%
Despite its crucial importance, systematic \emph{in vitro} validation for genAI generated molecules remains very challenging~\cite{Chen2023}. 
%
%
This process would start from effective sampling or prioritization of generated molecules to identify a feasible and manageable subset for \emph{in vitro} experiments.
Then, determining and executing viable synthesis reactions to make those molecules, if they do not exist, which is highly likely, 
also pose substantial difficulties~\cite{Gao2020}. 
%
Given the focus of this manuscript on developing \emph{in silico} genAI methods, \emph{in vitro} validation is beyond the scope
but remains a pivotal next step to investigate.
\subsection*{Conclusions}

\method generates novel binding molecules with realistic 3D structures based on the shapes of known ligands. 
It utilizes pre-trained shape embeddings and a customized diffusion model for binding molecule generation.
To better resemble the known ligand shapes, \method also modifies the generated 3D molecules iteratively under the guidance of the ligand shapes. 
Additionally, it can leverage the geometric information of protein binding pockets and tailor the generated molecules toward optimal binding affinities.
Experimental results demonstrate that \method outperforms SMG methods in generating molecules with highly similar shapes to known ligands, while incorporating shape guidance further boosts this performance.
When compared to PMG methods, \method with pocket guidance also achieves exceptional performance in generating molecules with high binding affinities.
The case studies involving two critical drug targets show that \method can generate binding molecules with desirable drug properties.
However, \method still has limitations.
In addition to the limitations and corresponding future research directions that have been discussed above, 
one limitation with {\method} is that the binding poses of generated molecules are typically constrained by those of known ligands.
This limitation can confine the ability of \method to explore novel binding poses.
Thus, a future research direction is on how to mitigate this limitation by inferring diverse ligand shapes from protein pockets.
\section*{Method}


%
\method aims to generate novel binding molecules based on the shapes of known ligands, following the principle that molecules with similar shapes tend to have similar binding activities.
%
%
%
Toward this end, 
\method consists of two modules: (1) a pre-trained equivariant shape embedding module {\SE} that learns expressive latent embeddings for the shapes of condition molecules (e.g, ligands), and 
(2) an equivariant molecule diffusion model {\methoddiff} that explicitly considers shape embeddings from \SE to generate new 3D molecules with similar shapes to condition molecules.  
%
Particularly, given a condition molecule, {\SE} represents its shape as a point cloud with points sampled over its molecular surface. 
{\SE} learns to map this point cloud into a latent embedding {\shapehiddenmat} using an encoder-decoder framework
(more details in Section ``Equivariant Condition Shape Representation Pre-training'').
Conditioned on the shape embedding \shapehiddenmat, \methoddiff  
learns to generate
molecules with desired shapes and realistic topologies
in an equivariant way. 
%
%
%
Particularly, \methoddiff utilizes equivariant graph neural networks to learn shape-aware atom embeddings and generate molecules tailored to the shape condition.
%
\methoddiff also leverages bond types as a training signal to fully capture the inherent topologies of molecules. 
%
%
%
%
%
%
During inference, 
%
\methoddiff utilizes shape guidance to further direct the generated molecules toward the shape condition.
%
Besides shape guidance, when the structure of the protein binding pocket is available, \methoddiff employs pocket guidance to adjust the atom positions of generated molecules for optimal binding affinities with the binding pocket.  
%
%
This design enables the applicability of \method for PMG. 
%
%
%
%
%
%
%
Fig.~\ref{fig:overall} presents the overall architecture of \method. 
All the algorithms are presented in Supplementary Section~\ref{supp:algorithms}.

{In the following sections, we will first introduce the key notations and the definitions of equivariance and invariance in Section ``Representations, Notations, and Preliminaries.''
We will then introduce the equivariant shape embedding module \SE in Section ``Equivariant Condition Shape Representation Pre-training. '' 
After that, we will discuss the shape-conditioned molecule diffusion model \methoddiff in Section ``Diffusion-based Molecule Generation.''
We will describe the shape-conditioned molecule prediction module \molpred used in \methoddiff in Section ``Shape-conditioned Molecule Prediction.''
Finally, we will describe the molecule generation process and how the shape guidance and pocket guidance are used during inference in Section ``Guidance-induced Inference.''
}

\begin{figure}
	\centering
	\includegraphics[width=.9\linewidth]{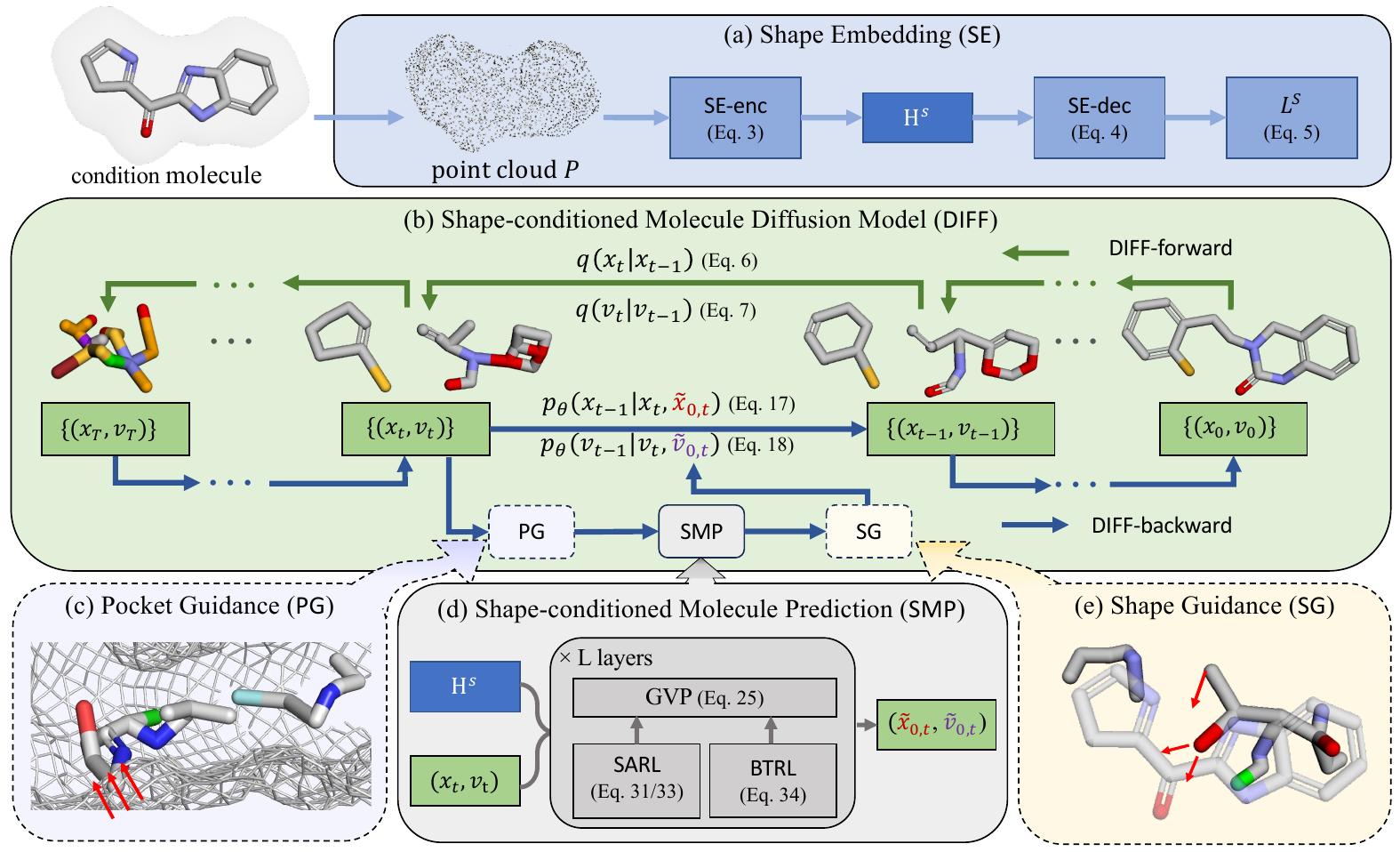}
	\caption{\textbf{Model Architecture of \method.} 
		\textbf{a,} Shape embedding module \SE. \method uses a shape embedding module \SE to map the 3D molecule surface shapes \pc into shape embeddings $\shapehiddenmat$. 
		\SE uses an encoder \SEE to map \pc into \shapehiddenmat, and a decoder \SED to optimize \shapehiddenmat with loss $\mathcal{L}^s$.
		\textbf{b,} Shape-conditioned molecule diffusion model \methoddiff. 
		\method uses \methoddiff to generate molecules conditioned on \shapehiddenmat. 
		\methoddiff includes a forward diffusion process, denoted as \diffnoise, which gradually adds noises step by step to the atom positions and features $\{(\pos_t, \atomfeat_t)\}$ at step $t$. 
		\methoddiff uses a backward generative process, denoted as \diffgenerative, to remove the noises in the noisy molecules. 
		\methoddiff generates a 3D molecule by first sampling noisy atom positions and features $\{(\pos_T, \atomfeat_T)\}$ at step $T$ and then removing the noises step by step until $t$ reaches 1.
		\textbf{c,} Pocket guidance \pocketguide. During the generation, \method can use \pocketguide to adjust atom positions $\atomfeat_t$ for minimizing steric clashes between generated molecules and protein pockets.
		\textbf{d,} Shape-conditioned molecule prediction module \molpred. 
		\methoddiff uses \molpred to predict the atom positions and features $(\tilde{\pos}_{0,t}, \tilde{\atomfeat}_{0,t})$ given the noisy data $(\pos_t, \atomfeat_t)$ and \shapehiddenmat.
		\molpred is a multi-layer graph neural network comprising $L$ layers.
		In the $l$-th layer, \molpred leverages a shape-aware atom representation learning  (SARL) module, a bond-type representation learning (BTRL) module, and a geometric vector perceptron (GVP) to jointly learn effective atom representations for the prediction.
		\textbf{e,} Shape guidance \shapeguide. During the generation, \method can use \shapeguide to explicitly push predicted atoms to the shapes of condition molecules.
		}
	\label{fig:overall}
\end{figure}

\subsection*{Representations, Notations, and Preliminaries}

\subsubsection{Representations and Notations}


We represent a molecule \mol
as a set of atoms \mbox{$\mol = \{a_1, a_2, \cdots, a_{\scriptsize{|\mol|}}| a_i = (\pos_i, \atomfeat_i)\}$},
where $|\mol|$ is the number of atoms in \mol; $a_i$ is the $i$-th atom in \mol;  
$\pos_i\in \mathbb{R}^3$ represents the position of $a_i$ in 3D space; 
and $\atomfeat_i\in \mathbb{R}^{K}$ is $a_i$'s one-hot atom feature vector indicating the atom type and its
aromaticity.
We represent the Euclidean distance between each pair of atoms $a_i$ and $a_j$ as $d_{ij} \in \mathbb{R}$, and 
the type of the bond in between 
as a one-hot vector $\bondemb_{ij} \in \mathbb{R}^4$, 
in which the four dimensions of $\bondemb_{ij}$ represent the absence of a bond, a single bond, a double bond, and an aromatic bond, respectively. 
%
%
%
%
%
Following Guan \etal~\cite{guan2023targetdiff}, bonds between atoms can be uniquely determined by the atom types and the atomic distances among atoms. 
%
%
We represent the 3D surface shape {\shape} of a molecule \mol as a point cloud constructed by sampling points over the molecular surface.
Details about the construction of point clouds from the surface of molecules are available in Supplementary Section~\ref{supp:point_clouds}.
We denote the point cloud as $\pc=\{z_1, z_2, \cdots\, z_{\scriptsize{|\pc|}} | z_j = (\pointpos_j)\}$, 
where $|\pc|$ is the number of points in $\pc$;
$z_j$ is the $j$-th point; and 
$\pointpos_j\in \mathbb{R}^3$ represents the position of $z_j$ in 3D space.
We denote the latent embedding of $\pc$ as $\shapehiddenmat\in \mathbb{R}^{d_p\times 3}$, where $d_p$ is the dimension of the latent embedding. 
We represent the distance of a point randomly sampled in 3D space to the molecule surface as $o$, referred to as a signed distance, with a positive (negative) sign indicating the point is inside (outside) the surface. 
Table~\ref{tbl:notations} summarizes the notations used in this manuscript.

\input{tables/notations}



\subsubsection{Equivariance and Invariance}

\paragraph{Equivariance}

{Equivariance refers to the property of a function $f(\pos)$ 
that any translation and rotation transformation from the special Euclidean group SE(3)~\cite{Atz2021} applied to a geometric object
$\pos\in\mathbb{R}^3$ is mirrored in the output of $f(\pos)$, accordingly.
This property ensures $f(\pos)$ to learn a consistent representation of an object's geometric information, regardless of its orientation or location in 3D space.
%
%
Formally, given any translation transformation $\mathbf{t}\in\mathbb{R}^3$ and rotation transformation $\mathbf{R}\in\mathbb{R}^{3\times3}$ ($\mathbf{R}^{\mathsf{T}}\mathbf{R}=\mathbb{I}$, 
$f(\pos)$ is equivariant with respect to these transformations 
if it satisfies
\begin{equation}
f(\mathbf{R}\pos+\mathbf{t}) = \mathbf{R}f(\pos) + \mathbf{t}. 
\end{equation}
%
%
In \method, both \SE and \methoddiff are developed to guarantee equivariance in capturing the geometric features of objects regardless of any translation or rotation transformations, as will be detailed in the following sections.
}

\paragraph{Invariance}

Invariance refers to the property of a function that its output {$f(\pos)$} remains constant under any translation and rotation transformations of the input $\pos$. 
This property enables $f(\pos)$ to accurately capture 
the inherent features (e.g., atom features for 3D molecules) that are invariant of its orientation or position in 3D space.
Formally, $f(\pos)$ is invariant under any translation $\mathbf{t}$ and  rotation $\mathbf{R}$ if it satisfies
\begin{equation}
f(\mathbf{R}\pos+\mathbf{t}) = f(\pos).
\end{equation}
In \method, both \SE and \methoddiff capture the inherent features of objects in an invariant way, regardless of any translation or rotation transformations, as will be detailed in the following sections.

\subsection*{Equivariant Condition Shape Representation Pre-training (\SE)}
%
\method pre-trains a shape embedding module \SE to generate surface shape embeddings 
$\shapehiddenmat$ of condition molecules. 
\SE uses an encoder \SEE to map $\pc$ to the equivariant latent embedding $\shapehiddenmat$.
\SE employs a decoder \SED to optimize $\shapehiddenmat$ by recovering the signed distances~\cite{park2019sdf} of randomly sampled points 
in 3D space to the molecule surface using 
$\shapehiddenmat$. 
%
\method uses $\hiddenmat^{\mathtt{s}}$ to guide the diffusion process {as will be detailed} later
(Section ``Diffusion-based Molecule Generation'').
We present \SE in detail in the following sections. Particularly, we present the encoder \SEE in Section ``Shape Encoder''; the decoder \SED in Section ``Shape Decoder''; and the optimization of \SE in Section ``\SE Pre-training.'' 
{Fig.~\ref{fig:overall}(a) presents the architecture of \SE. }
\subsubsection{Shape Encoder (\SEE)}

\SEE learns shape embeddings $\hiddenmat^{\mathtt{s}}$ from the 3D surface shape $\pc$ of molecules in an equivariant way, 
%
as described in Section ``Equivariance and Invariance''.
%
%
%
To ensure translation equivariance, 
{\SEE} shifts the center of each $\pc$ to zero to eliminate all translations.
To ensure rotation equivariance, \SEE leverages vector neurons (VNs)~\cite{deng2021vn} and dynamic graph convolutional neural networks (DGCNNs)~\cite{wang2019dynamic} to learn shape embeddings $\shapehiddenmat$ as follows:
\begin{equation*}
\{\hiddenmat^{\mathtt{p}}_1, \hiddenmat^{\mathtt{p}}_2, \cdots, \hiddenmat^{\mathtt{p}}_{|\scriptsize{\pc}|}\} = \text{VN-DGCNN}(\{\pointpos_1, \pointpos_2, \cdots, \pointpos_{|\scriptsize{\pc}|}\}), 
\vspace{-3pt}
\end{equation*}
\begin{equation}
\hiddenmat^{\mathtt{s}} = \sum\nolimits_{j}\hiddenmat^{\mathtt{p}}_j / {|\pc|},
\label{eqn:shape_embed}
\end{equation}
%
%
where $\text{VN-DGCNN}(\cdot)$ is a VN-based DGCNN network
to generate equivariant embedding $\hiddenmat^{\mathtt{p}}_j\in  \mathbb{R}^{3\times d_p}$ for each point $z_j$ in $\pc$; 
and $\hiddenmat^{\mathtt{s}} \in \mathbb{R}^{3\times d_p}$ is the embedding of $\pc$ generated via %
a mean-pooling over the embeddings of all the points.
%
%
$\text{VN-DGCNN}(\cdot)$ guarantees the rotation equivariance by learning embedding matrices $\hiddenmat^{\mathtt{p}}_j\in  \mathbb{R}^{3\times d_p}$ for points using only equivariant operations as detailed in Deng \etal~\cite{deng2021vn}
%
%

%
%
%
%


\subsubsection{Shape Decoder (\SED)}

To optimize $\hiddenmat^{\mathtt{s}}$, following Deng \etal~\cite{deng2021vn}, \SE learns a decoder \SED to predict the signed distance of a query point $z_q$ randomly sampled from  3D space to surface shape \shape using multilayer perceptrons (MLPs) as follows:
\begin{equation}
\tilde{o}_q = \text{MLP}([\langle \mathbf{z}_q, \hiddenmat^{\mathtt{s}}\rangle, \|\mathbf{z}_q\|^2, \text{VN-In}(\hiddenmat^{\mathtt{s}})]),
\label{eqn:se:decoder}
\end{equation}
where $\tilde{o}_q$ is the predicted signed distance of $z_q$, with positive and negative values indicating $z_q$ is inside or outside the surface shape \shape, respectively;
$[\cdot,\cdot]$ represents the concatenation operation;
$\langle\cdot,\cdot\rangle$ is the dot-product operator;
$\|\mathbf{z}_q\|^2$ is the squared Euclidean norm of the position of $z_q$;
$\text{VN-In}(\cdot)$ is an invariant 
VN network~\cite{deng2021vn} that converts the equivariant shape embedding $\shapehiddenmat\in \mathbb{R}^{d_p\times 3}$ into 
an invariant shape embedding $\text{VN-In}(\shapehiddenmat)\in \mathbb{R}^{d_p}$. 
Intuitively, \SED predicts the signed distance between the query point and 3D surface by jointly considering 
the interaction between the point and surface ($\langle\mathbf{z}_q, \hiddenmat^{\mathtt{s}}\rangle$), 
the distance 
of the query point ($\|\mathbf{z}_q\|^2=\langle\mathbf{z}_q, \mathbf{z}_q \rangle$) to the origin, 
and the molecule surface shape ($\text{VN-In}(\cdot)$).
All these three terms 
are invariant to any rotation transformations, as they are calculated from the dot-product operation $\langle \cdot, \cdot \rangle$. 
This operation is invariant to any rotations {as} $\langle \mathbf{R}\mathbf{z}, \mathbf{R}\mathbf{z} \rangle = \mathbf{z}^{\mathsf{T}}\mathbf{R}^{\mathsf{T}}\mathbf{R}\mathbf{z} = \mathbf{z}^{\mathsf{T}}\mathbf{z} = \langle \mathbf{z}, \mathbf{z} \rangle$.
Note that $\text{VN-In}(\cdot)$ comprises invariant dot-product operations and specifically designed invariant activations to learn invariant embeddings, as detailed in Deng~\etal~\cite{deng2021vn}. 
%
%
%
The predicted signed distance $\tilde{o}_q$ is used to calculate the loss for the optimization of $\shapehiddenmat$ (discussed below in Equation~\ref{eqn:loss_distance}).
%
%
%
We present the sampling process of $z_q$ in the Supplementary Section~{\ref{supp:training:shapeemb}}.

\subsubsection{\SE Pre-training}

\method pre-trains \SE by minimizing the squared-errors loss between the predicted and the ground-truth signed distances of query points to the surface shape \shape as follows: 
\begin{equation}
\label{eqn:loss_distance}
\mathcal{L}^{\mathtt{s}} = \sum\nolimits_{z_q \in \mathcal{Z}} \|o_q-\tilde{o}_q\|^2,
\end{equation}
where $\mathcal{Z}$ is the set of sampled query points and $o_q$ is the ground-truth
signed distance of query point $z_q$.
By pretraining \SE, \method learns $\shapehiddenmat$ that will be used as the condition in the following 3D molecule generation.

%

\subsection*{Diffusion-based Molecule Generation (\methoddiff)}
\label{section:diff}

In \method, a shape-conditioned molecule diffusion model, referred to as \methoddiff, 
is used to generate a 3D molecule structure (i.e., atom coordinates and features, and bonds) 
conditioned on a given 3D surface shape 
that is represented by the shape latent embedding $\shapehiddenmat$ (Equation~\ref{eqn:shape_embed}).
{Fig.~\ref{fig:overall}(b) presents the architecture of \methoddiff. }
%
%
Following the denoising diffusion probabilistic models~\cite{ho2020ddpm}, \methoddiff includes a forward diffusion process based on a Markov chain,
denoted as \diffnoise, which gradually adds noises step by step to the atom positions and features $\{(\pos_i, \atomfeat_i)\}$ in the training molecules with $i$ indexing the $i$-th atom.
%
The noisy atom positions and features at step $t$ are represented as $\{(\pos_{i,t}, \atomfeat_{i,t})\}$ ($t=1, \cdots, T$),
and the molecules without any noise 
are represented as $\{(\pos_{i,0}, \atomfeat_{i,0})\}$.
At the final step $T$, $\{(\pos_{i,T}, \atomfeat_{i,T})\}$ are completely unstructured and resemble 
a simple distribution like a Normal distribution $\mathcal{N}(\mathbf{0}, \mathbb{I})$ or a uniform categorical distribution {$\mathcal{C}(\mathbf{1}/K)$},
in which $\mathbf{I}$ and $\mathbf{1}$ denotes the identity matrix and identity vector, respectively.
When no ambiguity arises, we will eliminate subscript $i$ in the notations and use $(\pos_{t}, \atomfeat_{t})$ for brevity.

During training, \methoddiff is learned to reverse the forward diffusion process via another Markov chain, 
referred to as the backward generative process and denoted as \diffgenerative, to remove the noises in the noisy molecules.
During inference, 
\methoddiff first samples noisy atom positions and features at step $T$ 
from simple distributions and then generates a 3D molecule structure by removing the noises in the noisy molecules step by step until $t$ reaches 1.


\subsubsection{Forward Diffusion Process (\diffnoise)}
\label{section:diff:diff}


Following the previous work~\cite{guan2023targetdiff}, at step $t\in[1, T]$, a small Gaussian noise and a small categorical noise are added to the continuous atom positions and 
discrete atom features $\{(\pos_{t-1}, \atomfeat_{t-1})\}$, 
respectively.
%
%
The noise levels of the Gaussian and categorical noises are determined by two predefined variance schedules $(\beta_t^{\mathtt{x}}, \beta_t^{\mathtt{v}})\in (0,1)$, where $\beta_t^{\mathtt{x}}$ and $\beta_t^{\mathtt{v}}$
are selected to be sufficiently small to ensure the smoothness of \diffnoise.
The details about variance schedules are available in Supplementary Section~\ref{supp:forward:variance}. 
%
%
Formally, for atom positions, the probability of $\pos_t$ sampled given $\pos_{t-1}$, denoted as $q(\pos_t|\pos_{t-1})$, is defined as follows,
%
\begin{equation}
q(\pos_t|\pos_{t-1}) = \mathcal{N}(\pos_t|\sqrt{1-\beta^{\mathtt{x}}_t}\pos_{t-1}, \beta^{\mathtt{x}}_t\mathbb{I}), 
\label{eqn:noiseposinter}
\end{equation}
%
where 
$\mathcal{N}(\cdot)$ is a Gaussian distribution of $\pos_t$ with mean $\sqrt{1-\beta_t^{\mathtt{x}}}\pos_{t-1}$ and covariance $\beta_t^{\mathtt{x}}\mathbf{I}$.
%
Following Hoogeboom \etal~\cite{hoogeboom2021catdiff}, 
%
for atom features, the probability of $\atomfeat_t$ across $K$ classes given $\atomfeat_{t-1}$ is defined as follows,
\begin{equation}
q(\atomfeat_t|\atomfeat_{t-1}) = \mathcal{C}(\atomfeat_t|(1-\beta^{\mathtt{v}}_t) \atomfeat_{t-1}+\beta^{\mathtt{v}}_t\mathbf{1}/K),
\label{eqn:noisetypeinter}
\end{equation}
where 
$\mathcal{C}$ is a categorical distribution of $\atomfeat_t$ derived from the 
noising $\atomfeat_{t-1}$ with a uniform noise $\beta^{\mathtt{v}}_t\mathbf{1}/K$ across $K$ classes.

Since the above distributions form Markov chains, 
the probability of any $\pos_t$ or $\atomfeat_t$ can be derived from $\pos_0$ or $\atomfeat_0$:
%
\begin{eqnarray}
& q(\pos_t|\pos_{0}) & = \mathcal{N}(\pos_t|\sqrt{\cumalpha^{\mathtt{x}}_t}\pos_0, (1-\cumalpha^{\mathtt{x}}_t)\mathbb{I}), \label{eqn:noisepos}\\
& q(\atomfeat_t|\atomfeat_0)  & = \mathcal{C}(\atomfeat_t|\cumalpha^{\mathtt{v}}_t\atomfeat_0 + (1-\cumalpha^{\mathtt{v}}_t)\mathbf{1}/K), \label{eqn:noisetype}\\
& \text{where }\cumalpha^{\mathtt{u}}_t & = \prod\nolimits_{\tau=1}^{t}\alpha^{\mathtt{u}}_\tau, \ \alpha^{\mathtt{u}}_\tau=1 - \beta^{\mathtt{u}}_\tau, \ {\mathtt{u}}={\mathtt{x}} \text{ or } {\mathtt{v}}.\;\;\;\label{eqn:noiseschedule}
\label{eqn:pos_prior}
\end{eqnarray}
%
%
Note that $\bar{\alpha}^{\mathtt{u}}_t$ ($\mathtt{u}={\mathtt{x}}\text{ or }{\mathtt{v}}$)
is monotonically decreasing from 1 to 0 over $t=[1,T]$. 
As $t\rightarrow 1$, $\cumalpha^{\mathtt{x}}_t$ and $\cumalpha^{\mathtt{v}}_t$ are close to 1, leading to that $\pos_t$ or $\atomfeat_t$ approximates 
$\pos_0$ or $\atomfeat_0$.
Conversely, as  $t\rightarrow T$, $\cumalpha^{\mathtt{x}}_t$ and $\cumalpha^{\mathtt{v}}_t$ are close to 0,
leading to that $q(\pos_T|\pos_{0})$ 
resembles  {$\mathcal{N}(\mathbf{0}, \mathbb{I})$} 
and $q(\atomfeat_T|\atomfeat_0)$ 
resembles {$\mathcal{C}(\mathbf{1}/K)$}.

Using Bayes theorem, the ground-truth Normal posterior of atom positions $p(\pos_{t-1}|\pos_t, \pos_0)$ can be calculated in a
closed form~\cite{ho2020ddpm} as below,
\begin{eqnarray}
& p(\pos_{t-1}|\pos_t, \pos_0) = \mathcal{N}(\pos_{t-1}|\mu(\pos_t, \pos_0), \tilde{\beta}^\mathtt{x}_t\mathbb{I}), \label{eqn:gt_pos_posterior_1}\\
&\!\!\!\!\!\!\!\!\!\!\!\mu(\pos_t, \pos_0)\!=\!\frac{\sqrt{\bar{\alpha}^{\mathtt{x}}_{t-1}}\beta^{\mathtt{x}}_t}{1-\bar{\alpha}^{\mathtt{x}}_t}\pos_0\!+\!\frac{\sqrt{\alpha^{\mathtt{x}}_t}(1-\bar{\alpha}^{\mathtt{x}}_{t-1})}{1-\bar{\alpha}^{\mathtt{x}}_t}\pos_t, 
\tilde{\beta}^\mathtt{x}_t\!=\!\frac{1-\bar{\alpha}^{\mathtt{x}}_{t-1}}{1-\bar{\alpha}^{\mathtt{x}}_{t}}\beta^{\mathtt{x}}_t.\;\;\;
\end{eqnarray}
%
Similarly, the ground-truth categorical posterior of atom features $p(\atomfeat_{t-1}|\atomfeat_{t}, \atomfeat_0)$ can be calculated~\cite{hoogeboom2021catdiff} as below,
\begin{eqnarray}
& p(\atomfeat_{t-1}|\atomfeat_{t}, \atomfeat_0) = \mathcal{C}(\atomfeat_{t-1}|\mathbf{c}(\atomfeat_t, \atomfeat_0)), \label{eqn:gt_atomfeat_posterior_1}\\
& \mathbf{c}(\atomfeat_t, \atomfeat_0) = \tilde{\mathbf{c}}/{\sum_{k=1}^K \tilde{c}_k}, \label{eqn:gt_atomfeat_posterior_2} \\
& \tilde{\mathbf{c}} = [\alpha^{\mathtt{v}}_t\atomfeat_t + \frac{1 - \alpha^{\mathtt{v}}_t}{K}]\odot[\bar{\alpha}^{\mathtt{v}}_{t-1}\atomfeat_{0}+\frac{1-\bar{\alpha}^{\mathtt{v}}_{t-1}}{K}], 
\label{eqn:gt_atomfeat_posterior_3}
\end{eqnarray}
%
%
where $\tilde{c}_k$ denotes the likelihood of $k$-th class across $K$ classes in $\tilde{\mathbf{c}}$; 
$\odot$ denotes the element-wise product operation;
$\tilde{\mathbf{c}}$ is calculated using $\atomfeat_t$ and $\atomfeat_{0}$ and normalized into $\mathbf{c}(\atomfeat_t, \atomfeat_0)$ so as to represent
probabilities. 
%
The proof of the above equations is available in Supplementary Section~\ref{supp:forward:proof}.

\subsubsection{{Backward Generative Process (\diffgenerative)}}
\label{section:diff:backward}

\methoddiff learns to reverse {\diffnoise} by denoising from {$(\pos_{t}, \atomfeat_{t})$} to {$(\pos_{t-1}, \atomfeat_{t-1})$} 
at $t\in[1,T]$, conditioned on the shape latent embedding $\shapehiddenmat$.
Specifically, the probabilities of $(\pos_{t-1}, \atomfeat_{t-1})$ denoised from $(\pos_{t}, \atomfeat_{t})$ are estimated by 
the approximates of the ground-truth posteriors $p(\pos_{t-1}|\pos_t, \pos_0)$ (Equation~\ref{eqn:gt_pos_posterior_1}) and 
$p(\atomfeat_{t-1}|\atomfeat_{t}, \atomfeat_0)$ (Equation~\ref{eqn:gt_atomfeat_posterior_1}).
%
Given that $(\pos_0, \atomfeat_0)$ is unknown in the generative process,
a prediction module $\molpred$,  which is a graph neural network with multiple layers,  (Section ``Shape-conditioned Molecule Prediction'') 
is employed to predict the atom position and feature $(\pos_0, \atomfeat_0)$ at time step $t$
as below,
\begin{equation}
(\tilde{\pos}_{0,t}, \tilde{\atomfeat}_{0,t}) = \molpred(\pos_t, \atomfeat_t, \hiddenmat^{\mathtt{s}}),
\label{eqn:predictor}
\end{equation}
where $\tilde{\pos}_{0,t}$ and $\tilde{\atomfeat}_{0,t}$ are the predictions of $\pos_0$ and $\atomfeat_0$ based on 
the information at $t$ (i.e., $\pos_t$, $\atomfeat_t$ and $\shapehiddenmat$). 
%

%
Following Ho \etal~\cite{ho2020ddpm}, with $\tilde{\pos}_{0,t}$, the probability of $\pos_{t-1}$ denoised from $\pos_t$, denoted as $p(\pos_{t-1}|\pos_t)$,
can be estimated 
by the approximated posterior $p_{\boldsymbol{\Theta}}(\pos_{t-1}|\pos_t, \tilde{\pos}_{0,t})$ as below,
\begin{equation}
\begin{aligned}
p(\pos_{t-1}|\pos_t) & \approx p_{\boldsymbol{\Theta}}(\pos_{t-1}|\pos_t, \tilde{\pos}_{0,t}) \\
& = \mathcal{N}(\pos_{t-1}|\mu_{\boldsymbol{\Theta}}(\pos_t, \tilde{\pos}_{0,t}),\tilde{\beta}_t^{\mathtt{x}}\mathbb{I}),
\end{aligned}
\label{eqn:aprox_pos_posterior}
\end{equation}
where ${\boldsymbol{\Theta}}$ is the learnable parameter; $\mu_{\boldsymbol{\Theta}}(\pos_t, \tilde{\pos}_{0,t})$ is an estimate 
of $\mu(\pos_t, \pos_{0})$ by replacing $\pos_0$ with its estimate $\tilde{\pos}_{0,t}$ 
in Equation~{\ref{eqn:gt_pos_posterior_1}}.
Similarly, with $\tilde{\atomfeat}_{0,t}$, the probability of $\atomfeat_{t-1}$ denoised from $\atomfeat_t$, denoted as $p(\atomfeat_{t-1}|\atomfeat_t)$, 
can be estimated 
by the approximated posterior $p_{\boldsymbol{\Theta}}(\atomfeat_{t-1}|\atomfeat_t, \tilde{\atomfeat}_{0,t})$ as below,
\begin{equation}
\begin{aligned}
p(\atomfeat_{t-1}|\atomfeat_t)\approx p_{\boldsymbol{\Theta}}(\atomfeat_{t-1}|\atomfeat_{t}, \tilde{\atomfeat}_{0,t}) 
=\mathcal{C}(\atomfeat_{t-1}|\mathbf{c}_{\boldsymbol{\Theta}}(\atomfeat_t, \tilde{\atomfeat}_{0,t})),\!\!\!\!
\end{aligned}
\label{eqn:aprox_atomfeat_posterior}
\end{equation}
where $\mathbf{c}_{\boldsymbol{\Theta}}(\atomfeat_t, \tilde{\atomfeat}_{0,t})$ is an estimate of $\mathbf{c}(\atomfeat_t, \atomfeat_0)$
by replacing $\atomfeat_0$  
with its estimate $\tilde{\atomfeat}_{0,t}$ in Equation~\ref{eqn:gt_atomfeat_posterior_1}.

\subsubsection{Model Training}
\label{section:diff:opt}

\method optimizes \methoddiff by minimizing the following three losses.

\paragraph{Atom Position Loss}
%
\method measures the squared errors between the predicted positions ($\tilde{\pos}_{0,t}$) from the prediction module $\molpred$ (Equation~\ref{eqn:predictor}) and the ground-truth positions ($\pos_0$) of atoms in molecules.
%
%
Given a particular step $t$, the loss is calculated as follows:
\begin{equation}
\label{eqn:diff:obj:pos}
\begin{aligned}
& \mathcal{L}^\mathtt{x}_t({\mol}) = w_t^\mathtt{x}\sum\nolimits_{\forall a \in {\scriptsize{\mol}}}\|\tilde{\pos}_{0,t} - \pos_{0}\|^2, \\
& \text{where} \ w_t^\mathtt{x} = \min(\lambda_t, \delta),\ \lambda_t={\bar{\alpha}^{\mathtt{x}}_t}/({1-\bar{\alpha}^{\mathtt{x}}_t}),
\end{aligned}
\end{equation}
%
where $w_t^\mathtt{x}$ is a weight at step $t$, and is calculated by clipping the signal-to-noise ratio 
$\lambda_t>0$ 
with a threshold $\delta > 0$. 
Note that because $\bar{\alpha}_t^{\mathtt{x}}$ decreases monotonically as $t$ increases from 1 to $T$ (Equation~\ref{eqn:noiseschedule}), $w_t^\mathtt{x}$ decreases monotonically over $t$ as well until it is clipped. 
Thus, $w_t^\mathtt{x}$ imposes lower weights on the loss when the noise level in $\pos_t$ is higher (i.e., $t$ close to $T$). 
This encourages the model training to focus more on accurately recovering molecule structures when there are 
sufficient signals in the data, rather than being potentially confused by major noises in the data. 
%


\paragraph{Atom Feature Loss}
%
\method also minimizes the KL divergence~\cite{kullback1951information} between the 
ground-truth posterior $p(\atomfeat_{t-1}|\atomfeat_t, \atomfeat_0)$ (Equation~\ref{eqn:gt_atomfeat_posterior_1}) 
and its approximate
$p_\theta(\atomfeat_{t-1}|\atomfeat_{t}, \tilde{\atomfeat}_{0,t})$ (Equation~\ref{eqn:aprox_atomfeat_posterior}) 
for discrete atom features
to optimize \methoddiff, following the literature~\cite{hoogeboom2021catdiff}. 
%
Particularly, the KL divergence at $t$ for a given molecule $\mol$ is calculated as follows:
%
\begin{equation*}
\mathcal{L}^{\mathtt{v}}_t({\mol})  = \sum\nolimits_{\forall a \in \scriptsize{\mol}}\text{KL}(p(\atomfeat_{t-1}|\atomfeat_{t}, \atomfeat_{0}) | p_{\boldsymbol{\Theta}}(\atomfeat_{t-1}|\atomfeat_{t}, \tilde{\atomfeat}_{0,t}) ),
\end{equation*}
\vspace{-10pt}
\begin{equation}
\label{eqn:lossv}
\! \! \! \! \! \!\! \! \! \! \! \!= \sum\nolimits_{\forall a\in \scriptsize{\mol}}\text{KL}(\mathbf{c}(\atomfeat_{t}, \atomfeat_{0})|\mathbf{c}_{\boldsymbol{\Theta}}(\atomfeat_{t}, \tilde{\atomfeat}_{0,t})),
\end{equation}
where $\mathbf{c}(\atomfeat_{t}, \atomfeat_{0})$ is a categorical distribution of $\atomfeat_{t-1}$  (Equation~\ref{eqn:gt_atomfeat_posterior_2});
$\mathbf{c}_{\boldsymbol{\Theta}}(\atomfeat_{t}, \tilde{\atomfeat}_{0,t})$ is an estimate of $\mathbf{c}(\atomfeat_{t}, \atomfeat_{0})$ (Equation~\ref{eqn:aprox_atomfeat_posterior}).
%
%

\paragraph{Bond Type Loss}
%
\method also minimizes the classification errors between the predicted bond types $(\bondpred_{ji})$ and the ground-truth types $\bondemb_{ij}$ of bonds in molecules. 
At step $t$, for each $l$-th layer of $\molpred$, \method predicts the bond types $(\bondpred_{ij,t,l})$ to understand the 
relations among atoms. 
%
%
Details about the calculation of $\bondpred_{ij,t,l}$ will be discussed later in Equation~\ref{eqn:bond_type}.
Given a particular step $t$, the error 
on bond type prediction
at the $l$-th layer is calculated as follows:
%
\begin{equation}
\label{eqn:loss_bond}
\mathcal{L}^{\mathtt{b}}_{t,l}({\mol}) = \sum_{\forall a_i\in \mol}\sum_{\forall a_j \in N(\scriptsize{\pos_{i,t}})} \text{H}(\bondpred_{ij,t,l}, \bondemb_{ij}),
\end{equation}
where {$N(\pos_{i,t})$ denotes the $k$-nearest neighbors of atom $a_i$ in position $\pos_{i,t}$}; 
$\text{H}(\cdot)$ denotes the cross-entropy loss. 
%
The bond type prediction loss across different layers is then aggregated as follows:
\begin{equation}
\mathcal{L}^{\mathtt{b}}_{t}({\mol}) = \frac{w_t^\mathtt{x}}{L-1}\sum_{l=1}^{L-1} \mathcal{L}^{\mathtt{b}}_{t,l}({\mol}) + w_t^\mathtt{x}\mathcal{L}^{\mathtt{b}}_{t, L}({\mol}),
\label{eqn:diff:obj:bond}
\end{equation}
where $w_t^\mathtt{x}$ is the weight at step $t$ used in Equation~\ref{eqn:diff:obj:pos}; $L$ is the number of layers in \molpred. 
Same with the Equation~\ref{eqn:diff:obj:pos}, the $w_t^\mathtt{x}$ in Equation~\ref{eqn:diff:obj:bond} is used to encourage the model training to focus more on accurately predicting bond types when the data provides sufficient signals, 
rather than being confused by major noises in the data. 
%
%
%
%
Note that, similar to Jumper \etal~\cite{Jumper2021}, in Equation~\ref{eqn:diff:obj:bond}, \method uses different weights on the last layer (i.e., $l$=$L$) and all the other layers, 
as we empirically find this design benefits the generation performance.
%
%


\paragraph{Overal \method Loss}
%
The overall \method loss function is defined as follows:
\begin{equation}
\label{eqn:loss}
\mathcal{L} =\sum\nolimits_{\forall \scriptsize{\mol}\in\mathcal{M}}\sum\nolimits_{\forall t \in \mathcal{T}} (\mathcal{L}^{\mathtt{x}}_t(\mol) + \xi \mathcal{L}^{\mathtt{v}}_t(\mol) + \zeta \mathcal{L}^{\mathtt{b}}_{t}({\mol})),
\end{equation}
where $\mathcal{M}$ is the set of all the molecules in training; 
$\mathcal{T}$ is the set of timesteps; 
$\xi > 0$ and $\zeta > 0$ are two \mbox{hyper-parameters} to balance $\mathcal{L}^{\mathtt{x}}_t$(\mol), $\mathcal{L}^{\mathtt{v}}_t$(\mol) and $\mathcal{L}^{\mathtt{b}}_{t}({\mol})$.
%
During training, step $t$ is uniformly sampled from $\mathcal{T}=\{1, 2, \cdots, 1000\}$.
The derivation of the loss functions
is available in Supplementary Section~\ref{supp:training:loss}.

\subsection*{Shape-conditioned Molecule Prediction ($\molpred$)}
\label{sec:method:rep:gvp}

In \diffgenerative, the prediction module $\molpred$ (Equation~\ref{eqn:predictor}) 
predicts the atom positions and features $(\tilde{\pos}_{0,t}, \tilde{\atomfeat}_{0,t})$ 
given the noisy data $(\pos_t, \atomfeat_t)$ conditioned on 
$\shapehiddenmat$.
For brevity, in this section, we eliminate the subscript $t$ in the notations when no ambiguity arises.
Particularly, as presented in Fig.~\ref{fig:overall}(d), 
$\molpred$ is a multi-layer graph neural network (GNN) comprising $L$ layers. 
In the $l$-th layer,
$\molpred$ 
uses the geometric vector perceptron (GVP) 
to learn a scalar embedding $\hiddenatomfeat_{i,l} \in\mathbb{R}^{d_a}$
and a vector embedding $\hiddenpos_i \in\mathbb{R}^{3\times d_r}$
for atom $a_i$ in an alternative manner that guarantees the invariance of $\hiddenatomfeat_{i,l}$ and the equivariance of $\hiddenpos_{i,l}$~\cite{jing2021learning}.
Intuitively, $\hiddenatomfeat_{i,l}$ captures inherent properties (e.g., atom types) of atom $a_i$, which are invariant of the molecule's orientation or position in 3D space.
%
Different from $\hiddenatomfeat_{i,l}$, $\hiddenpos_{i,l}$ captures geometric information (e.g., atom positions) of atom $a_i$, which will change under different 
transformations. 
We note that existing work primarily employs equivariant graph neural networks (EGNN)~\cite{satorras2021} 
for the prediction.
However, EGNN could suffer from limited capacity in capturing rich geometric information within molecules as it can only represent geometric features in a 3-dimensional latent space.
In contrast, GVP exhibits stronger expressiveness, capable of learning latent embeddings in spaces of any dimensions~\cite{torge2023diffhopp}.
Equipped with GVP, {$\molpred$} enables the learning of effective representations for geometric information.
Note that to ensure translation equivariance, \molpred shifts a fixed point (i.e., the center of shape condition $\pc$) to zero to eliminate all translations.
Therefore, only rotation equivariance needs to be considered. 

\molpred also leverages shape-aware scalar embeddings $\hat{\hiddenatomfeat}_{i,l}$ and vector embeddings $\hat{\hiddenpos}_{i,l}$ to generate molecules tailored to the shape condition. 
\molpred learns $\hat{\hiddenatomfeat}_{i,l}$ from $\hiddenatomfeat_{i,l}$ using the shape representation $\shapehiddenmat$ in an invariant way (Equation~\ref{eqn:shapea}). 
Similarly, \molpred learns $\hat{\hiddenpos}_{i,l}$ from $\hiddenpos_{i,l}$ and $\shapehiddenmat$ in an equivariant manner (Equation~\ref{eqn:shaper}).
%
%
%
In addition, \molpred utilizes the bond type embeddings to enhance the understanding of relations among atoms for better prediction (Equation~\ref{eqn:bond_type}).

{
Particularly, 
$\molpred$
estimates the type 
and position 
of the $i$-th atom $a_i$ as follows, }
%

\begin{equation}
\tilde{\pos}_{0, i} = \pos_i + 
\hiddenpos_{i,L}
, \quad \tilde{\atomfeat}_{0, i} = \text{softmax}(\text{MLP}(\hiddenatomfeat_{i,L})), 
\label{eqn:pred:pred}
\end{equation}
where $\tilde{\atomfeat}_{0, i}$ (Equation~\ref{eqn:predictor}) is the predicted probability distribution across all the types of atom features; 
$\tilde{\pos}_{0, i}$ (Equation~\ref{eqn:predictor}) is the predicted position of $a_i$; 
$\pos_i$ is the noisy position of $a_i$;
$\hiddenatomfeat_{i,L}$ and $\hiddenpos_{i,L}$ are the invariant scalar embedding and equivariant vector embedding for atom $a_i$, respectively. 
%
%
In each $l$-th layer, $\hiddenatomfeat_{i,l}$ and $\hiddenpos_{i,l}$ of atom $a_i$ are updated by propagating its neighborhood's inherent features and geometric features as follows,
\begin{equation}
\hiddenatomfeat_{i,l}, \hiddenpos_{i,l} = \text{GVP}(\mathbf{h}_{i,l}, \mathbf{y}_{i,l}),
\label{eqn:pred:gvp}
\end{equation}
\begin{equation}
\label{eqn:geometric_embedding}
\mathbf{h}_{i,l}= 
[\atomfeat_i, \hat{\hiddenatomfeat}_{i,l-1}, \sum_{j\in N(i)}e_{ji,l}\hiddenmessatom_{ji,l}, t]
, \quad
 \mathbf{y}_{i,l} = 
 [\pos_i, \hat{\hiddenpos}_{i,l-1}, \sum_{j\in N(i)}e_{ji,l}\hiddenmesspos_{ji,l}],
\end{equation}
\begin{equation}
\hat{\hiddenatomfeat}_{i, l-1}, \hat{\hiddenpos}_{i, l-1} = \SA(\hiddenatomfeat_{i, l-1}, \hiddenpos_{i, l-1}, \shapehiddenmat),
\end{equation}
where $\text{GVP}(\cdot)$ is a function that learns $\hiddenatomfeat_{i,l}$ and $\hiddenpos_{i,l}$ jointly from $\mathbf{h}_{i,l}\in \mathbb{R}^{d_h}$ and $\mathbf{y}_{i,l}\in\mathbb{R}^{3\times d_y}$; 
$[\cdot,\cdot]$ is the concatenation operation;
{$N(i)$ denotes the $k$-nearest neighbor atoms of atom $a_i$ over the 3D space;} 
%
$t$ denotes the time step;
$\atomfeat_i$ is the noisy feature vector of $a_i$;
%
%
$\hiddenmessatom_{ji,l}\in\mathbb{R}^{d_m}$ and $\hiddenmesspos_{ji,l}\in\mathbb{R}^{3\times d_n}$ are messages to propagate information from $a_j$ to $a_i$ as will be described in Equation~\ref{eqn:message};
$e_{ji,l}$ is the attention weight used to aggregate information from neighboring atoms;
\SA is a module to learn shape-aware atom embeddings as will be introduced later; 
$\hat{\hiddenatomfeat}_{i,l-1}$ and $\hat{\hiddenpos}_{i,l-1}$ are the shape-aware atom scalar and vector embedding, respectively (detailed in Equation~\ref{eqn:shapea} and Equation~\ref{eqn:shaper}). 
The weight $e_{ji,l}$ is calculated to estimate how much the neighboring atom $a_j$ should contribute to the 
learning of $\mathbf{h}_{i,l}$ and $\mathbf{y}_{i,l}$ as follows,
\begin{equation}
\label{eqn:attention}
\begin{aligned}
e_{ji,l} & = \frac{\exp(Q_{i,l}K_{ji,l})}{\sum_{k\in N(i)}\exp(Q_{i,l}K_{ki,l})},\\
 \text{where}\  Q_{i,l}& =\text{MLP}([\hat{\hiddenatomfeat}_{i,l-1}, \|\hat{\hiddenpos}_{i,l-1}\|^2]), \\
K_{ji,l}&=\text{MLP}([\hiddenmessatom_{ji,l}, \|\hiddenmesspos_{ji,l}\|^2]).
\end{aligned}
\end{equation}
%
%
In both Equation~\ref{eqn:geometric_embedding} and \ref{eqn:attention}, the messages $\hiddenmessatom_{ji,l}$ and $\hiddenmesspos_{ji,l}$ are calculated from the scalar embeddings (e.g., $\hat{\hiddenatomfeat}_{i,l}$) and vector embeddings (e.g., $\hat{\hiddenpos}_{i,l}$) of atoms as follows,
\begin{equation}
\hiddenmessatom_{ji,l}, \hiddenmesspos_{ji,l} = \text{GVP}(\hat{\hiddenmessatom}_{ji,l}, \hat{\hiddenmesspos}_{ji,l}),
\label{eqn:mess:gvp}
\end{equation}
\begin{equation}
\label{eqn:message}
\hat{\hiddenmessatom}_{ji,l} = 
[\hat{\hiddenatomfeat}_{j,l-1}, d_{ji}, \bondpred_{ji,l-1}], 
\quad \hat{\hiddenmesspos}_{ji,l} = 
[\hat{\hiddenpos}_{j,l-1}, \pos_j - \pos_i], 
\end{equation}
where $\text{GVP}(\cdot)$ is a function that learns $\hiddenmessatom_{ji,l}$ and $\hiddenmesspos_{ji,l}$ jointly from $\hat{\hiddenmessatom}_{ji,l}\in\mathbb{R}^{d_m}$ and $\hat{\hiddenmesspos}_{ji,l}\in\mathbb{R}^{3\times d_n}$; 
$[\cdot,\cdot]$ is the concatenation operation;
$\bondpred_{ji,l-1}$ is the embedding of the bond type between $a_i$ and $a_j$ (detailed in Equation~\ref{eqn:bond_type}); and $d_{ji}$ is the distance between $\pos_i$ and $\pos_j$.
\subsubsection*{Shape-aware Atom Representation Learning (\SA)}


To generate molecules that tailored to the shape condition represented by $\shapehiddenmat$, 
\molpred adapts the scalar embedding $\hiddenatomfeat_{i,l}$ and 
the vector embedding $\hiddenpos_{i,l}$ of each atom $a_i$ into the shape-aware scalar embedding $\hat{\hiddenatomfeat}_{i,l}$ and the shape-aware vector embedding $\hat{\hiddenpos}_{i,l}$ by incorporating $\shapehiddenmat$ at each layer.
Particularly, $\molpred$ learns $\hat{\hiddenatomfeat}_{i,l}$ for each atom $a_i$ using $\shapehiddenmat$ as follows,
%
%
%
%
\begin{equation}
\label{eqn:shapea}
\hat{\hiddenatomfeat}_{i,l} = \text{MLP}([\hiddenatomfeat_{i,l}, \interacthidden_{i,l}]), 
\end{equation}

%
%
where $[\cdot,\cdot]$ is the concatenation operation;
$\hiddenatomfeat_{i,L}$ is the scalar embedding of atom $a_i$ at the $l$-th layer; 
$\interacthidden_{i,l}\in\mathbb{R}^{d_o}$ 
represents the inherent relations 
between $a_i$ and the molecular surface shape, such as the signed distance from $a_i$ to the shape.
%
{\molpred learns $\interacthidden_{i,l}$ in a similar way to Equation~\ref{eqn:se:decoder} as follows,}
\begin{equation}
\interacthidden_{i,l} = \text{MLP}([\hiddenatomfeat_{i,l}, \langle \hiddenpos_{i,l}, \hiddenmat^{\mathtt{s}}\rangle, \|\hiddenpos_{i,l}\|, \text{VN-In}(\hiddenmat^{\mathtt{s}})]),
\label{eqn:diff:interact}
\end{equation}
where $\langle\hiddenpos_{i,l}, \hiddenmat^{\mathtt{s}}\rangle$ is the dot-product between $\hiddenpos_{i,l}$ and $\hiddenmat^{\mathtt{s}}$; $\|\hiddenpos_{i,l}\|^2$ is the column-wise Euclidean norm of the vector feature $\hiddenpos_{i,l}$; $\text{VN-In}(\hiddenmat^{\mathtt{s}})$ encodes the inherent geometry of shape condition and thus is shared across all the layers.
%
%
%
%
Apart from scalar embeddings, \molpred also incorporates shape information into the vector embeddings as follows, 
\begin{equation}
\label{eqn:shaper}
\hat{\hiddenpos}_{i,l} = \text{VN-MLP}([\hiddenpos_{i,l}, \shapehiddenmat]),
\end{equation}
where $\hiddenpos_{i,l}$ is the vector embedding of atom $a_i$ at the $l$-th layer; $\text{VN-MLP}(\cdot)$ is an equivariant VN network~\cite{deng2021vn} that learns non-linear interactions $\hat{\hiddenpos}_{i,l}\in\mathbb{R}^{3\times d_r}$ between $\hiddenpos_{i,l}$ and $\shapehiddenmat$ in an equivariant way. 
\subsubsection*{Bond Type Representation Learning ($\mathsf{BTRL}$)}
\label{sec:method:rep:bond}

As shown in Equation~\ref{eqn:message}, $\molpred$ leverages the types of bonds within \mol to facilitate its understanding of relations among atoms.
%
%
Particularly, for the bond between $a_j$ and $a_i$, \molpred generates the bond type embedding as follows, 
\begin{equation}
\bondpred_{ji,l} = 
	\begin{cases}
	\text{MLP}([\hiddenatomfeat_{i,l} + \hiddenatomfeat_{j,l}, \text{abs}(\hiddenatomfeat_{i,l} - \hiddenatomfeat_{j,l}), d_{ji}]), & \text{if}\ l = 0, \\
	\text{MLP}([\hiddenatomfeat_{i,l} + \hiddenatomfeat_{j,l}, \text{abs}(\hiddenatomfeat_{i,l} - \hiddenatomfeat_{j,l}), \|\hiddenpos_{i}\|^2 + \|\hiddenpos_{j}\|^2, \text{abs}(\|\hiddenpos_{i}\|^2 - \|\hiddenpos_{j}\|^2)]), & \text{if}\ l > 0,
	\end{cases}
	\label{eqn:bond_type}
\end{equation}
where $\hiddenatomfeat_{i,l}$ and $\hiddenpos_{i,l}$ is the scalar embedding and vector embedding of $a_i$ (Equation~\ref{eqn:geometric_embedding}),
respectively; $\text{abs}(\cdot)$  
represents the absolute difference; $d_{ji}$ is the distance between the positions $\pos_j$ and $\pos_i$.
\molpred guarantees that the predictions $\bondpred_{ij,l}$ and $\bondpred_{ji,l}$ are invariant to the permutation of atom $a_i$ and $a_j$.
This is achieved by using two invariant operations: the sum and the absolute difference operation.
To learn effective bond-type embeddings, we also
use the sum and the absolute difference of column-wise Euclidean norm of $\hiddenpos^l_{i}$ and $\hiddenpos^l_{j}$ to implicitly estimate the distance between $a_i$ and $a_j$.
When $l=0$, we directly use the distance $d_{ji}$ to calculate $\bondpred_{ji,l}$.
\subsection*{Guidance-induced Inference}
\label{section:method:guidance}


During inference, \method generates novel molecules by gradually denoising $(\pos_T, \atomfeat_T)$ to $(\pos_0, \atomfeat_0)$ using the prediction module $\molpred$.  
%
Specifically, \method samples $\pos_T$ and $\atomfeat_T$ from $\mathcal{N}(\mathbf{0}, \mathbb{I})$ and $\mathcal{C}(\mathbf{1}/K)$, respectively.
After that, \method samples $\pos_{t-1}$ from $\pos_t$ using $p_{\boldsymbol{\Theta}}(\pos_{t-1}|\pos_t, \tilde{\pos}_{0,t})$ (Equation~\ref{eqn:aprox_pos_posterior}).
Similarly, \method samples $\atomfeat_{t-1}$ from $\atomfeat_{t}$ using $p_{\boldsymbol{\Theta}}(\atomfeat_{t-1}|\atomfeat_{t}, \tilde{\atomfeat}_{0,t})$ (Equation~\ref{eqn:aprox_atomfeat_posterior}) until $t$ reaches 1.
\method uses post-processing to determine the bond type between atoms based on atomic distances following the previous work~\cite{peng22pocket2mol,guan2023targetdiff}.
Though the learned bond type embeddings in \method could provide valuable topology information for molecule prediction (\molpred), we observe that directly using predicted bond types in generated molecules could lead to sub-optimal performance. 

\subsubsection{{\method with Shape Guidance} {($\shapeguide$)}}
\label{section:method:guidance:shape-guide}

%
During molecule generation, as shown in Figure~\ref{fig:overall}(c), \method can also utilize additional shape guidance
by pushing the predicted atoms to the shape 
of the condition molecule \molx.
This approach is motivated by previous work~\cite{dhariwal2021diffusion}, which demonstrates that incorporating additional guidance into conditional diffusion models can further ensure the generated objects closely following the given condition.
%
%
Note that different from the shape for conditions, when used as guidance, 
we define molecule shapes as a set of points $\mathcal{Q}$ sampled according to atom positions in the condition molecule \molx following Adams and Coley \etal~\cite{adams2023equivariant}
We empirically find that this design leads to improved generation performance.
%
%
Particularly, for each atom $a_i$ in \molx, 20 points are randomly sampled into $\mathcal{Q}$ from a 
Gaussian distribution centered at $\pos_i$. 
%
Given the predicted atom position $\tilde{\pos}_{0,t}$ at step $t$, \method applies the shape guidance 
by {adjusting the predicted positions to $\mathcal{Q}$ 
as follows:
\begin{equation}
\label{eqn:shape_guidance}
\begin{aligned}
\pos_{0,t}^*=(1-\sigma) \tilde{\pos}_{0,t}+\sigma \!{\sum_{\mathclap{\mathbf{z}\in N(\tilde{\scriptsize{\pos}}_{0,t}; \mathcal{Q})}} \mathbf{z}}/{k},
\text{when }\sum_{\mathclap{\mathbf{z}\in N(\tilde{\scriptsize{\pos}}_{0,t}; \mathcal{Q})}} d(\tilde{\pos}_{0,t}, \mathbf{z}) / k>\gamma,\end{aligned}
\end{equation}
where $\sigma>0$ is the weight used to balance the prediction $\tilde{\pos}_{0,t}$ and the adjustment;
$d(\tilde{\pos}_{0, t}, \mathbf{z})$ is the Euclidean distance between $\tilde{\pos}_{0, t}$ and $\mathbf{z}$;
$N(\tilde{\pos}_{0,t};\mathcal{Q})$ is the set of $k$-nearest neighbors of $\tilde{\pos}_{0,t}$ in $\mathcal{Q}$ 
based on $d(\cdot)$;
$\gamma>0$ is a distance threshold. 
By doing the above adjustment, the predicted atom positions will be pushed to those of {\molx} if they
are sufficiently far away. 
Note that the shape guidance is applied exclusively for steps 
\begin{equation}
\label{eqn:steps}
t=T, T-1, \cdots, S\text{, where } S>1, 
\end{equation}
not for all the steps, 
and thus it only adjusts predicted atom positions when there are a lot of noises and the prediction needs 
more guidance. 
%
%
\method with the shape guidance is referred to as \methodwithsguide.
\subsubsection{\method with Protein Pocket Guidance {(\pocketguide)}}
\label{section:method:guidance:pocket}

%
%
%

When applying \method to PMG (i.e.,  protein pocket of the condition molecule is available), 
we observe that atoms in the generated molecules could be too close to the protein pocket atoms \pocket, thereby leading to steric clashes and thus undesirable binding affinities.
To address this issue, as shown in Figure~\ref{fig:overall}(e), \method utilizes pocket guidance to further adjust atom positions and maintain sufficient distances between molecule atoms and protein atoms.
%
%
%
Particularly, \method refines the atom positions in the generated molecules based on \pocket as follows,
\begin{equation}
\label{eqn:pocket_guidance}
\begin{aligned}
\pos_{t}^*=\pos_{t} + \frac{\pos_t - \mathbf{z}}{d(\pos_{t}, \mathbf{z})} * (\rho - d(\pos_{t}, \mathbf{z}) + \epsilon)
\quad \text{if }\exists\ \mathbf{z}\in N(\pos_{t}; \mathcal{K}), d(\pos_{t}, \mathbf{z}) <\rho, 
\end{aligned}
\end{equation}
where $\pos_{t}$ is the sampled atom positions at the step $t$; $N(\pos_{t};\mathcal{K})$ is the set of $k$-nearest neighbors of $\pos_{t}$ within the protein atoms $\mathcal{K}$; 
$d(\pos_{t}, \mathbf{z})$ is the distance between $\pos_{t}$ and $\mathbf{z}$, 
{and $\frac{\scriptsize{\pos_{t}} - \mathbf{z}}{d(\scriptsize{\pos_{t}}, \mathbf{z})}$ calculates the unit vector in the direction that moves $\pos_{t}$ far from $\mathbf{z}$.}
%
\method introduces a threshold $\rho$ 
to assess if protein atoms and molecule atoms are too close.
{\method} identifies this threshold from known protein-ligand complexes in the training dataset.}
{\method also introduces a hyper-parameter $\epsilon$ to control the margin. }
%
%
Note that different from the shape guidance that is applied on $\tilde{\pos}_{0, t}$, 
the pocket guidance is applied on $\pos_{t}$.
We empirically find this design benefits the generated molecules in their binding affinities to protein pockets.
\method with the pocket guidance is referred to as \methodwithpguide. 
%
%
%
%


\section*{Data Availability}
\label{section:data_availability}

The data used in this manuscript is made publicly available at the link \url{https://github.com/ninglab/DiffSMol}.

\section*{Code Availability}
\label{section:code_availability}

The code for \method is made publicly available at the link \url{https://github.com/ninglab/DiffSMol}..

\section*{Acknowledgements}
\label{section:acknowledgements}

This project was made possible, in part, by support from the National Science Foundation grant no. IIS-2133650 (X.N., Z.C.), 
and the National Library of Medicine grant no. 1R01LM014385 (X.N.). 
Any opinions, findings, and conclusions or recommendations expressed in this manuscript are those of the authors, and do not necessarily reflect the views of the funding agencies.
We thank Patrick J. Lawrence, Frazier N. Baker, and Vishal Dey for their constructive comments.

\section*{Author Contributions}
\label{section:author_contribution}

X.N. conceived the research. X.N. obtained funding for the research. Z.C. and X.N. designed the research. Z.C.
and X.N. conducted the research, including data curation, formal analysis, methodology design and implementation,
result analysis and visualization. Z.C., B.P. and X.N. drafted the original manuscript. T.Z. and D.A. provided comments
on case studies. Z.C., B.P. and X.N. conducted the manuscript editing and revision. All authors reviewed the final manuscript.

\section*{Competing Interests}
\label{section:author_contribution}

The authors declare that the research was conducted in the absence of any commercial or financial relationships that could be construed as a potential conflict of interest.

\clearpage

\bibliographystyle{naturemag}
\bibliography{paper}

\clearpage

\input{supp}

\end{document}

%% file: define.tex
\newcommand{\etal}{\mbox{\emph{et al.}}\xspace}

\newcommand{\ziqi}[1]{\textcolor{blue}{#1}}
\newcommand{\xia}[1]{\textcolor{red}{#1}}
\newcommand{\bo}[1]{\textcolor{magenta}{#1}}
\newcommand{\todo}[1]{\textcolor{olive}{[todo: #1]}}

\newcommand{\method}{\mbox{$\mathop{\mathsf{DiffSMol}}\limits$}\xspace}
\newcommand{\SE}{\mbox{$\mathop{\mathsf{SE}}\limits$}\xspace}
\newcommand{\SEE}{\mbox{$\mathop{\mathsf{SE}\text{-}\mathsf{enc}}\limits$}\xspace}
\newcommand{\SED}{\mbox{$\mathop{\mathsf{SE}\text{-}\mathsf{dec}}\limits$}\xspace}
\newcommand{\molpred}{\mbox{$\mathop{\mathsf{SMP}}\limits$}\xspace}
\newcommand{\methoddiff}{\mbox{$\mathop{\mathsf{DIFF}}\limits$}\xspace}
\newcommand{\SA}{\mbox{$\mathop{\mathsf{SARL}}\limits$}\xspace}
\newcommand{\shapeguide}{\mbox{$\mathop{\mathsf{SG}}\limits$}\xspace}
\newcommand{\pocketguide}{\mbox{$\mathop{\mathsf{PG}}\limits$}\xspace}

\newcommand{\diffnoise}{\mbox{$\mathop{\mathsf{DIFF}\texttt{-}\mathsf{forward}}\limits$}\xspace}
\newcommand{\diffgenerative}{\mbox{$\mathop{\mathsf{DIFF}\texttt{-}\mathsf{backward}}\limits$}\xspace}
\newcommand{\methodwithsguide}{\mbox{$\mathop{\mathsf{DiffSMol}\texttt{+}\mathsf{s}}\limits$}\xspace}
\newcommand{\methodwithsandpguide}{\mbox{$\mathop{\mathsf{DiffSMol}\texttt{+}\mathsf{s}\texttt{+}\mathsf{p}}\limits$}\xspace}
\newcommand{\methodwithpguide}{\mbox{$\mathop{\mathsf{DiffSMol}\texttt{+}\mathsf{p}}\limits$}\xspace}

\newcommand{\eqgnn}{\mbox{$\mathop{\mathsf{EQ}\text{-}\mathsf{GNN}}\limits$}\xspace}
\newcommand{\invgnn}{\mbox{$\mathop{\mathsf{INV}\text{-}\mathsf{GNN}}\limits$}\xspace}

\newcommand{\squid}{\mbox{$\mathop{\mathsf{SQUID}}\limits$}\xspace}
\newcommand{\dataset}{\mbox{$\mathop{\mathsf{VS}}\limits$}\xspace}

\newcommand{\pockettwomol}{\mbox{$\mathop{\mathsf{Pocket2Mol}}\limits$}\xspace}
\newcommand{\targetdiff}{\mbox{$\mathop{\mathsf{TargetDiff}}\limits$}\xspace}
\newcommand{\decompdiff}{\mbox{$\mathop{\mathsf{DecompDiff}}\limits$}\xspace}
\newcommand{\decompdiffbeta}{\mbox{$\mathop{\mathsf{DecompDiff}\text{+}\mathsf{b}}\limits$}\xspace}
\newcommand{\decompdiffref}{\mbox{$\mathop{\mathsf{DecompDiff}}\limits$}\xspace}
\newcommand{\AR}{\mbox{$\mathop{\mathsf{AR}}\limits$}\xspace}

\newcommand{\shape}{\mbox{$\mathop{\mathtt{s}}\limits$}\xspace}
\newcommand{\mol}{\mbox{$\mathop{\mathtt{M}}\limits$}\xspace}
\newcommand{\molx}{\mbox{$\mathop{\mathtt{M}_x}\limits$}\xspace}
\newcommand{\moly}{\mbox{$\mathop{\mathtt{M}_y}\limits$}\xspace}
\newcommand{\pc}{\mbox{$\mathop{\mathcal{P}}\limits$}\xspace}
\newcommand{\pocket}{\mbox{$\mathop{\mathcal{K}}\limits$}\xspace}

\newcommand{\pos}{\mbox{$\mathop{\mathbf{x}}\limits$}\xspace}
\newcommand{\pointpos}{\mbox{$\mathop{\mathbf{z}}\limits$}\xspace}
\newcommand{\atomfeat}{\mbox{$\mathop{\mathbf{v}}$}\xspace}

\newcommand{\hiddenvec}{\mbox{$\mathop{\mathbf{h}}\limits$}\xspace}
\newcommand{\hiddenmat}{\mbox{$\mathop{\mathbf{H}}\limits$}\xspace}
\newcommand{\interacthidden}{\mbox{$\mathop{\mathbf{o}}\limits$}\xspace}
\newcommand{\hiddenatomfeat}{\mbox{$\mathop{\mathbf{a}}\limits$}\xspace}
\newcommand{\hiddenpos}{\mbox{$\mathop{\mathbf{r}}\limits$}\xspace}
\newcommand{\hiddenmessatom}{\mbox{$\mathop{\mathbf{m}}\limits$}\xspace}
\newcommand{\hiddenmesspos}{\mbox{$\mathop{\mathbf{n}}\limits$}\xspace}
\newcommand{\bondemb}{\mbox{$\mathop{\mathbf{b}}\limits$}\xspace}
\newcommand{\bondpred}{\mbox{$\mathop{\mathbf{e}}\limits$}\xspace}

\newcommand{\shapehiddenmat}{\mbox{$\mathop{\mathbf{H}^{\mathtt{s}}}\limits$}\xspace}

\newcommand{\cumalpha}{\mbox{$\mathop{\bar{\alpha}}\limits$}\xspace}

\newcommand{\alphav}{\mbox{$\mathop{\alpha^{\scriptsize{v}}}\limits$}\xspace}

\newcommand{\shapesim}{\mbox{$\mathop{\mathsf{Sim}_{\mathtt{s}}}\limits$}\xspace}
\newcommand{\avgshapesim}{\mbox{$\mathop{\mathsf{avgASim}_{\mathtt{s}}}\limits$}}
\newcommand{\maxshapesim}{\mbox{$\mathop{\mathsf{avgMSim}_{\mathtt{s}}}\limits$}}

\newcommand{\graphsim}{\mbox{$\mathop{\mathsf{Sim}_{\mathtt{g}}}\limits$}\xspace}
\newcommand{\avggraphsim}{\mbox{$\mathop{\mathsf{avgASim}_{\mathtt{g}}}\limits$}}
\newcommand{\maxgraphsim}{\mbox{$\mathop{\mathsf{avgMSim}_{\mathtt{g}}}\limits$}}

\newcommand{\diversity}{\mbox{$\mathop{\mathsf{Div_d}}\limits$}\xspace}

\newcommand{\qed}{\mbox{$\mathop{\mathsf{QED}}\limits$}\xspace}
\newcommand{\desire}{\mbox{$\mathop{\#\mathsf{d}\%}\limits$}\xspace}
\newcommand{\novel}{\mbox{$\mathop{\#\mathsf{n}\%}\limits$}\xspace}

\algnewcommand{\FullLineComment}[1]{\Statex{$\triangleright$~#1}}
\algnewcommand{\IndentLineComment}[2]{\Statex{\hspace{#2em}$\triangleright$~#1}}

%% file: tables/dataset.tex
\begin{table*}[!t]
  \centering
      \caption{Data Statistics for SMG and PMG}
  \label{tbl:data}
  \begin{threeparttable}
 \begin{scriptsize}
      \begin{tabular}{
	@{\hspace{2pt}}l@{\hspace{10pt}}
	@{\hspace{10pt}}l@{\hspace{10pt}} 
	@{\hspace{10pt}}l@{\hspace{10pt}} 
	@{\hspace{10pt}}r@{\hspace{2pt}}         
	}
        \toprule
        Task & Dataset & Description & Statistics \\
        \midrule
        \multirow{3}{*}{SMG} &  \multirow{3}{*}{MOSES}
         & \#training molecules            & 1,592,653 \\
         & & \#validation molecules    & 1,000 \\
         & & \#test molecules        &  1,000 \\
         \midrule
         PMG & CrossDocked2020
         & \#test protein-ligand complexes &  72          \\
        \bottomrule
      \end{tabular}
%
%
\end{scriptsize}
  \end{threeparttable}
\end{table*}

%% file: tables/overall_results_desirable.tex
\begin{table*}[h]
	\centering
		\caption{Comparison on Desirable Molecules for SMG}
	\label{tbl:overall_desirable}
\begin{threeparttable}
	\begin{scriptsize}
	\begin{tabular}{
		@{\hspace{3pt}}l@{\hspace{3pt}}
		@{\hspace{3pt}}r@{\hspace{8pt}}
		@{\hspace{3pt}}c@{\hspace{3pt}}
		@{\hspace{3pt}}r@{\hspace{3pt}}
		@{\hspace{4pt}}c@{\hspace{4pt}}
		@{\hspace{3pt}}r@{\hspace{3pt}}
		@{\hspace{3pt}}c@{\hspace{3pt}}
		@{\hspace{3pt}}r@{\hspace{3pt}}
		@{\hspace{4pt}}c@{\hspace{4pt}}
		@{\hspace{3pt}}r@{\hspace{3pt}}
		@{\hspace{3pt}}c@{\hspace{3pt}}
		@{\hspace{3pt}}r@{\hspace{3pt}}
		@{\hspace{4pt}}c@{\hspace{4pt}}
		@{\hspace{3pt}}r@{\hspace{3pt}}
		@{\hspace{3pt}}c@{\hspace{3pt}}
		@{\hspace{3pt}}r@{\hspace{3pt}}
		%
		}
		\toprule
		\multirow{2}{*}{method}  & \multicolumn{3}{c}{$\delta_g$=0.3} & & \multicolumn{3}{c}{$\delta_g$=0.5} & & \multicolumn{3}{c}{$\delta_g$=0.7} & & \multicolumn{3}{c}{$\delta_g$=1.0}  \\
		\cmidrule{2-4} \cmidrule{6-8} \cmidrule{10-12} \cmidrule{14-16}
		& \desire & \diversity & \novel & & \desire & \diversity & \novel & &  \desire & \diversity & \novel & & \desire & \diversity & \novel \\ 
		\midrule
		\dataset &  10.6     & 0.736         &  0.0 & & 12.2            & 0.734 & 0.0 &  &  12.3 & 0.734 &  0.0 & &  12.3 & 0.734 & 0.0 \\
		\squid($\lambda$=0.3) & 8.3             & 0.669     & 96.6 & & 21.8            & 0.649    & 96.2 &  &  27.5            & 0.633 & 95.7 & &  31.3            & 0.617  & 92.6 \\
		\squid($\lambda$=1.0) & 11.2            & 0.728    & 96.9 & & 14.4            & 0.721    & 96.7 &  &  14.6            & 0.720 & 96.6 & & 14.7            & 0.720   & 96.6 \\
		\method               &  \underline{28.4}    & \textbf{0.762} & \underline{99.8} & &  \underline{32.3}    & \textbf{0.751} & \underline{99.8} & &  \underline{32.4}    & \textbf{0.751} & \underline{99.8}  &  &  \underline{32.4}    & \textbf{0.751}   & \underline{99.8}  \\
		\methodwithsguide     & \textbf{61.4}   & \underline{0.760}   & \textbf{99.9} & & \textbf{70.9}   & \underline{0.748}   & \textbf{99.9} & &  \textbf{71.0}   & \underline{0.748}  & \textbf{99.9}  &  &  \textbf{71.0}   & \underline{0.748}   & \textbf{99.9}  \\
	\bottomrule
	\end{tabular}%
	\begin{tablenotes}[normal,flushleft]
		\begin{footnotesize}
	\item 
\!\!Columns represent: {``$\delta_g$'': the graph similarity constraint; 
``\desire'': the percentage of generated molecules that are desirable, satisfying $\delta_g$ and exhibiting high \shapesim ($\shapesim>=0.8$);
``\diversity'': the diversity among the desirable molecules;
``\novel'': the percentage of desirable molecules that cannot be found in the MOSES dataset.
%
%
 Best values are in \textbf{bold}, and second-best values are \underline{underlined}. 
} 
\par
		\par
		\end{footnotesize}
	\end{tablenotes}
	\end{scriptsize}
\end{threeparttable}
\end{table*}

%% file: tables/overall_results_quality_desired.tex
\begin{table*}[ht]
	\centering
		\caption{{Comparison on Quality of Generated Desirable Molecules between \method and \squid ($\delta_g$=0.3)}}
	\label{tbl:overall_results_quality_desired}
\begin{threeparttable}
	\begin{scriptsize}
	\begin{tabular}{
		@{\hspace{0pt}}l@{\hspace{14pt}}
		@{\hspace{0pt}}l@{\hspace{2pt}}
		@{\hspace{4pt}}c@{\hspace{4pt}}
		%
		@{\hspace{3pt}}c@{\hspace{3pt}}
		@{\hspace{3pt}}c@{\hspace{3pt}}
		@{\hspace{3pt}}c@{\hspace{3pt}}
		@{\hspace{3pt}}c@{\hspace{3pt}}
		}
		\toprule
		group & metric & 
        & \squid ($\lambda$=0.3) & \squid ($\lambda$=1.0)  &  \method & \methodwithsguide \\
		\midrule

		\multirow{2}{*}{stability}
		& atom stability ($\uparrow$) & 
        & \textbf{0.996} & \textbf{0.996} & 0.993 & 0.989     \\
		& molecule stability ($\uparrow$) & 
        & \textbf{0.953} & 0.951 & 0.891 & 0.850     \\
		\midrule
		\multirow{4}{*}{3D structures} 
		& RMSD ($\downarrow$) & 
        & 0.912 & 0.902 & 0.895 & \textbf{0.882}    \\
		& JS. bond lengths ($\downarrow$) & 
        & 0.457 & 0.477 & 0.436 & \textbf{0.428}    \\
		& JS. bond angles ($\downarrow$) & 
        & 0.269 & 0.289 & \textbf{0.186} & 0.200   \\
		& JS. dihedral angles ($\downarrow$) & 
        & 0.199 & 0.209 & \textbf{0.168} & 0.170    \\
		\midrule
		\multirow{5}{*}{2D structures} 
		& JS. \#bonds per atom ($\downarrow$) & 
        & 0.313 & 0.328 & \textbf{0.176} & 0.180    \\
		& JS. basic bond types ($\downarrow$) & 
        & \textbf{0.070} & 0.081 & 0.180 & 0.190    \\
		& JS. \#rings ($\downarrow$) & 
        & 0.309 & 0.328 & \textbf{0.042} & 0.048    \\
		& JS. \#n-sized rings ($\downarrow$) & 
        & \textbf{0.088} & 0.091 & 0.098 & 0.111    \\
		& \#Intersecting rings ($\uparrow$) & 
        & \textbf{6} & 5 & 4 & 5    \\
		%
		\bottomrule
	\end{tabular}%
	\begin{tablenotes}[normal,flushleft]
		\begin{footnotesize}
	\item 
\!\!Rows represent:  {``atom stability'': the proportion of stable atoms that have the correct valency; 
		``molecule stability'': the proportion of generated molecules with all atoms stable;
		``RMSD'': the root mean square deviation (RMSD) between the generated 3D structures of molecules and their optimal conformations; 
		``JS. bond lengths/bond angles/dihedral angles'': the Jensen-Shannon (JS) divergences of bond lengths, bond angles and dihedral angles;
		``JS. \#bonds per atom/basic bond types/\#rings/\#n-sized rings'': the JS divergences of bond counts per atom, basic bond types, counts of all rings, and counts of n-sized rings;
		``\#Intersecting rings'': the number of rings observed in the top-10 frequent rings of both generated and real molecules. }
	\par
	\par
		\end{footnotesize}
	\end{tablenotes}
\end{scriptsize}
\end{threeparttable}
\end{table*}

%% file: tables/overall_results_docking3.tex
\begin{table*}[!ht]
	\centering
		\caption{Overall Comparison on PMG}
	\label{tbl:overall_results_docking2}
\begin{threeparttable}
	\begin{scriptsize}
	\begin{tabular}{
		@{\hspace{2pt}}l@{\hspace{2pt}}
		@{\hspace{2pt}}r@{\hspace{2pt}}
		@{\hspace{2pt}}r@{\hspace{2pt}}
		@{\hspace{4pt}}r@{\hspace{4pt}}
		@{\hspace{2pt}}r@{\hspace{2pt}}
		@{\hspace{2pt}}r@{\hspace{2pt}}
		@{\hspace{5pt}}r@{\hspace{5pt}}
		@{\hspace{2pt}}r@{\hspace{2pt}}
		@{\hspace{2pt}}r@{\hspace{2pt}}
		@{\hspace{5pt}}r@{\hspace{5pt}}
		@{\hspace{2pt}}r@{\hspace{2pt}}
	         @{\hspace{2pt}}r@{\hspace{2pt}}
		@{\hspace{5pt}}r@{\hspace{5pt}}
		@{\hspace{2pt}}r@{\hspace{2pt}}
		@{\hspace{2pt}}r@{\hspace{2pt}}
		@{\hspace{5pt}}r@{\hspace{5pt}}
		@{\hspace{2pt}}r@{\hspace{2pt}}
		@{\hspace{2pt}}r@{\hspace{2pt}}
		@{\hspace{5pt}}r@{\hspace{5pt}}
		@{\hspace{2pt}}r@{\hspace{2pt}}
		@{\hspace{2pt}}r@{\hspace{2pt}}
		@{\hspace{5pt}}r@{\hspace{5pt}}
		@{\hspace{2pt}}r@{\hspace{2pt}}
		}
		\toprule
		\multirow{2}{*}{method} & \multicolumn{2}{c}{Vina S$\downarrow$} & & \multicolumn{2}{c}{Vina M$\downarrow$} & & \multicolumn{2}{c}{Vina D$\downarrow$} & & \multicolumn{2}{c}{{HA\%$\uparrow$}}  & & \multicolumn{2}{c}{QED$\uparrow$} & & \multicolumn{2}{c}{SA$\uparrow$} & & \multicolumn{2}{c}{Div$\uparrow$} & 
		& \multirow{2}{*}{time$\downarrow$} \\
	    \cmidrule{2-3}\cmidrule{5-6} \cmidrule{8-9} \cmidrule{11-12} \cmidrule{14-15} \cmidrule{17-18} \cmidrule{20-21}
		& Avg. & Med. &  & Avg. & Med. &  & Avg. & Med. & & Avg. & Med.  & & Avg. & Med.  & & Avg. & Med.  & & Avg. & Med.  & & \\ 
		\midrule
		Reference                          & -5.32 & -5.66 & & -5.78 & -5.76 & & -6.63 & -6.67 & & - & - & & 0.53 & 0.49 & & 0.77 & 0.77 & & - & - & 
		& - \\
		\midrule
		\AR & -5.06 & -4.99 & &  -5.59 & -5.29 & &  -6.16 & -6.05 & &  37.69 & 31.00 & &  0.50 & 0.49 & &  0.66 & 0.65 & & 0.70 & 0.70 & 
		& 7,789 \\
		\pockettwomol   & -4.50 & -4.21 & &  -5.70 & -5.27 & &  -6.43 & -6.25 & &  48.00 & 51.00 & &  0.58 & 0.58 & &  \textbf{0.77} & \textbf{0.78} & &  0.69 & 0.71 &  
		& 2,150 \\
		\targetdiff     & -4.88 & \underline{-5.82} & &  -6.20 & \underline{-6.36} & &  \textbf{-7.37} & \underline{-7.51} & &  57.57 & 58.27 & &  0.50 & 0.51 & &  0.60 & 0.59 & &  \textbf{0.72} & 0.71 & 
		& 1,252 \\
		\decompdiffref  & -4.58 & -4.77 & &  -5.47 & -5.51 & &  -6.43 & -6.56 & &  47.76 & 48.66 & &  0.56 & 0.56 & &  0.70 & 0.69  & &  \textbf{0.72} & \textbf{0.72} &  
		& 1,859 \\
		\midrule
		\methodwithpguide       &  \underline{-5.53} & -5.64 & &  \underline{-6.37} & -6.33 & &  \underline{-7.19} & \textbf{-7.52} & &  \underline{78.75} & \textbf{94.00} & &  \textbf{0.77} & \textbf{0.80} & &  \underline{0.76} & \underline{0.76} & & 0.63 & 0.66 & 
		& 462 \\
		 \methodwithsandpguide   & \textbf{-5.81} & \textbf{-5.96} & &  \textbf{-6.50} & \textbf{-6.58} & & -7.16 & \underline{-7.51} & &  \textbf{79.92} & \underline{93.00} & &  \underline{0.76} & \underline{0.79} & &  0.75 & 0.74 & & 0.64 & 0.66 &
		& 561 \\
		\bottomrule
	\end{tabular}%
	\begin{tablenotes}[normal,flushleft]
		\begin{footnotesize}
	\item 
\!\!Columns represent: {``Vina S'': the binding affinities between the initially generated poses of molecules and the protein pockets; 
		``Vina M'': the binding affinities between the poses after local structure minimization and the protein pockets;
		``Vina D'': the binding affinities between the poses determined by AutoDock Vina~\cite{Eberhardt2021} and the protein pockets;
		``HA'': the percentage of generated molecules with Vina D higher than those of condition molecules;
		``QED'': the drug-likeness score;
		``SA'': the synthesizability score;
		``Div'': the diversity among generated molecules;
		``time'': the time cost to generate molecules.}
		
		\par
		\par
		\end{footnotesize}
	\end{tablenotes}
	\end{scriptsize}
\end{threeparttable}
  \vspace{-10pt}    
\end{table*}

%% file: tables/overall_docking_results_quality1.0.tex
\begin{table*}[t!]
	\centering
		\caption{Comparison on Quality of Generated Molecules for PMG}
	\label{tbl:overall_docking_results_quality_10}
	\begin{scriptsize}
\begin{threeparttable}
	\begin{tabular}{
		@{\hspace{0pt}}l@{\hspace{8pt}}
		@{\hspace{0pt}}l@{\hspace{2pt}}
		@{\hspace{4pt}}c@{\hspace{4pt}}
		%
		@{\hspace{3pt}}c@{\hspace{3pt}}
		@{\hspace{3pt}}c@{\hspace{3pt}}
		@{\hspace{3pt}}c@{\hspace{3pt}}
		@{\hspace{3pt}}c@{\hspace{3pt}}
		@{\hspace{3pt}}c@{\hspace{3pt}}
		@{\hspace{3pt}}c@{\hspace{3pt}}
		}
		\toprule
		group & metric & & \AR & \pockettwomol & \targetdiff & \decompdiffref & \methodwithpguide & \methodwithsandpguide \\
		\midrule
		\multirow{2}{*}{stability}
		& atom stability ($\uparrow$) & 
			& 0.907 & {0.841} & \textbf{0.949} & 0.920 & {0.934} & 0.910     \\
		& molecule stability ($\uparrow$) & 
			& 0.499 & 0.167 & 0.456 & 0.391 & \textbf{0.581} & 0.485    \\
		\midrule
		\multirow{4}{*}{3D structures} 
		& RMSD ($\downarrow$) & & 0.656 & \textbf{0.369} & 0.918 & 0.815 & 0.663 & 0.675    \\
		& JS. bond lengths ($\downarrow$) & & {0.472} & 0.428 & 0.340 & 0.278 & \textbf{0.274} & 0.278    \\
		& JS. bond angles ($\downarrow$) & & 0.342 & 0.227 & 0.212 & \textbf{0.137} & 0.197 & 0.219    \\
		& JS. dihedral angles ($\downarrow$) & & {0.415} & 0.292 & 0.268 & 0.203 & \textbf{0.185} & 0.186    \\
		\midrule
		\multirow{5}{*}{2D structures} 
		& JS. \#bonds per atom ($\downarrow$) & & {0.318} & 0.293 & \textbf{0.140} & 0.266 & 0.279 & 0.288    \\
		& JS. basic bond types ($\downarrow$) & & 0.223 & \textbf{0.055} & {0.244} & 0.155 & 0.061 & 0.080    \\
		& JS. \#rings ($\downarrow$) & & 0.213 & 0.208 & 0.109 & {0.262} & \textbf{0.067} & 0.071    \\
		& JS. \#n-sized rings ($\downarrow$) & & 0.141 & \textbf{0.077} & {0.149} & 0.126 & 0.115 & 0.124   \\
		& \#Intersecting rings ($\uparrow$) & & 6 & 4 & \textbf{7} & \textbf{7} & 6 & \textbf{7} \\
		\bottomrule
	\end{tabular}%
	\begin{footnotesize}
	\begin{tablenotes}[normal,flushleft]
\item \!\!Rows represent:  {``atom stability'': the proportion of stable atoms that have the correct valency; 
		``molecule stability'': the proportion of generated molecules with all atoms stable;
		``RMSD'': the root mean square deviation (RMSD) between the generated 3D structures of molecules and their optimal conformations; 
		``JS. bond lengths/bond angles/dihedral angles'': the Jensen-Shannon (JS) divergences of bond lengths, bond angles and dihedral angles;
		``JS. \#bonds per atom/basic bond types/\#rings/\#n-sized rings'': the JS divergences of bond counts per atom, basic bond types, counts of all rings, and counts of n-sized rings;
		``\#Intersecting rings'': the number of rings observed in the top-10 frequent rings of both generated and real molecules. }
		\par
	\end{tablenotes}
\end{footnotesize}
\end{threeparttable}
\end{scriptsize}
  \vspace{-10pt}    
\end{table*}

%% file: tables/notations.tex
\begin{table}[!h]
  \caption{{Notations}}
  \vspace{-10pt}
  \label{tbl:notations}
  \centering
  \begin{threeparttable}
     \begin{footnotesize}
      \begin{tabular}{
	@{\hspace{3pt}}l@{\hspace{3pt}}
    @{\hspace{3pt}}l@{\hspace{3pt}}
	}
        \toprule
        notations & meanings \\
        \midrule
        \mol & a molecule\\
        $a_i$ & the $i$-th atom in \mol\\
        $\pos_i$ & the position of $a_i$ in 3D space\\
        $\atomfeat_i$ & the feature vector of $a_i$\\
        $d_{ij}$ & the distance between $a_i$ and $a_j$\\
        $\mathbf{b}_{ij}$ & the one-hot feature vector indicating the bond type between $a_i$ and $a_j$\\ 
        $\shape$ & the 3D surface shape of \mol\\
        $\pc$ & the point cloud for $\shape$\\
        $z_i$ & the $i$-th point in $\pc$\\
        $\shapehiddenmat$ & the latent embedding of $\pc$ \\
        $o$ & the signed distance of a point randomly sampled in 3D space to the molecule surface\\
        \bottomrule
      \end{tabular}
      \end{footnotesize}
  \end{threeparttable}
\end{table}

%% file: supp.tex



\setcounter{secnumdepth}{2} 

\setcounter{section}{0}
\renewcommand{\thesection}{S\arabic{section}}

\setcounter{table}{0}
\renewcommand{\thetable}{S\arabic{table}}

\setcounter{figure}{0}
\renewcommand{\thefigure}{S\arabic{figure}}

\setcounter{algorithm}{0}
\renewcommand{\thealgorithm}{S\arabic{algorithm}}

\setcounter{equation}{0}
\renewcommand{\theequation}{S\arabic{equation}}

\begin{center}
	\begin{minipage}{0.95\linewidth}
		\centering
		\LARGE 
	Generating 3D Binding Molecules Using Shape-Conditioned Diffusion Models with Guidance (Supplementary Information)
	\end{minipage}
\end{center}
\vspace{10pt}

\section{Parameters for Reproducibility}
\label{supp:experiments:parameters}

We implemented both \SE and \methoddiff using Python-3.7.16, PyTorch-1.11.0, PyTorch-scatter-2.0.9, Numpy-1.21.5, Scikit-learn-1.0.2.
We trained the models using a Tesla V100 GPU with 32GB memory and a CPU with 80GB memory on Red Hat Enterprise 7.7.
%

\subsection{Parameters of \SE}

In \SE, we tuned the dimension of all the hidden layers including VN-DGCNN layers
(Eq.~\ref{eqn:shape_embed}), MLP layers (Eq.~\ref{eqn:se:decoder}) and
VN-In layer (Eq.~\ref{eqn:se:decoder}), and the dimension $d_p$ of generated shape latent embeddings $\shapehiddenmat$ with the grid-search algorithm in the 
parameter space presented in Table~\ref{tbl:hyper_se}.
We determined the optimal hyper-parameters according to the mean squared errors of the predictions of signed distances for 1,000 validation molecules that are selected as described in Section ``Data'' 
in the main manuscript.
The optimal dimension of all the hidden layers is 256, and the optimal dimension $d_p$ of shape latent embedding \shapehiddenmat is 128.
The optimal number of points $|\pc|$ in the point cloud \pc is 512.
We sampled 1,024 query points in $\mathcal{Z}$ for each molecule shape.
We constructed graphs from point clouds, which are employed to learn $\shapehiddenmat$ with VN-DGCNN layer (Eq.~\ref{eqn:shape_embed}), using the $k$-nearest neighbors based on Euclidean distance with $k=20$.
We set the number of VN-DGCNN layers as 4.
We set the number of MLP layers in the decoder (Eq.~\ref{eqn:se:decoder}) as 2.
We set the number of VN-In layers as 1.

We optimized the \SE model via Adam~\cite{adam} with its parameters (0.950, 0.999), 
learning rate 0.001, and batch size 16.
We evaluated the validation loss every 2,000 training steps.
We scheduled to decay the learning rate with a factor of 0.6 and a minimum learning rate of 1e-6 if 
the validation loss does not decrease in 5 consecutive evaluations.
The optimal \SE model has 28.3K learnable parameters. 
We trained the \SE model 
with $\sim$156,000 training steps.
The training took 80 hours with our GPUs.
The trained \SE model achieved the minimum validation loss at 152,000 steps.

\input{tables/hyper_para_se}
\input{tables/hyper_para_diff}

\subsection{Parameters of \methoddiff}

Table~\ref{tbl:hyper_diff} presents the parameters used to train \methoddiff.
In \methoddiff, we set the hidden dimensions of all the MLP layers and the scalar hidden layers in GVPs (Eq.~\ref{eqn:pred:gvp} and Eq.~\ref{eqn:mess:gvp}) as 128. 
We set the dimensions of all the vector hidden layers in GVPs as 32.
We set the number of layers $L$ in \molpred as 8.
%
Both two GVP modules in Eq.~\ref{eqn:pred:gvp} and Eq.~\ref{eqn:mess:gvp} consist of three GVP layers. 
We set the number of VN-MLP layers in Eq.~\ref{eqn:shaper} as 1 and the number of MLP layers as 2 for all the involved MLP functions.

We constructed graphs from atoms in molecules, which are employed to learn the scalar embeddings and vector embeddings for atoms 
(Eq.~\ref{eqn:geometric_embedding} and \ref{eqn:attention}), using the $N$-nearest neighbors based on Euclidean distance with $N=8$. 
We used $K=15$ atom features in total, indicating the atom types and its aromaticity.
These atom features include 10 non-aromatic atoms (i.e., ``H'', ``C'', ``N'', ``O'', ``F'', ``P'', ``S'', ``Cl'', ``Br'', ``I''), 
and 5 aromatic atoms (i.e., ``C'', ``N'', ``O'', ``P'', ``S'').
We set the number of diffusion steps $T$ as 1,000.
We set the weight $\xi$ of atom type loss (Eq.~\ref{eqn:loss}) as $100$ to balance the values of atom type loss and atom coordinate loss.
We set the threshold $\delta$ (Eq.~\ref{eqn:diff:obj:pos}) as 10.
The parameters $\beta_t^{\mathtt{x}}$ and $\beta_t^{\mathtt{v}}$ of variance scheduling in the forward diffusion process of \methoddiff are discussed in 
Supplementary Section~\ref{supp:forward:variance}.
%
%
Following \squid, we did not perform extensive hyperparameter tunning for \methoddiff given that the used 
hyperparameters have enabled good performance.

We optimized the \methoddiff model via Adam~\cite{adam} with its parameters (0.950, 0.999), learning rate 0.001, and batch size 32.
We evaluated the validation loss every 2,000 training steps.
We scheduled to decay the learning rate with a factor of 0.6 and a minimum learning rate of 1e-5 if 
the validation loss does not decrease in 10 consecutive evaluations.
The \methoddiff model has 7.8M learnable parameters. 
We trained the \methoddiff model 
with $\sim$770,000 training steps.
The training took 70 hours with our GPUs.
The trained \methoddiff achieved the minimum validation loss at 758,000 steps.

During inference, 
following Adams and Coley~\cite{adams2023equivariant}, we set the variance $\phi$ 
of atom-centered Gaussians as 0.049, which is used to build a set of points for shape guidance in Section ``\method with Shape Guidance'' 
in the main manuscript.
We determined the number of atoms in the generated molecule using the atom number distribution of training molecules that have surface shape sizes similar to the condition molecule.
The optimal distance threshold $\gamma$ is 0.2, and the optimal stop step $S$ for shape guidance is 300.
With shape guidance, each time we updated the atom position (Eq.~\ref{eqn:shape_guidance}), we randomly sampled the weight $\sigma$ from $[0.2, 0.8]$. 
Moreover, when using pocket guidance as mentioned in Section ``\method with Pocket Guidance'' in the main manuscript, each time we updated the atom position (Eq.~\ref{eqn:pocket_guidance}), we randomly sampled the weight $\epsilon$ from $[0, 0.5]$. 
For each condition molecule, it took around 40 seconds on average to generate 50 molecule candidates with our GPUs.

\section{Performance of \decompdiff with Protein Pocket Prior}
\label{supp:app:decompdiff}

In this section, we demonstrate that \decompdiff with protein pocket prior, referred to as \decompdiffbeta, shows very limited performance in generating drug-like and synthesizable molecules compared to all the other methods, including \methodwithpguide and \methodwithsandpguide.
We evaluate the performance of \decompdiffbeta in terms of binding affinities, drug-likeness, and diversity.
We compare \decompdiffbeta with \methodwithpguide and \methodwithsandpguide and report the results in Table~\ref{tbl:comparison_results_decompdiff}.
Note that the results of \methodwithpguide and \methodwithsandpguide here are consistent with those in Table~\ref{tbl:overall_results_docking2} in the main manuscript.
As shown in Table~\ref{tbl:comparison_results_decompdiff}, while \decompdiffbeta achieves high binding affinities in Vina M and Vina D, it substantially underperforms \methodwithpguide and \methodwithsandpguide in QED and SA.
Particularly, \decompdiffbeta shows a QED score of 0.36, while \methodwithpguide substantially outperforms \decompdiffbeta in QED (0.77) with 113.9\% improvement.
\decompdiffbeta also substantially underperforms \methodwithpguide in terms of SA scores (0.55 vs 0.76).
These results demonstrate the limited capacity of \decompdiffbeta in generating drug-like and synthesizable molecules.
As a result, the generated molecules from \decompdiffbeta can have considerably lower utility compared to other methods.
Considering these limitations of \decompdiffbeta, we exclude it from the baselines for comparison.

\input{tables/decompdiff_method_compare}

\section{{Additional Experimental Results on SMG}}
\label{supp:app:results}

\subsection{Comparison on Shape and Graph Similarity}
\label{supp:app:results:overall_shape}


\input{tables/overall_results_sims}

{We evaluate the shape similarity \shapesim and graph similarity \graphsim of molecules generated from}
\dataset, \squid, \method and \methodwithsguide under different graph similarity constraints  ($\delta_g$=1.0, 0.7, 0.5, 0.3). 
%
%
We calculate evaluation metrics using all the generated molecules satisfying the graph similarity constraints.
Particularly, when $\delta_g$=1.0, we do not filter out any molecules based on the constraints and directly calculate metrics on all the generated molecules.
When $\delta_g$=0.7, 0.5 or 0.3, we consider only generated molecules with similarities lower than $\delta_g$.
Based on \shapesim and \graphsim as described in Section ``Evaluation Metrics'' in the main manuscript,
we calculate the following metrics using the subset of molecules with \graphsim lower than $\delta_g$, from a set of 50 generated molecules for each test molecule and report the average of  these metrics across all test molecules:
(1) \avgshapesim\ measures the average \shapesim across each subset of generated molecules with $\graphsim$ lower than $\delta_g$; 
(2) \avggraphsim\ calculates the average \graphsim for each set; 
(3) \maxshapesim\ determines the maximum \shapesim within each set; 
(4) \maxgraphsim\ measures the \graphsim of the molecule with maximum \shapesim in each set. 

As shown in Table~\ref{tbl:overall_sim}, \method and \methodwithsguide demonstrate outstanding performance in terms of the average shape similarities (\avgshapesim) and the average graph similarities (\avggraphsim) among generated molecules.
%
%
Specifically, when $\delta_g$=0.3, \methodwithsguide achieves a substantial 10.5\% improvement in \avgshapesim\ over the best baseline \dataset. 
In terms of \avggraphsim, \methodwithsguide also achieves highly comparable performance with \dataset (0.217 vs 0.211, in \avggraphsim, lower values indicate better performance).
%
This trend remains consistent when applying various similarity constraints (i.e., $\delta_g$) as shown in Table~\ref{tbl:overall_sim}.

Similarly, \method and \methodwithsguide demonstrate superior performance in terms of the average maximum shape similarity across generated molecules for all test molecules (\maxshapesim), as well as the average graph similarity of the molecules with the maximum shape similarities (\maxgraphsim). 
%
%
Specifically, at \maxshapesim, Table~\ref{tbl:overall_sim} shows that \methodwithsguide outperforms the best baseline \squid ($\lambda$=0.3) when $\delta_g$=0.3, 0.5, and 0.7, and only underperforms
it by 0.7\% when $\delta$=1.0.
We also note that the molecules generated by {\methodwithsguide} with the maximum shape similarities have substantially lower graph similarities ({\maxgraphsim}) compared to those generated by {\squid} ({$\lambda$}=0.3).
%
%
%
%
As evidenced by these results, \methodwithsguide features strong capacities of generating molecules with similar shapes yet novel graph structures compared to the condition molecule, facilitating the discovery of promising drug candidates.

Table~\ref{tbl:overall_sim} also shows that by incorporating shape guidance, \methodwithsguide
substantially outperforms \method in both \avgshapesim and \maxshapesim, while maintaining comparable graph similarities (i.e., \avggraphsim\ and \maxgraphsim).
Particularly, when $\delta_g$=0.3, \methodwithsguide 
establishes a considerable improvement of 6.9\% and 4.9\%
over \method in \avgshapesim and \maxshapesim, respectively. 
%
Meanwhile, \methodwithsguide achieves the same \avggraphsim with \method and only slightly underperforms \method in \maxgraphsim (0.223 vs 0.220).
%
%
%
%
%
%
The superior performance of \methodwithsguide suggests that the incorporation of shape guidance effectively boosts the shape similarities of generated molecules without compromising graph similarities.
%

\subsection{Comparison on Validity and Novelty}
\label{supp:app:results:valid_novel}

We evaluate the ability of \method and \squid to generate molecules with valid and novel 2D molecular graphs.
We calculate the percentages of the valid and novel molecules among all the generated molecules.
As shown in Table~\ref{tbl:validity_novelty}, both \method and \methodwithsguide outperform \squid with $\lambda$=0.3 and $\lambda$=1.0 in generating novel molecules.
Particularly, almost all valid molecules generated by \method and \methodwithsguide are novel (99.8\% and 99.9\% at \#n\%), while the best baseline \squid with $\lambda$=0.3 achieves 98.4\% in novelty.
In terms of the percentage of valid and novel molecules among all the generated ones (\#v\&n\%), \method and \methodwithsguide again outperform \squid with $\lambda$=0.3 and $\lambda$=1.0.
We also note that at \#v\%,  \method (99.1\%) and \methodwithsguide (99.2\%) slightly underperform \squid with $\lambda$=0.3 and $\lambda$=1.0 (100.0\%) in generating valid molecules.
\squid guarantees the validity of generated molecules by incorporating valence rules into the generation process and ensuring it to avoid fragments that violate these rules.
Conversely, \method and \methodwithsguide use a purely data-driven approach to learn the generation of valid molecules.
These results suggest that, even without integrating valence rules, \method and \methodwithsguide can still achieve a remarkably high percentage of valid and novel generated molecules.

\input{tables/validity_novelty}

\subsection{Additional Quality Comparison between Desirable Molecules Generated by \method and \squid}
\label{supp:app:results:quality_desirable}

\input{tables/overall_results_quality0.5}

\input{tables/overall_results_quality0.7}

\input{tables/overall_results_quality1.0}

Similar to Table~\ref{tbl:overall_results_quality_desired} in the main manuscript, we present the performance comparison on the quality of desirable molecules generated by different methods under different graph similarity constraints $\delta_g$=0.5, 0.7 and 1.0, as detailed in Table~\ref{tbl:overall_results_quality_05}, Table~\ref{tbl:overall_results_quality_07}, and Table~\ref{tbl:overall_results_quality_10}, respectively.
Overall, these tables show that under varying graph similarity constraints, \method and \methodwithsguide can always generate desirable molecules with comparable quality to baselines in terms of stability, 3D structures, and 2D structures.
These results demonstrate the strong effectiveness of \method and \methodwithsguide in generating high-quality desirable molecules with stable and realistic structures in both 2D and 3D.
This enables the high utility of \method and \methodwithsguide in discovering promising drug candidates.

\section{Additional Experimental Results on PMG}
\label{supp:app:results_PMG}



In this section, we present the results of \methodwithpguide and \methodwithsandpguide when generating 100 molecules. 
Please note that both \methodwithpguide and \methodwithsandpguide show remarkable efficiency over the PMG baselines.
\methodwithpguide and \methodwithsandpguide generate 100 molecules in 48 and 58 seconds on average, respectively, while the most efficient baseline \targetdiff requires 1,252 seconds.
We report the performance of \methodwithpguide and \methodwithsandpguide against state-of-the-art PMG baselines in Table~\ref{tbl:overall_results_docking_100}.

According to Table~\ref{tbl:overall_results_docking_100}, \methodwithpguide and \methodwithsandpguide achieve comparable performance with the PMG baselines in generating molecules with high binding affinities.
Particularly, in terms of Vina S, \methodwithsandpguide achieves very comparable performance (-4.56 kcal/mol) to the third-best baseline \decompdiff (-4.58 kcal/mol) in average Vina S; it also achieves the third-best performance (-4.82 kcal/mol) among all the methods and slightly underperforms the second-best baseline \AR (-4.99 kcal/mol) in median Vina S
\methodwithsandpguide also achieves very close average Vina M (-5.53 kcal/mol) with the third-best baseline \AR (-5.59 kcal/mol) and the third-best performance (-5.47 kcal/mol) in median Vina M.
Notably, for Vina D, \methodwithpguide and \methodwithsandpguide achieve the second and third performance among all the methods.
In terms of the average percentage of generated molecules with Vina D higher than those of known ligands (i.e., HA), \methodwithpguide (58.52\%) and \methodwithsandpguide (58.28\%) outperform the best baseline \targetdiff (57.57\%).
These results signify the high utility of \methodwithpguide and \methodwithsandpguide in generating molecules that effectively bind with protein targets and have better binding affinities than known ligands.

In addition to binding affinities, \methodwithpguide and \methodwithsandpguide also demonstrate similar performance compared to the baselines in metrics related to drug-likeness and diversity.
For drug-likeness, both \methodwithpguide and \methodwithsandpguide achieve the best (0.67) and the second-best (0.66) QED scores.
They also achieve the third and fourth performance in SA scores.
In terms of the diversity among generated molecules,  \methodwithpguide and \methodwithsandpguide slightly underperform the baselines, possibly due to the design that generates molecules with similar shapes to the ligands.
These results highlight the strong ability of \methodwithpguide and \methodwithsandpguide in efficiently generating effective binding molecules with favorable drug-likeness and diversity.
This ability enables them to potentially serve as promising tools to facilitate effective and efficient drug development.

\input{tables/overall_results_docking4}

\section{Properties of Molecules in Case Studies for Targets}
\label{supp:app:results:properties}

\subsection{Drug Properties of Generated Molecules}
\label{supp:app:results:properties:drug}

Table~\ref{tbl:drug_property} presents the drug properties of three generated molecules: NL-001, NL-002, and NL-003.
As shown in Table~\ref{tbl:drug_property}, each of these molecules has a favorable profile, making them promising drug candidates. 
{As discussed in Section ``Case Studies for Targets'' in the main manuscript, all three molecules have high binding affinities in terms of Vina S, Vina M and Vina D, and favorable QED and SA values.
In addition, all of them meet the Lipinski's rule of five criteria~\cite{Lipinski1997}.}
In terms of physicochemical properties, all these properties of NL-001, NL-002 and NL-003, including number of rotatable bonds, molecule weight, LogP value, number of hydrogen bond doners and acceptors, and molecule charges, fall within the desired range of drug molecules. 
This indicates that these molecules could potentially have good solubility and membrane permeability, essential qualities for effective drug absorption.

These generated molecules also demonstrate promising safety profiles based on the predictions from ICM~\cite{Neves2012}.
In terms of drug-induced liver injury prediction scores, all three molecules have low scores (0.188 to 0.376), indicating a minimal risk of hepatotoxicity. 
NL-001 and NL-002 fall under `Ambiguous/Less concern' for liver injury, while NL-003 is categorized under 'No concern' for liver injury. 
Moreover, all these molecules have low toxicity scores (0.000 to 0.236). 
NL-002 and NL-003 do not have any known toxicity-inducing functional groups. 
NL-001 and NL-003 are also predicted not to include any known bad groups that lead to inappropriate features.
These attributes highlight the potential of NL-001, NL-002, and NL-003 as promising treatments for cancers and Alzheimer’s disease.


\input{tables/drug_property_generated_mols}

\subsection{Comparison on ADMET Profiles between Generated Molecules and Approved Drugs}
\label{supp:app:results:properties:admet}

\paragraph{Generated Molecules for CDK6}
Table~\ref{tbl:admet_cdk6} presents the comparison on ADMET profiles between two generated molecules for CDK6 and the approved CDK6 inhibitors, including Abemaciclib~\cite{Patnaik2016}, Palbociclib~\cite{Lu2015}, and Ribociclib~\cite{Tripathy2017}.
As shown in Table~\ref{tbl:admet_cdk6}, the generated molecules, NL-001 and NL-002, exhibit comparable ADMET profiles with those of the approved CDK6 inhibitors. 
Importantly, both molecules demonstrate good potential in most crucial properties, including Ames mutagenesis, favorable oral toxicity, carcinogenicity, estrogen receptor binding, high intestinal absorption and favorable oral bioavailability.
Although the generated molecules are predicted as positive in hepatotoxicity and mitochondrial toxicity, all the approved drugs are also predicted as positive in these two toxicity.
This result suggests that these issues might stem from the limited prediction accuracy rather than being specific to our generated molecules.
Notably, NL-001 displays a potentially better plasma protein binding score compared to other molecules, which may improve its distribution within the body. 
Overall, these results indicate that NL-001 and NL-002 could be promising candidates for further drug development.

\input{tables/admet_property_cdk6}

\paragraph{Generated Molecule for NEP}
Table~\ref{tbl:admet_nep} presents the comparison on ADMET profiles between a generated molecule for NEP targeting Alzheimer's disease and three approved drugs, Donepezil, Galantamine, and Rivastigmine, for Alzheimer's disease~\cite{Hansen2008}.
Overall, NL-003 exhibits a comparable ADMET profile with the three approved drugs.  
Notably, same as other approved drugs, NL-003 is predicted to be able to penetrate the blood brain barrier, a crucial property for Alzheimer's disease.
In addition, it demonstrates a promising safety profile in terms of Ames mutagenesis, favorable oral toxicity, carcinogenicity, estrogen receptor binding, high intestinal absorption, nephrotoxicity and so on.
These results suggest that NL-003 could be promising candidates for the drug development of Alzheimer's disease.

\input{tables/admet_property_nep}

\clearpage
\section{Algorithms}
\label{supp:algorithms}

Algorithm~\ref{alg:shapemol} describes the molecule generation process of \method.
Given a known ligand \molx, \method generates a novel molecule \moly that has a similar shape to \molx and thus potentially similar binding activity.
\method can also take the protein pocket \pocket as input and adjust the atoms of generated molecules for optimal fit and improved binding affinities.
Specifically, \method first calculates the shape embedding \shapehiddenmat for \molx using the shape encoder \SEE described in Algorithm~\ref{alg:see_shaperep}.
Based on \shapehiddenmat, \method then generates a novel molecule with a similar shape to \molx using the diffusion-based generative model \methoddiff as in Algorithm~\ref{alg:diffgen}.
During generation, \method can use shape guidance to directly modify the shape of \moly to closely resemble the shape of \molx.
When the protein pocket \pocket is available, \method can also use pocket guidance to ensure that \moly is specifically tailored to closely fit within \pocket.

\input{algorithms/shapemol}

\input{algorithms/shaperep}

\input{algorithms/diffgen}



\section{{Equivariance and Invariance}}
\label{supp:ei}

\subsection{Equivariance}
\label{supp:ei:equivariance}

{Equivariance refers to the property of a function $f(\pos)$ 
that any translation and rotation transformation from the special Euclidean group SE(3)~\cite{Atz2021} applied to a geometric object
$\pos\in\mathbb{R}^3$ is mirrored in the output of $f(\pos)$, accordingly.
This property ensures $f(\pos)$ to learn a consistent representation of an object's geometric information, regardless of its orientation or location in 3D space.
%
%
Formally, given any translation transformation $\mathbf{t}\in\mathbb{R}^3$ and rotation transformation $\mathbf{R}\in\mathbb{R}^{3\times3}$ ($\mathbf{R}^{\mathsf{T}}\mathbf{R}=\mathbb{I}$), 
$f(\pos)$ is equivariant with respect to these transformations 
if it satisfies
\begin{equation}
f(\mathbf{R}\pos+\mathbf{t}) = \mathbf{R}f(\pos) + \mathbf{t}. 
\end{equation}
%
%
In \method, both \SE and \methoddiff are developed to guarantee equivariance in capturing the geometric features of objects regardless of any translation or rotation transformations, as will be detailed in the following sections.
}

\subsection{Invariance}
\label{supp:ei:invariance}

Invariance refers to the property of a function that its output {$f(\pos)$} remains constant under any translation and rotation transformations of the input $\pos$. 
This property enables $f(\pos)$ to accurately capture 
the inherent features (e.g., atom features for 3D molecules) that are invariant of its orientation or position in 3D space.
Formally, $f(\pos)$ is invariant under any translation $\mathbf{t}$ and  rotation $\mathbf{R}$ if it satisfies
\begin{equation}
f(\mathbf{R}\pos+\mathbf{t}) = f(\pos).
\end{equation}
In \method, both \SE and \methoddiff capture the inherent features of objects in an invariant way, regardless of any translation or rotation transformations, as will be detailed in the following sections.

\section{Point Cloud Construction}
\label{supp:point_clouds}

In \method, we represented molecular surface shapes using point clouds (\pc).
$\pc$
serves as input to \SE, from which we derive shape latent embeddings.
To generate $\pc$, 
we initially generated a molecular surface mesh using the algorithm from the Open Drug Discovery Toolkit~\cite{Wjcikowski2015oddt}.
Following this, we uniformly sampled points on the mesh surface with probability proportional to the face area, 
using the algorithm from PyTorch3D~\cite{ravi2020pytorch3d}.
This point cloud $\pc$ is then centralized by setting the center of its points to zero.
%
%

\section{Query Point Sampling}
\label{supp:training:shapeemb}

As described in Section ``Shape Decoder (\SED)'', the signed distances of query points $z_q$ to molecule surface shape $\pc$ are used to optimize \SE.
In this section, we present how to sample these points $z_q$ in 3D space.
Particularly, we first determined the bounding box around the molecular surface shape, using the maximum and minimum \mbox{($x$, $y$, $z$)-axis} coordinates for points in our point cloud \pc,
denoted as $(x_\text{min}, y_\text{min}, z_\text{min})$ and $(x_\text{max}, y_\text{max}, z_\text{max})$.
We extended this box slightly by defining its corners as \mbox{$(x_\text{min}-1, y_\text{min}-1, z_\text{min}-1)$} and \mbox{$(x_\text{max}+1, y_\text{max}+1, z_\text{max}+1)$}.
For sampling $|\mathcal{Z}|$ query points, we wanted an even distribution of points inside and outside the molecule surface shape.
%
%
When a bounding box is defined around the molecule surface shape, there could be a lot of empty spaces within the box that the molecule does not occupy due to 
its complex and irregular shape.
This could lead to that fewer points within the molecule surface shape could be sampled within the box.
Therefore, we started by randomly sampling $3k$ points within our bounding box to ensure that there are sufficient points within the surface.
We then determined whether each point lies within the molecular surface, using an algorithm from Trimesh~\footnote{https://trimsh.org/} based on the molecule surface mesh.
If there are $n_w$ points found within the surface, we selected $n=\min(n_w, k/2)$ points from these points, 
and randomly choose the remaining 
$k-n$ points 
from those outside the surface.
For each query point, we determined its signed distance to the molecule surface by its closest distance to points in \pc with a sign indicating whether it is inside the surface.

\section{Forward Diffusion (\diffnoise)}
\label{supp:forward}

\subsection{{Forward Process}}
\label{supp:forward:forward}

Formally, for atom positions, the probability of $\pos_t$ sampled given $\pos_{t-1}$, denoted as $q(\pos_t|\pos_{t-1})$, is defined as follows,
%
\begin{equation}
q(\pos_t|\pos_{t-1}) = \mathcal{N}(\pos_t|\sqrt{1-\beta^{\mathtt{x}}_t}\pos_{t-1}, \beta^{\mathtt{x}}_t\mathbb{I}), 
\label{eqn:noiseposinter}
\end{equation}
%
where 
$\mathcal{N}(\cdot)$ is a Gaussian distribution of $\pos_t$ with mean $\sqrt{1-\beta_t^{\mathtt{x}}}\pos_{t-1}$ and covariance $\beta_t^{\mathtt{x}}\mathbf{I}$.
%
Following Hoogeboom \etal~\cite{hoogeboom2021catdiff}, 
%
for atom features, the probability of $\atomfeat_t$ across $K$ classes given $\atomfeat_{t-1}$ is defined as follows,
\begin{equation}
q(\atomfeat_t|\atomfeat_{t-1}) = \mathcal{C}(\atomfeat_t|(1-\beta^{\mathtt{v}}_t) \atomfeat_{t-1}+\beta^{\mathtt{v}}_t\mathbf{1}/K),
\label{eqn:noisetypeinter}
\end{equation}
where 
$\mathcal{C}$ is a categorical distribution of $\atomfeat_t$ derived from the 
noising $\atomfeat_{t-1}$ with a uniform noise $\beta^{\mathtt{v}}_t\mathbf{1}/K$ across $K$ classes.

Since the above distributions form Markov chains, 
the probability of any $\pos_t$ or $\atomfeat_t$ can be derived from $\pos_0$ or $\atomfeat_0$:
%
\begin{eqnarray}
& q(\pos_t|\pos_{0}) & = \mathcal{N}(\pos_t|\sqrt{\cumalpha^{\mathtt{x}}_t}\pos_0, (1-\cumalpha^{\mathtt{x}}_t)\mathbb{I}), \label{eqn:noisepos}\\
& q(\atomfeat_t|\atomfeat_0)  & = \mathcal{C}(\atomfeat_t|\cumalpha^{\mathtt{v}}_t\atomfeat_0 + (1-\cumalpha^{\mathtt{v}}_t)\mathbf{1}/K), \label{eqn:noisetype}\\
& \text{where }\cumalpha^{\mathtt{u}}_t & = \prod\nolimits_{\tau=1}^{t}\alpha^{\mathtt{u}}_\tau, \ \alpha^{\mathtt{u}}_\tau=1 - \beta^{\mathtt{u}}_\tau, \ {\mathtt{u}}={\mathtt{x}} \text{ or } {\mathtt{v}}.\;\;\;\label{eqn:noiseschedule}
\label{eqn:pos_prior}
\end{eqnarray}
%
%
Note that $\bar{\alpha}^{\mathtt{u}}_t$ ($\mathtt{u}={\mathtt{x}}\text{ or }{\mathtt{v}}$)
is monotonically decreasing from 1 to 0 over $t=[1,T]$. 
As $t\rightarrow 1$, $\cumalpha^{\mathtt{x}}_t$ and $\cumalpha^{\mathtt{v}}_t$ are close to 1, leading to that $\pos_t$ or $\atomfeat_t$ approximates 
$\pos_0$ or $\atomfeat_0$.
Conversely, as  $t\rightarrow T$, $\cumalpha^{\mathtt{x}}_t$ and $\cumalpha^{\mathtt{v}}_t$ are close to 0,
leading to that $q(\pos_T|\pos_{0})$ 
resembles  {$\mathcal{N}(\mathbf{0}, \mathbb{I})$} 
and $q(\atomfeat_T|\atomfeat_0)$ 
resembles {$\mathcal{C}(\mathbf{1}/K)$}.

Using Bayes theorem, the ground-truth Normal posterior of atom positions $p(\pos_{t-1}|\pos_t, \pos_0)$ can be calculated in a
closed form~\cite{ho2020ddpm} as below,
\begin{eqnarray}
& p(\pos_{t-1}|\pos_t, \pos_0) = \mathcal{N}(\pos_{t-1}|\mu(\pos_t, \pos_0), \tilde{\beta}^\mathtt{x}_t\mathbb{I}), \label{eqn:gt_pos_posterior_1}\\
&\!\!\!\!\!\!\!\!\!\!\!\mu(\pos_t, \pos_0)\!=\!\frac{\sqrt{\bar{\alpha}^{\mathtt{x}}_{t-1}}\beta^{\mathtt{x}}_t}{1-\bar{\alpha}^{\mathtt{x}}_t}\pos_0\!+\!\frac{\sqrt{\alpha^{\mathtt{x}}_t}(1-\bar{\alpha}^{\mathtt{x}}_{t-1})}{1-\bar{\alpha}^{\mathtt{x}}_t}\pos_t, 
\tilde{\beta}^\mathtt{x}_t\!=\!\frac{1-\bar{\alpha}^{\mathtt{x}}_{t-1}}{1-\bar{\alpha}^{\mathtt{x}}_{t}}\beta^{\mathtt{x}}_t.\;\;\;
\end{eqnarray}
%
Similarly, the ground-truth categorical posterior of atom features $p(\atomfeat_{t-1}|\atomfeat_{t}, \atomfeat_0)$ can be calculated~\cite{hoogeboom2021catdiff} as below,
\begin{eqnarray}
& p(\atomfeat_{t-1}|\atomfeat_{t}, \atomfeat_0) = \mathcal{C}(\atomfeat_{t-1}|\mathbf{c}(\atomfeat_t, \atomfeat_0)), \label{eqn:gt_atomfeat_posterior_1}\\
& \mathbf{c}(\atomfeat_t, \atomfeat_0) = \tilde{\mathbf{c}}/{\sum_{k=1}^K \tilde{c}_k}, \label{eqn:gt_atomfeat_posterior_2} \\
& \tilde{\mathbf{c}} = [\alpha^{\mathtt{v}}_t\atomfeat_t + \frac{1 - \alpha^{\mathtt{v}}_t}{K}]\odot[\bar{\alpha}^{\mathtt{v}}_{t-1}\atomfeat_{0}+\frac{1-\bar{\alpha}^{\mathtt{v}}_{t-1}}{K}], 
\label{eqn:gt_atomfeat_posterior_3}
\end{eqnarray}
%
%
where $\tilde{c}_k$ denotes the likelihood of $k$-th class across $K$ classes in $\tilde{\mathbf{c}}$; 
$\odot$ denotes the element-wise product operation;
$\tilde{\mathbf{c}}$ is calculated using $\atomfeat_t$ and $\atomfeat_{0}$ and normalized into $\mathbf{c}(\atomfeat_t, \atomfeat_0)$ so as to represent
probabilities. 
%
The proof of the above equations is available in Supplementary Section~\ref{supp:forward:proof}.

\subsection{Variance Scheduling in \diffnoise}
\label{supp:forward:variance}

Following Guan \etal~\cite{guan2023targetdiff}, we used a sigmoid $\beta$ schedule for the variance schedule $\beta_t^{\mathtt{x}}$ of atom coordinates as below,

\begin{equation}
\beta_t^{\mathtt{x}} = \text{sigmoid}(w_1(2 t / T - 1)) (w_2 - w_3) + w_3
\end{equation}
in which $w_i$($i$=1,2, or 3) are hyperparameters; $T$ is the maximum step.
We set $w_1=6$, $w_2=1.e-7$ and $w_3=0.01$.
For atom types, we used a cosine $\beta$ schedule~\cite{nichol2021} for $\beta_t^{\mathtt{v}}$ as below,

\begin{equation}
\begin{aligned}
& \bar{\alpha}_t^{\mathtt{v}} = \frac{f(t)}{f(0)}, f(t) = \cos(\frac{t/T+s}{1+s} \cdot \frac{\pi}{2})^2\\
& \beta_t^{\mathtt{v}} = 1 - \alpha_t^{\mathtt{v}} = 1 - \frac{\bar{\alpha}_t^{\mathtt{v}} }{\bar{\alpha}_{t-1}^{\mathtt{v}} }
\end{aligned}
\end{equation}
in which $s$ is a hyperparameter and set as 0.01.

\begin{figure}
	\begin{subfigure}[t]{.45\linewidth}
		\centering
		\includegraphics[width=.7\linewidth]{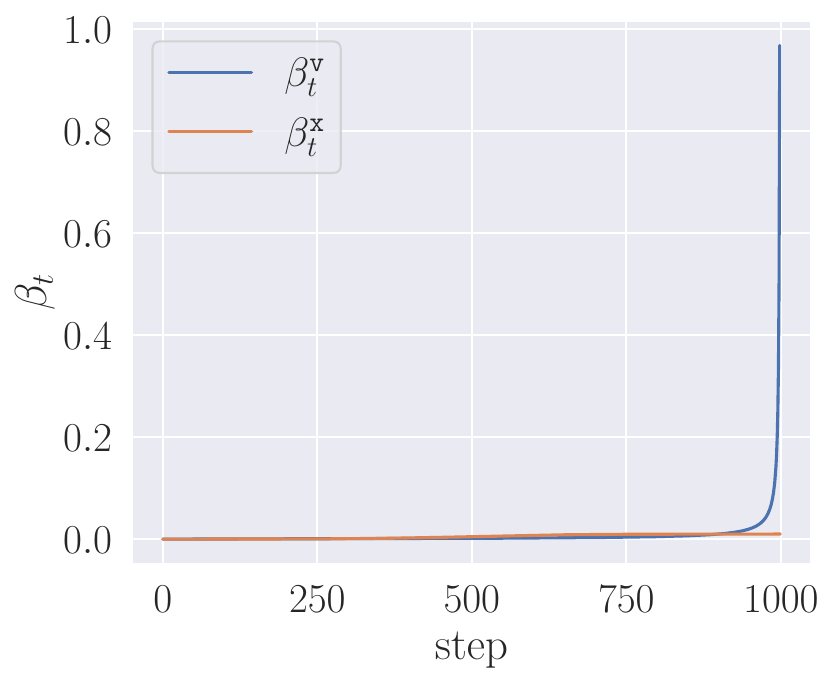}
	\end{subfigure}
	\hfill
	\begin{subfigure}[t]{.45\linewidth}
		\centering
		\includegraphics[width=.7\linewidth]{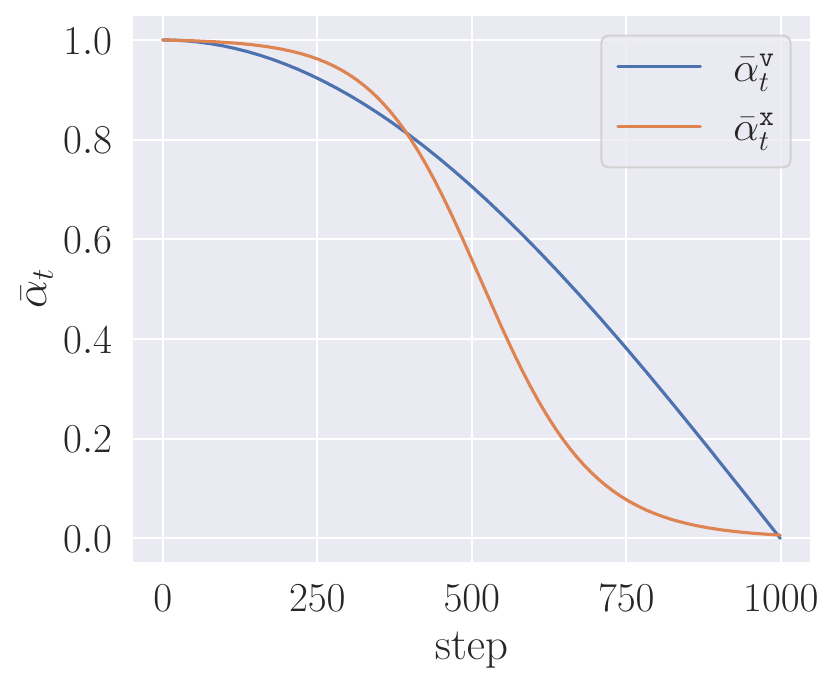}
	\end{subfigure}
	\caption{Schedule}
	\label{fig:schedule}
\end{figure}

As shown in Section ``Forward Diffusion Process'', the values of $\beta_t^{\mathtt{x}}$ and $\beta_t^{\mathtt{v}}$ should be 
sufficiently small to ensure the smoothness of forward diffusion process. In the meanwhile, their corresponding $\bar{\alpha}_t$
values should decrease from 1 to 0 over $t=[1,T]$.
Figure~\ref{fig:schedule} shows the values of $\beta_t$ and $\bar{\alpha}_t$ for atom coordinates and atom types with our hyperparameters.
Please note that the value of $\beta_{t}^{\mathtt{x}}$ is less than 0.1 for 990 out of 1,000 steps. 
This guarantees the smoothness of the forward diffusion process.

\subsection{Derivation of Forward Diffusion Process}
\label{supp:forward:proof}

In \method, a Gaussian noise and a categorical noise are added to continuous atom position and discrete atom features, respectively.
Here, we briefly describe the derivation of posterior equations (i.e., Eq.~\ref{eqn:gt_pos_posterior_1}, and   \ref{eqn:gt_atomfeat_posterior_1}) for atom positions and atom types in our work.
We refer readers to Ho \etal~\cite{ho2020ddpm} and Kong \etal~\cite{kong2021diffwave} 
for a detailed description of diffusion process for continuous variables and Hoogeboom \etal~\cite{hoogeboom2021catdiff} for
the description of diffusion process for discrete variables.

For continuous atom positions, as shown in Kong \etal~\cite{kong2021diffwave}, according to Bayes theorem, given $q(\pos_t|\pos_{t-1})$ defined in Eq.~\ref{eqn:noiseposinter} and 
$q(\pos_t|\pos_0)$ defined in Eq.~\ref{eqn:noisepos}, the posterior $q(\pos_{t-1}|\pos_{t}, \pos_0)$ is derived as below (superscript $\mathtt{x}$ is omitted for brevity),

\begin{equation}
\begin{aligned}
& q(\pos_{t-1}|\pos_{t}, \pos_0)  = \frac{q(\pos_t|\pos_{t-1}, \pos_0)q(\pos_{t-1}|\pos_0)}{q(\pos_t|\pos_0)} \\
& =  \frac{\mathcal{N}(\pos_t|\sqrt{1-\beta_t}\pos_{t-1}, \beta_{t}\mathbf{I}) \mathcal{N}(\pos_{t-1}|\sqrt{\bar{\alpha}_{t-1}}\pos_{0}, (1-\bar{\alpha}_{t-1})\mathbf{I}) }{ \mathcal{N}(\pos_{t}|\sqrt{\bar{\alpha}_t}\pos_{0}, (1-\bar{\alpha}_t)\mathbf{I})}\\
& =  (2\pi{\beta_t})^{-\frac{3}{2}} (2\pi{(1-\bar{\alpha}_{t-1})})^{-\frac{3}{2}} (2\pi(1-\bar{\alpha}_t))^{\frac{3}{2}} \times \exp( \\
& -\frac{\|\pos_t - \sqrt{\alpha}_t\pos_{t-1}\|^2}{2\beta_t} -\frac{\|\pos_{t-1} - \sqrt{\bar{\alpha}}_{t-1}\pos_{0} \|^2}{2(1-\bar{\alpha}_{t-1})} \\
& + \frac{\|\pos_t - \sqrt{\bar{\alpha}_t}\pos_0\|^2}{2(1-\bar{\alpha}_t)}) \\
& = (2\pi\tilde{\beta}_t)^{-\frac{3}{2}} \exp(-\frac{1}{2\tilde{\beta}_t}\|\pos_{t-1}-\frac{\sqrt{\bar{\alpha}_{t-1}}\beta_t}{1-\bar{\alpha}_t}\pos_0 \\
& - \frac{\sqrt{\alpha_t}(1-\bar{\alpha}_{t-1})}{1-\bar{\alpha}_t}\pos_{t}\|^2) \\
& \text{where}\ \tilde{\beta}_t = \frac{1-\bar{\alpha}_{t-1}}{1-\bar{\alpha}_t}\beta_t.
\end{aligned}
\end{equation}

Therefore, the posterior of atom positions is derived as below,

\begin{equation}
q(\pos_{t-1}|\pos_{t}, \pos_0)\!\!=\!\!\mathcal{N}(\pos_{t-1}|\frac{\sqrt{\bar{\alpha}_{t-1}}\beta_t}{1-\bar{\alpha}_t}\pos_0 + \frac{\sqrt{\alpha_t}(1-\bar{\alpha}_{t-1})}{1-\bar{\alpha}_t}\pos_{t}, \tilde{\beta}_t\mathbf{I}).
\end{equation}

For discrete atom features, as shown in Hoogeboom \etal~\cite{hoogeboom2021catdiff} and Guan \etal~\cite{guan2023targetdiff},
according to Bayes theorem, the posterior $q(\atomfeat_{t-1}|\atomfeat_{t}, \atomfeat_0)$ is derived as below (supperscript $\mathtt{v}$ is omitted for brevity),

\begin{equation}
\begin{aligned}
& q(\atomfeat_{t-1}|\atomfeat_{t}, \atomfeat_0) =  \frac{q(\atomfeat_t|\atomfeat_{t-1}, \atomfeat_0)q(\atomfeat_{t-1}|\atomfeat_0)}{\sum_{\scriptsize{\atomfeat}_{t-1}}q(\atomfeat_t|\atomfeat_{t-1}, \atomfeat_0)q(\atomfeat_{t-1}|\atomfeat_0)} \\
\end{aligned}
\end{equation}

For $q(\atomfeat_t|\atomfeat_{t-1}, \atomfeat_0)$, we have 
%
%
\begin{equation}
\begin{aligned}
q(\atomfeat_t|\atomfeat_{t-1}, \atomfeat_0) & = \mathcal{C}(\atomfeat_t|(1-\beta_t)\atomfeat_{t-1} + \beta_t/{K})\\
& = \begin{cases}
1-\beta_t+\beta_t/K,\!&\text{when}\ \atomfeat_{t} = \atomfeat_{t-1},\\
\beta_t / K,\! &\text{when}\ \atomfeat_{t} \neq \atomfeat_{t-1},
\end{cases}\\
& = \mathcal{C}(\atomfeat_{t-1}|(1-\beta_t)\atomfeat_{t} + \beta_t/{K}).
\end{aligned}
\end{equation}
Therefore, we have
%
\begin{equation}
\begin{aligned}
& q(\atomfeat_t|\atomfeat_{t-1}, \atomfeat_0)q(\atomfeat_{t-1}|\atomfeat_0) \\
& = \mathcal{C}(\atomfeat_{t-1}|(1-\beta_t)\atomfeat_{t} + \beta_t\frac{\mathbf{1}}{K}) \mathcal{C}(\atomfeat_{t-1}|\bar{\alpha}_{t-1}\atomfeat_0+(1-\bar{\alpha}_{t-1})\frac{\mathbf{1}}{K}) \\
& = [\alpha_t\atomfeat_t + \frac{1 - \alpha_t}{K}]\odot[\bar{\alpha}_{t-1}\atomfeat_{0}+\frac{1-\bar{\alpha}_{t-1}}{K}].
\end{aligned}
\end{equation}
%
Therefore, with $q(\atomfeat_t|\atomfeat_{t-1}, \atomfeat_0)q(\atomfeat_{t-1}|\atomfeat_0) = \tilde{\mathbf{c}}$, the posterior is as below,

\begin{equation}
q(\atomfeat_{t-1}|\atomfeat_{t}, \atomfeat_0) = \mathcal{C}(\atomfeat_{t-1}|\mathbf{c}(\atomfeat_t, \atomfeat_0)) = \frac{\tilde{\mathbf{c}}}{\sum_{k}^K\tilde{c}_k}.
\end{equation}

\section{{Backward Generative Process} (\diffgenerative)}
\label{supp:backward}

Following Ho \etal~\cite{ho2020ddpm}, with $\tilde{\pos}_{0,t}$, the probability of $\pos_{t-1}$ denoised from $\pos_t$, denoted as $p(\pos_{t-1}|\pos_t)$,
can be estimated 
by the approximated posterior $p_{\boldsymbol{\Theta}}(\pos_{t-1}|\pos_t, \tilde{\pos}_{0,t})$ as below,
\begin{equation}
\begin{aligned}
p(\pos_{t-1}|\pos_t) & \approx p_{\boldsymbol{\Theta}}(\pos_{t-1}|\pos_t, \tilde{\pos}_{0,t}) \\
& = \mathcal{N}(\pos_{t-1}|\mu_{\boldsymbol{\Theta}}(\pos_t, \tilde{\pos}_{0,t}),\tilde{\beta}_t^{\mathtt{x}}\mathbb{I}),
\end{aligned}
\label{eqn:aprox_pos_posterior}
\end{equation}
where ${\boldsymbol{\Theta}}$ is the learnable parameter; $\mu_{\boldsymbol{\Theta}}(\pos_t, \tilde{\pos}_{0,t})$ is an estimate 
of $\mu(\pos_t, \pos_{0})$ by replacing $\pos_0$ with its estimate $\tilde{\pos}_{0,t}$ 
in Equation~{\ref{eqn:gt_pos_posterior_1}}.
Similarly, with $\tilde{\atomfeat}_{0,t}$, the probability of $\atomfeat_{t-1}$ denoised from $\atomfeat_t$, denoted as $p(\atomfeat_{t-1}|\atomfeat_t)$, 
can be estimated 
by the approximated posterior $p_{\boldsymbol{\Theta}}(\atomfeat_{t-1}|\atomfeat_t, \tilde{\atomfeat}_{0,t})$ as below,
\begin{equation}
\begin{aligned}
p(\atomfeat_{t-1}|\atomfeat_t)\approx p_{\boldsymbol{\Theta}}(\atomfeat_{t-1}|\atomfeat_{t}, \tilde{\atomfeat}_{0,t}) 
=\mathcal{C}(\atomfeat_{t-1}|\mathbf{c}_{\boldsymbol{\Theta}}(\atomfeat_t, \tilde{\atomfeat}_{0,t})),\!\!\!\!
\end{aligned}
\label{eqn:aprox_atomfeat_posterior}
\end{equation}
where $\mathbf{c}_{\boldsymbol{\Theta}}(\atomfeat_t, \tilde{\atomfeat}_{0,t})$ is an estimate of $\mathbf{c}(\atomfeat_t, \atomfeat_0)$
by replacing $\atomfeat_0$  
with its estimate $\tilde{\atomfeat}_{0,t}$ in Equation~\ref{eqn:gt_atomfeat_posterior_1}.

\section{\method Loss Function Derivation}
\label{supp:training:loss}

In this section, we demonstrate that a step weight $w_t^{\mathtt{x}}$ based on the signal-to-noise ratio $\lambda_t$ should be 
included into the atom position loss (Eq.~\ref{eqn:diff:obj:pos}).
In the diffusion process for continuous variables, the optimization problem is defined 
as below~\cite{ho2020ddpm},
\begin{equation*}
\begin{aligned}
& \arg\min_{\boldsymbol{\Theta}}KL(q(\pos_{t-1}|\pos_t, \pos_0)|p_{\boldsymbol{\Theta}}(\pos_{t-1}|\pos_t, \tilde{\pos}_{0,t})) \\
& = \arg\min_{\boldsymbol{\Theta}} \frac{\bar{\alpha}_{t-1}(1-\alpha_t)}{2(1-\bar{\alpha}_{t-1})(1-\bar{\alpha}_{t})}\|\tilde{\pos}_{0, t}-\pos_0\|^2 \\
& = \arg\min_{\boldsymbol{\Theta}} \frac{1-\alpha_t}{2(1-\bar{\alpha}_{t-1})\alpha_{t}} \|\tilde{\boldsymbol{\epsilon}}_{0,t}-\boldsymbol{\epsilon}_0\|^2,
\end{aligned}
\end{equation*}
where $\boldsymbol{\epsilon}_0 = \frac{\pos_t - \sqrt{\bar{\alpha}_t}\pos_0}{\sqrt{1-\bar{\alpha}_t}}$ is the ground-truth noise variable sampled from $\mathcal{N}(\mathbf{0}, \mathbf{1})$ and is used to sample $\pos_t$ from $\mathcal{N}(\pos_t|\sqrt{\cumalpha_t}\pos_0, (1-\cumalpha_t)\mathbf{I})$ in Eq.~\ref{eqn:noisetype};
$\tilde{\boldsymbol{\epsilon}}_0 = \frac{\pos_t - \sqrt{\bar{\alpha}_t}\tilde{\pos}_{0, t}}{\sqrt{1-\bar{\alpha}_t}}$ is the predicted noise variable. 

Ho \etal~\cite{ho2020ddpm} further simplified the above objective as below and
demonstrated that the simplified one can achieve better performance:
\begin{equation}
\begin{aligned}
& \arg\min_{\boldsymbol{\Theta}} \|\tilde{\boldsymbol{\epsilon}}_{0,t}-\boldsymbol{\epsilon}_0\|^2 \\
& = \arg\min_{\boldsymbol{\Theta}} \frac{\bar{\alpha}_t}{1-\bar{\alpha}_t}\|\tilde{\pos}_{0,t}-\pos_0\|^2,
\end{aligned}
\label{eqn:supp:losspos}
\end{equation}
where $\lambda_t=\frac{\bar{\alpha}_t}{1-\bar{\alpha}_t}$ is the signal-to-noise ratio.
While previous work~\cite{guan2023targetdiff} applies uniform step weights across
different steps, we demonstrate that a step weight should be included into the atom position loss according to Eq.~\ref{eqn:supp:losspos}.
However, the value of $\lambda_t$ could be very large when $\bar{\alpha}_t$ is close to 1 as $t$ approaches 1.
Therefore, we clip the value of $\lambda_t$ with threshold $\delta$ in Eq.~\ref{eqn:diff:obj:pos}.

%% file: tables/hyper_para_se.tex
\begin{table*}[!h]
  \centering
      \caption{{Hyper-Parameter Space for \SE Optimization}}
  \label{tbl:hyper_se}
  \begin{threeparttable}
 \begin{scriptsize}
      \begin{tabular}{
	@{\hspace{2pt}}l@{\hspace{5pt}} 
	@{\hspace{2pt}}r@{\hspace{2pt}}         
	}
        \toprule
          Hyper-parameters &  Space\\
        \midrule
         hidden layer dimension            & \{128, 256\}\\
         dimension $d_p$ of \shapehiddenmat        &  \{64, 128\} \\
         \#points in \pc        & \{512, 1,024\} \\
         \#query points in $\mathcal{Z}$                & 1,024 \\
         \#nearest neighbors              & 20          \\
         \#VN-DGCNN layers (Eq~\ref{eqn:shape_embed})               & 4            \\
         \#MLP layers in Eq~\ref{eqn:se:decoder} & 4           \\
        \bottomrule
      \end{tabular}
%
%
\end{scriptsize}
  \end{threeparttable}
\end{table*}

%% file: tables/hyper_para_diff.tex
\begin{table*}[!h]
  \centering
      \caption{{Hyper-Parameter Space for \methoddiff Optimization}}
  \label{tbl:hyper_diff}
  \begin{threeparttable}
 \begin{scriptsize}
      \begin{tabular}{
	@{\hspace{2pt}}l@{\hspace{5pt}} 
	@{\hspace{2pt}}r@{\hspace{2pt}}         
	}
        \toprule
          Hyper-parameters &  Space\\
        \midrule
         scalar hidden layer dimension         & 128 \\
         vector hidden layer dimension         & 32 \\
         weight of atom type loss $\xi$ (Eq.~\ref{eqn:loss})  & 100           \\
         threshold of step weight $\delta$ (Eq.~\ref{eqn:diff:obj:pos}) & 10 \\
         \#atom features $K$                   & 15 \\
         \#layers $L$ in \molpred             & 8 \\
         \#nearest neighbors {$N$}  (Eq.~\ref{eqn:geometric_embedding} and \ref{eqn:attention})            & 8          \\
         {\#diffusion steps $T$}                  & 1,000 \\
        \bottomrule
      \end{tabular}
%
%
\end{scriptsize}
  \end{threeparttable}

\end{table*}

%% file: tables/decompdiff_method_compare.tex
\begin{table*}[!h]
	\centering
		\caption{Comparison on PMG among \methodwithpguide, \methodwithsandpguide and \decompdiffbeta}
	\label{tbl:comparison_results_decompdiff}
\begin{threeparttable}
	\begin{scriptsize}
	\begin{tabular}{
		@{\hspace{2pt}}l@{\hspace{2pt}}
		%
		%
		@{\hspace{2pt}}r@{\hspace{2pt}}
		@{\hspace{2pt}}r@{\hspace{2pt}}
		@{\hspace{6pt}}r@{\hspace{6pt}}
		@{\hspace{2pt}}r@{\hspace{2pt}}
		@{\hspace{2pt}}r@{\hspace{2pt}}
		@{\hspace{5pt}}r@{\hspace{5pt}}
		@{\hspace{2pt}}r@{\hspace{2pt}}
		@{\hspace{2pt}}r@{\hspace{2pt}}
		@{\hspace{5pt}}r@{\hspace{5pt}}
		@{\hspace{2pt}}r@{\hspace{2pt}}
	         @{\hspace{2pt}}r@{\hspace{2pt}}
		@{\hspace{5pt}}r@{\hspace{5pt}}
		@{\hspace{2pt}}r@{\hspace{2pt}}
		@{\hspace{2pt}}r@{\hspace{2pt}}
		@{\hspace{5pt}}r@{\hspace{5pt}}
		@{\hspace{2pt}}r@{\hspace{2pt}}
		@{\hspace{2pt}}r@{\hspace{2pt}}
		@{\hspace{5pt}}r@{\hspace{5pt}}
		@{\hspace{2pt}}r@{\hspace{2pt}}
		@{\hspace{2pt}}r@{\hspace{2pt}}
		@{\hspace{5pt}}r@{\hspace{5pt}}
		@{\hspace{2pt}}r@{\hspace{2pt}}
		}
		\toprule
		\multirow{2}{*}{method} & \multicolumn{2}{c}{Vina S$\downarrow$} & & \multicolumn{2}{c}{Vina M$\downarrow$} & & \multicolumn{2}{c}{Vina D$\downarrow$} & & \multicolumn{2}{c}{{HA\%$\uparrow$}}  & & \multicolumn{2}{c}{QED$\uparrow$} & & \multicolumn{2}{c}{SA$\uparrow$} & & \multicolumn{2}{c}{Div$\uparrow$} & 
		& \multirow{2}{*}{time$\downarrow$} \\
	    \cmidrule{2-3}\cmidrule{5-6} \cmidrule{8-9} \cmidrule{11-12} \cmidrule{14-15} \cmidrule{17-18} \cmidrule{20-21}
		& Avg. & Med. &  & Avg. & Med. &  & Avg. & Med. & & Avg. & Med.  & & Avg. & Med.  & & Avg. & Med.  & & Avg. & Med.  & & \\ 
		\midrule
		 \decompdiffbeta             & -4.72 & -4.86 & & \textbf{-6.84} & \textbf{-6.91} & & \textbf{-8.85} & \textbf{-8.90} & &  {72.16} & {72.16} & &  0.36 & 0.36 & &  0.55 & 0.55 & & 0.59 & 0.59 & & 3,549 \\ 
		\methodwithpguide       &  \underline{-5.53} & \underline{-5.64} & & {-6.37} & -6.33 & &  \underline{-7.19} & \underline{-7.52} & &  \underline{78.75} & \textbf{94.00} & &  \textbf{0.77} & \textbf{0.80} & &  \textbf{0.76} & \textbf{0.76} & & 0.63 & 0.66 & & 462 \\
		\methodwithsandpguide   & \textbf{-5.81} & \textbf{-5.96} & &  \underline{-6.50} & \underline{-6.58} & & -7.16 & {-7.51} & &  \textbf{79.92} & \underline{93.00} & &  \underline{0.76} & \underline{0.79} & &  \underline{0.75} & \underline{0.74} & & 0.64 & 0.66 & & 561\\
		\bottomrule
	\end{tabular}%
	\begin{tablenotes}[normal,flushleft]
		\begin{footnotesize}
	\item 
\!\!Columns represent: {``Vina S'': the binding affinities between the initially generated poses of molecules and the protein pockets; 
		``Vina M'': the binding affinities between the poses after local structure minimization and the protein pockets;
		``Vina D'': the binding affinities between the poses determined by AutoDock Vina~\cite{Eberhardt2021} and the protein targets;
		``QED'': the drug-likeness score;
		``SA'': the synthesizability score;
		``Div'': the diversity among generated molecules;
		``time'': the time cost to generate molecules.}
		
		\par
		\par
		\end{footnotesize}
	\end{tablenotes}
	\end{scriptsize}
\end{threeparttable}
  \vspace{-10pt}    
\end{table*}

%% file: tables/overall_results_sims.tex
\begin{table*}[!h]
	\centering
		\caption{Similarity Comparison on SMG}
	\label{tbl:overall_sim}
\begin{threeparttable}
	\begin{scriptsize}
	\begin{tabular}{
		@{\hspace{0pt}}l@{\hspace{8pt}}
		@{\hspace{8pt}}l@{\hspace{8pt}}
		@{\hspace{8pt}}c@{\hspace{8pt}}
		@{\hspace{8pt}}c@{\hspace{8pt}}
	    	@{\hspace{0pt}}c@{\hspace{0pt}}
		@{\hspace{8pt}}c@{\hspace{8pt}}
		@{\hspace{8pt}}c@{\hspace{8pt}}
		%
		}
		\toprule
		$\delta_g$  & method          & \avgshapesim$\uparrow$(std) & \avggraphsim$\downarrow$(std) & & \maxshapesim$\uparrow$(std) & \maxgraphsim$\downarrow$(std)       \\ 
		\midrule
		\multirow{6}{0.059\linewidth}{\hspace{0pt}0.3} & \dataset             & 0.745(0.037)          & \textbf{0.211}(0.026) &  & 0.815(0.039)          & \textbf{0.215}(0.047)      \\ 
			& \squid($\lambda$=0.3) & 0.709(0.076)          & 0.237(0.033)          &  & 0.841(0.070)          & 0.253(0.038)        \\ 
		    & \squid($\lambda$=1.0) & 0.695(0.064)          & \underline{0.216}(0.034)  &  & 0.841(0.056)          & 0.231(0.047)        \\ 
			& \method               & \underline{0.770}(0.039)  & 0.217(0.031)          &  & \underline{0.858}(0.038)  & \underline{0.220}(0.046)  \\ 
			& \methodwithsguide     & \textbf{0.823}(0.029) & 0.217(0.032)          &  & \textbf{0.900}(0.028) & 0.223(0.048)  \\ 
		\midrule
		\multirow{6}{0.059\linewidth}{\hspace{0pt}0.5} & \dataset & 0.750(0.037)          & \textbf{0.225}(0.037) &  & 0.819(0.039)          & \textbf{0.236}(0.070)          \\ 
			& \squid($\lambda$=0.3)  & 0.728(0.072)          & 0.301(0.054)          &  & \underline{0.888}(0.061)  & 0.355(0.088)          \\ 
			& \squid($\lambda$=1.0)  & 0.699(0.063)          & 0.233(0.043)          &  & 0.850(0.057)          & 0.263(0.080)          \\ 
			& \method               & \underline{0.771}(0.039)  & \underline{0.229}(0.043)  &  & 0.862(0.036)          & \textbf{0.236}(0.065) \\ 
			& \methodwithsguide    & \textbf{0.824}(0.029) & \underline{0.229}(0.044)  &  & \textbf{0.903}(0.027) & \underline{0.242}(0.069)  \\ 
		\midrule
		\multirow{6}{0.059\linewidth}{\hspace{0pt}0.7} 
		& \dataset &  0.750(0.037) & \textbf{0.226}(0.038) & & 0.819(0.039) & \underline{0.240}(0.081) \\ 
			& \squid($\lambda$=0.3) &  0.735(0.074)          & 0.328(0.070)          &  & \underline{0.900}(0.062)  & 0.435(0.143)          \\ 
			& \squid($\lambda$=1.0) &  0.699(0.064)          & 0.234(0.045)          &  & 0.851(0.057)          & 0.268(0.090)          \\ 
			& \method               &  \underline{0.771}(0.039)  & \underline{0.229}(0.043)  &  & 0.862(0.036)          & \textbf{0.237}(0.066) \\ 
			& \methodwithsguide     &  \textbf{0.824}(0.029) & 0.230(0.045)          &  & \textbf{0.903}(0.027) & 0.244(0.074)          \\ 
		\midrule
		\multirow{6}{0.059\linewidth}{\hspace{0pt}1.0} 
		& \dataset & 0.750(0.037)          & \textbf{0.226}(0.038) &  & 0.819(0.039)          & \underline{0.242}(0.085)  \\
		& \squid($\lambda$=0.3) & 0.740(0.076)          & 0.349(0.088)          &  & \textbf{0.909}(0.065) & 0.547(0.245)       \\ 
		& \squid($\lambda$=1.0) & 0.699(0.064)          & 0.235(0.045)          &  & 0.851(0.057)          & 0.271(0.097)          \\ 
		& \method               & \underline{0.771}(0.039)  & \underline{0.229}(0.043)  &  & 0.862(0.036)          & \textbf{0.237}(0.066) \\ 
		& \methodwithsguide      & \textbf{0.824}(0.029) & 0.230(0.045)          &  & \underline{0.903}(0.027)  & 0.244(0.076)          \\ 
		\bottomrule
	\end{tabular}%
	\begin{tablenotes}[normal,flushleft]
		\begin{footnotesize}
	\item 
\!\!Columns represent: ``$\delta_g$'': the graph similarity constraint; 
``\avgshapesim/\avggraphsim'': the average of shape or graph similarities between the condition molecules and generated molecules with $\graphsim<=\delta_g$;
``\maxshapesim'': the maximum of shape similarities between the condition molecules and generated molecules with $\graphsim<=\delta_g$;
``\maxgraphsim'': the graph similarities between the condition molecules and the molecules with the maximum shape similarities and $\graphsim<=\delta_g$;
%
``$\uparrow$'' represents higher values are better, and ``$\downarrow$'' represents lower values are better.
 Best values are in \textbf{bold}, and second-best values are \underline{underlined}. 
\par
		\par
		\end{footnotesize}
	\end{tablenotes}
\end{scriptsize}
\end{threeparttable}
  \vspace{-10pt}    
\end{table*}

%% file: tables/validity_novelty.tex
\begin{table*}
	\centering
		\caption{Comparison on Validity and Novelty between \method and \squid}
	\label{tbl:validity_novelty}
	\begin{scriptsize}
\begin{threeparttable}
	\begin{tabular}{
		@{\hspace{3pt}}l@{\hspace{10pt}}
		@{\hspace{10pt}}r@{\hspace{10pt}}
		@{\hspace{10pt}}r@{\hspace{10pt}}
		@{\hspace{10pt}}r@{\hspace{3pt}}
		}
		\toprule
		method & \#v\% & \#n\% & \#v\&n\% \\
		\midrule
		\squid ($\lambda$=0.3) & \textbf{100.0} & 96.7 & 96.7 \\
		\squid ($\lambda$=1.0) & \textbf{100.0} & 98.4 & 98.4 \\
		\method & 99.1 & 99.8 & 98.9 \\
		\methodwithsguide & 99.2 & \textbf{99.9} & \textbf{99.1} \\
		\bottomrule
	\end{tabular}%
	\begin{tablenotes}[normal,flushleft]
		\begin{footnotesize}
	\item 
\!\!Columns represent: ``\#v\%'': the percentage of generated molecules that are valid;
		``\#n\%'': the percentage of valid molecules that are novel;
		``\#v\&n\%'': the percentage of generated molecules that are valid and novel.
		Best values are in \textbf{bold}. 
		\par
		\end{footnotesize}
	\end{tablenotes}
\end{threeparttable}
\end{scriptsize}
\end{table*}

%% file: tables/overall_results_quality0.5.tex
\begin{table*}[!h]
	\centering
		\caption{Comparison on Quality of Generated Desirable Molecules between \method and \squid ($\delta_g$=0.5)}
	\label{tbl:overall_results_quality_05}
	\begin{scriptsize}
\begin{threeparttable}
	\begin{tabular}{
		@{\hspace{0pt}}l@{\hspace{16pt}}
		@{\hspace{0pt}}l@{\hspace{2pt}}
		@{\hspace{6pt}}c@{\hspace{6pt}}
		%
		@{\hspace{3pt}}c@{\hspace{3pt}}
		@{\hspace{3pt}}c@{\hspace{3pt}}
		@{\hspace{3pt}}c@{\hspace{3pt}}
		@{\hspace{3pt}}c@{\hspace{3pt}}
		}
		\toprule
		group & metric & 
        & \squid ($\lambda$=0.3) & \squid ($\lambda$=1.0)  &  \method & \methodwithsguide  \\
		\midrule
		\multirow{2}{*}{stability}
		& atom stability ($\uparrow$) & 
        & \textbf{0.996} & 0.995 & 0.992 & 0.989     \\
		& mol stability ($\uparrow$) & 
        & \textbf{0.948} & 0.947 & 0.886 & 0.839    \\
		\midrule
		\multirow{4}{*}{3D structures} 
		& RMSD ($\downarrow$) & 
        & 0.907 & 0.906 & 0.897 & \textbf{0.881}    \\
		& JS. bond lengths ($\downarrow$) & 
        & 0.457 & 0.477 & 0.436 & \textbf{0.428}    \\
		& JS. bond angles ($\downarrow$) & 
        & 0.269 & 0.289 & \textbf{0.186} & 0.200    \\
		& JS. dihedral angles ($\downarrow$) & 
        & 0.199 & 0.209 & \textbf{0.168} & 0.170    \\
		\midrule
		\multirow{5}{*}{2D structures} 
		& JS. \#bonds per atoms ($\downarrow$) & 
        & 0.291 & 0.331 & \textbf{0.176} & 0.181    \\
		& JS. basic bond types ($\downarrow$) & 
        & \textbf{0.071} & 0.083 & 0.181 & 0.191    \\
		& JS. \#rings ($\downarrow$) & 
        & 0.280 & 0.330 & \textbf{0.043} & 0.049    \\
		& JS. \#n-sized rings ($\downarrow$) & 
        & \textbf{0.077} & 0.091 & 0.099 & 0.112    \\
		& \#Intersecting rings ($\uparrow$) & 
        & \textbf{6} & 5 & 4 & 5    \\
		%
		\bottomrule
	\end{tabular}%
	\begin{tablenotes}[normal,flushleft]
		\begin{footnotesize}
	\item 
\!\!Rows represent:  {``atom stability'': the proportion of stable atoms that have the correct valency; 
		``molecule stability'': the proportion of generated molecules with all atoms stable;
		``RMSD'': the root mean square deviation (RMSD) between the generated 3D structures of molecules and their optimal conformations; 
		``JS. bond lengths/bond angles/dihedral angles'': the Jensen-Shannon (JS) divergences of bond lengths, bond angles and dihedral angles;
		``JS. \#bonds per atom/basic bond types/\#rings/\#n-sized rings'': the JS divergences of bond counts per atom, basic bond types, counts of all rings, and counts of n-sized rings;
		``\#Intersecting rings'': the number of rings observed in the top-10 frequent rings of both generated and real molecules. } \par
		\par
		\end{footnotesize}
	\end{tablenotes}
\end{threeparttable}
\end{scriptsize}
\end{table*}

%% file: tables/overall_results_quality0.7.tex
\begin{table*}[!h]
	\centering
		\caption{Comparison on Quality of Generated Desirable Molecules between \method and \squid ($\delta_g$=0.7)}
	\label{tbl:overall_results_quality_07}
	\begin{scriptsize}
\begin{threeparttable}
	\begin{tabular}{
		@{\hspace{0pt}}l@{\hspace{14pt}}
		@{\hspace{0pt}}l@{\hspace{2pt}}
		@{\hspace{4pt}}c@{\hspace{4pt}}
		%
		@{\hspace{3pt}}c@{\hspace{3pt}}
		@{\hspace{3pt}}c@{\hspace{3pt}}
		@{\hspace{3pt}}c@{\hspace{3pt}}
		@{\hspace{3pt}}c@{\hspace{3pt}}
		}
		\toprule
		group & metric & 
        & \squid ($\lambda$=0.3) & \squid ($\lambda$=1.0)  &  \method & \methodwithsguide  \\
		\midrule
		\multirow{2}{*}{stability} 
		& atom stability ($\uparrow$) & 
        & \textbf{0.995} & 0.995 & 0.992 & 0.988 \\
		& molecule stability ($\uparrow$) & 
        & 0.944 & \textbf{0.947} & 0.885 & 0.839 \\
		\midrule
		\multirow{4}{*}{3D structures} 
		& RMSD ($\downarrow$) & 
        & 0.897 & 0.906 & 0.897 & \textbf{0.881}    \\
		& JS. bond lengths ($\downarrow$) & 
        & 0.457 & 0.477 & 0.436 & \textbf{0.428}    \\
		& JS. bond angles ($\downarrow$) & 
        & 0.269 & 0.289 & \textbf{0.186} & 0.200    \\
		& JS. dihedral angles ($\downarrow$) & 
        & 0.199 & 0.209 & \textbf{0.168} & 0.170    \\
		\midrule
		\multirow{5}{*}{2D structures} 
		& JS. \#bonds per atoms ($\downarrow$) & 
        & 0.285 & 0.329 & \textbf{0.176} & 0.181    \\
		& JS. basic bond types ($\downarrow$) & 
        & \textbf{0.067} & 0.083 & 0.181 & 0.191    \\
		& JS. \#rings ($\downarrow$) & 
        & 0.273 & 0.328 & \textbf{0.043} & 0.049    \\
		& JS. \#n-sized rings ($\downarrow$) & 
        & \textbf{0.076} & 0.091 & 0.099 & 0.112    \\
		& \#Intersecting rings ($\uparrow$) & 
        & \textbf{6} & 5 & 4 & 5    \\
		%
		\bottomrule
	\end{tabular}%
	\begin{tablenotes}[normal,flushleft]
		\begin{footnotesize}
	\item 
\!\!Rows represent:  {``atom stability'': the proportion of stable atoms that have the correct valency; 
		``molecule stability'': the proportion of generated molecules with all atoms stable;
		``RMSD'': the root mean square deviation (RMSD) between the generated 3D structures of molecules and their optimal conformations; 
		``JS. bond lengths/bond angles/dihedral angles'': the Jensen-Shannon (JS) divergences of bond lengths, bond angles and dihedral angles;
		``JS. \#bonds per atom/basic bond types/\#rings/\#n-sized rings'': the JS divergences of bond counts per atom, basic bond types, counts of all rings, and counts of n-sized rings;
		``\#Intersecting rings'': the number of rings observed in the top-10 frequent rings of both generated and real molecules. } \par
		\par
		\end{footnotesize}
	\end{tablenotes}
\end{threeparttable}
\end{scriptsize}
\end{table*}

%% file: tables/overall_results_quality1.0.tex
\begin{table*}[!h]
	\centering
		\caption{Comparison on Quality of Generated Desirable Molecules between \method and \squid ($\delta_g$=1.0)}
	\label{tbl:overall_results_quality_10}
	\begin{scriptsize}
\begin{threeparttable}
	\begin{tabular}{
		@{\hspace{0pt}}l@{\hspace{14pt}}
		@{\hspace{0pt}}l@{\hspace{2pt}}
		@{\hspace{4pt}}c@{\hspace{4pt}}
		%
		@{\hspace{3pt}}c@{\hspace{3pt}}
		@{\hspace{3pt}}c@{\hspace{3pt}}
		@{\hspace{3pt}}c@{\hspace{3pt}}
		@{\hspace{3pt}}c@{\hspace{3pt}}
		}
		\toprule
		group & metric & 
        & \squid ($\lambda$=0.3) & \squid ($\lambda$=1.0)  &  \method & \methodwithsguide \\
		\midrule
		\multirow{2}{*}{stability}
		& atom stability ($\uparrow$) & 
        & \textbf{0.995} & \textbf{0.995} & 0.992 & 0.988     \\
		& mol stability ($\uparrow$) & 
        & 0.942 & \textbf{0.947} & 0.885 & 0.839    \\
		\midrule
		\multirow{4}{*}{3D structures} 
		& RMSD ($\downarrow$) & 
        & 0.898 & 0.906 & 0.897 & \textbf{0.881}    \\
		& JS. bond lengths ($\downarrow$) & 
        & 0.457 & 0.477 & 0.436 & \textbf{0.428}    \\
		& JS. bond angles ($\downarrow$) & 
        & 0.269 & 0.289 & \textbf{0.186} & 0.200   \\
		& JS. dihedral angles ($\downarrow$) & 
        & 0.199 & 0.209 & \textbf{0.168} & 0.170    \\
		\midrule
		\multirow{5}{*}{2D structures} 
		& JS. \#bonds per atoms ($\downarrow$) & 
        & 0.280 & 0.330 & \textbf{0.176} & 0.181    \\
		& JS. basic bond types ($\downarrow$) & 
        & \textbf{0.066} & 0.083 & 0.181 & 0.191   \\
		& JS. \#rings ($\downarrow$) & 
        & 0.269 & 0.328 & \textbf{0.043} & 0.049    \\
		& JS. \#n-sized rings ($\downarrow$) & 
        & \textbf{0.075} & 0.091 & 0.099 & 0.112    \\
		& \#Intersecting rings ($\uparrow$) & 
        & \textbf{6} & 5 & 4 & 5    \\
		%
		\bottomrule
	\end{tabular}%
	\begin{tablenotes}[normal,flushleft]
		\begin{footnotesize}
	\item 
\!\!Rows represent:  {``atom stability'': the proportion of stable atoms that have the correct valency; 
		``molecule stability'': the proportion of generated molecules with all atoms stable;
		``RMSD'': the root mean square deviation (RMSD) between the generated 3D structures of molecules and their optimal conformations; 
		``JS. bond lengths/bond angles/dihedral angles'': the Jensen-Shannon (JS) divergences of bond lengths, bond angles and dihedral angles;
		``JS. \#bonds per atom/basic bond types/\#rings/\#n-sized rings'': the JS divergences of bond counts per atom, basic bond types, counts of all rings, and counts of n-sized rings;
		``\#Intersecting rings'': the number of rings observed in the top-10 frequent rings of both generated and real molecules. } \par
		\par
		\end{footnotesize}
	\end{tablenotes}
\end{threeparttable}
\end{scriptsize}
\end{table*}

%% file: tables/overall_results_docking4.tex
\begin{table*}[!h]
	\centering
		\caption{Additional Comparison on PMG When All Methods Generate 100 Molecules}
	\label{tbl:overall_results_docking_100}
\begin{threeparttable}
	\begin{scriptsize}
	\begin{tabular}{
		@{\hspace{2pt}}l@{\hspace{2pt}}
		@{\hspace{2pt}}r@{\hspace{2pt}}
		@{\hspace{2pt}}r@{\hspace{2pt}}
		@{\hspace{2pt}}r@{\hspace{2pt}}
		@{\hspace{6pt}}r@{\hspace{6pt}}
		@{\hspace{2pt}}r@{\hspace{2pt}}
		@{\hspace{2pt}}r@{\hspace{2pt}}
		@{\hspace{5pt}}r@{\hspace{5pt}}
		@{\hspace{2pt}}r@{\hspace{2pt}}
		@{\hspace{2pt}}r@{\hspace{2pt}}
		@{\hspace{5pt}}r@{\hspace{5pt}}
		@{\hspace{2pt}}r@{\hspace{2pt}}
	         @{\hspace{2pt}}r@{\hspace{2pt}}
		@{\hspace{5pt}}r@{\hspace{5pt}}
		@{\hspace{2pt}}r@{\hspace{2pt}}
		@{\hspace{2pt}}r@{\hspace{2pt}}
		@{\hspace{5pt}}r@{\hspace{5pt}}
		@{\hspace{2pt}}r@{\hspace{2pt}}
		@{\hspace{2pt}}r@{\hspace{2pt}}
		@{\hspace{5pt}}r@{\hspace{5pt}}
		@{\hspace{2pt}}r@{\hspace{2pt}}
		@{\hspace{2pt}}r@{\hspace{2pt}}
		@{\hspace{5pt}}r@{\hspace{5pt}}
		@{\hspace{2pt}}r@{\hspace{2pt}}
		}
		\toprule
		\multirow{2}{*}{method} & \multicolumn{2}{c}{Vina S$\downarrow$} & & \multicolumn{2}{c}{Vina M$\downarrow$} & & \multicolumn{2}{c}{Vina D$\downarrow$} & & \multicolumn{2}{c}{{HA\%$\uparrow$}}  & & \multicolumn{2}{c}{QED$\uparrow$} & & \multicolumn{2}{c}{SA$\uparrow$} & & \multicolumn{2}{c}{Div$\uparrow$} & 
		& \multirow{2}{*}{time$\downarrow$} \\
	    \cmidrule{2-3}\cmidrule{5-6} \cmidrule{8-9} \cmidrule{11-12} \cmidrule{14-15} \cmidrule{17-18} \cmidrule{20-21}
		 & Avg. & Med. &  & Avg. & Med. &  & Avg. & Med. & & Avg. & Med.  & & Avg. & Med.  & & Avg. & Med.  & & Avg. & Med.  & & \\ 
		\midrule
		Reference                          & -5.32 & -5.66 & & -5.78 & -5.76 & & -6.63 & -6.67 & & - & - & & 0.53 & 0.49 & & 0.77 & 0.77 & & - & - & 
		& - \\
		\midrule
		\AR & \textbf{-5.06} & -4.99 & &  -5.59 & -5.29 & &  -6.16 & -6.05 & &  37.69 & 31.00 & &  0.50 & 0.49 & &  0.66 & 0.65 & & 0.70 & 0.70 & 
		& 7,789 \\
		\pockettwomol   & -4.50 & -4.21 & &  -5.70 & -5.27 & &  -6.43 & -6.25 & &  48.00 & 51.00 & &  0.58 & 0.58 & &  \textbf{0.77} & \textbf{0.78} & &  0.69 & 0.71 &  
		& 2,150 \\
		\targetdiff     & -4.88 & \textbf{-5.82} & &  \textbf{-6.20} & \textbf{-6.36} & &  \textbf{-7.37} & \textbf{-7.51} & &  57.57 & 58.27 & &  0.50 & 0.51 & &  0.60 & 0.59 & &  \textbf{0.72} & 0.71 & 
		& 1,252 \\
		\decompdiffref  & -4.58 & -4.77 & &  -5.47 & -5.51 & &  -6.43 & -6.56 & &  47.76 & 48.66 & &  0.56 & 0.56 & &  0.70 & 0.69  & &  \textbf{0.72} & \textbf{0.72} &  
		& 1,859 \\
		\midrule
		\methodwithpguide      & -4.15 & -4.59 & &  -5.41 & -5.34 & &  -6.49 & -6.74 & &  \textbf{58.52} & 59.00 & &  \textbf{0.67} & \textbf{0.69} & &  0.68 & 0.68 & & 0.67 & 0.70 & 
		& 48 \\
		\methodwithsandpguide  & -4.56 & -4.82 & &  -5.53 & -5.47 & &  -6.60 & -6.78 & &  58.28 & \textbf{60.00} & &  0.66 & 0.68 & &  0.67 & 0.66 & & 0.68 & 0.71 &
		& 58 \\
		\bottomrule
	\end{tabular}%
	\begin{tablenotes}[normal,flushleft]
		\begin{footnotesize}
	\item 
\!\!Columns represent: {``Vina S'': the binding affinities between the initially generated poses of molecules and the protein pockets; 
		``Vina M'': the binding affinities between the poses after local structure minimization and the protein pockets;
		``Vina D'': the binding affinities between the poses determined by AutoDock Vina~\cite{Eberhardt2021} and the protein pockets;
		``HA'': the percentage of generated molecules with Vina D higher than those of condition molecules;
		``QED'': the drug-likeness score;
		``SA'': the synthesizability score;
		``Div'': the diversity among generated molecules;
		``time'': the time cost to generate molecules.}
		\par
		\par
		\end{footnotesize}
	\end{tablenotes}
	\end{scriptsize}
\end{threeparttable}
\end{table*}

%% file: tables/drug_property_generated_mols.tex
\begin{table*}
	\centering
		\caption{Drug Properties of Generated Molecules}
	\label{tbl:drug_property}
	\begin{scriptsize}
\begin{threeparttable}
	\begin{tabular}{
		@{\hspace{0pt}}p{0.23\linewidth}@{\hspace{5pt}}
		@{\hspace{1pt}}r@{\hspace{2pt}}
		@{\hspace{2pt}}r@{\hspace{6pt}}
		@{\hspace{6pt}}r@{\hspace{6pt}}
		}
		\toprule
		Property Name & NL-001 & NL-002 & NL-003 \\
		\midrule
Vina S & -6.817 &  -6.970 & -11.953 \\
Vina M & -7.251 & -7.605 & -12.165 \\
Vina D & -8.319 & -8.986 & -12.308 \\
QED    & 0.834  & 0.851  & 0.772 \\
SA       & 0.72    & 0.74    & 0.57    \\
Lipinski & 5 & 5 & 5 \\
\#rotatable bonds          & 3                                                                                        & 2                                                                                        & 2      \\
molecule weight         & 267.112                                                                                      & 270.117                                                                                      & 390.206    \\
molecule LogP           & 1.698                                                                                        & 2.685                                                                                        & 2.382     \\
\#hydrogen bond doners           & 1                                                                                        & 1                                                                                        & 2      \\
\#hydrogen bond acceptors           & 5                                                                                       & 3                                                                                        & 5      \\
\#molecule charges   & 1                                                                                        & 0                                                                                        & 0      \\
drug-induced liver injury predScore    & 0.227                                                                                        & 0.376                                                                                        & 0.188      \\
drug-induced liver injury predConcern  & Ambiguous/Less concern                                                                       & Ambiguous/Less concern                                                                       & No concern \\
drug-induced liver injury predLabel    & Warnings/Precautions/Adverse reactions & Warnings/Precautions/Adverse reactions & No match   \\
drug-induced liver injury predSeverity & 2                                                                                        & 3                                                                                        & 2      \\
toxicity names         & hydrazone                                                                                    &   -                                                                                           &   -         \\
toxicity score         & 0.236                                                                                        & 0.000                                                                                        & 0.000      \\
bad groups         & -                                                                                             & Tetrahydroisoquinoline:   allergies                                                          &   -         \\
		\bottomrule
	\end{tabular}%
	\begin{tablenotes}[normal,flushleft]
		\begin{footnotesize}
	\item ``-'': no results found by algorithms
\!\! \par
		\par
		\end{footnotesize}
	\end{tablenotes}
\end{threeparttable}
\end{scriptsize}
  \vspace{-10pt}    
\end{table*}

%% file: tables/admet_property_cdk6.tex
\begin{table*}
	\centering
		\caption{Comparison on ADMET Profiles among Generated Molecules and Approved Drugs Targeting CDK6}
	\label{tbl:admet_cdk6}
	\begin{scriptsize}
\begin{threeparttable}
	\begin{tabular}{
		%
		@{\hspace{6pt}}l@{\hspace{5pt}}
		@{\hspace{6pt}}r@{\hspace{6pt}}
		@{\hspace{6pt}}r@{\hspace{6pt}}
		@{\hspace{6pt}}r@{\hspace{6pt}}
		@{\hspace{6pt}}r@{\hspace{6pt}}
		@{\hspace{6pt}}r@{\hspace{6pt}}
		@{\hspace{6pt}}r@{\hspace{6pt}}
		%
		}
		\toprule
		\multirow{2}{*}{Property name} & \multicolumn{2}{c}{Generated molecules} & & \multicolumn{3}{c}{FDA-approved drugs} \\
		\cmidrule{2-3}\cmidrule{5-7}
		 & NL--001 & NL--002 & & Abemaciclib & Palbociclib & Ribociclib \\
		\midrule
\rowcolor[HTML]{D2EAD9}Ames   mutagenesis                             & --   &  --  & & + &  --  & --  \\
\rowcolor[HTML]{D2EAD9}Acute oral toxicity (c)                           & III & III & &  III          & III          & III         \\
Androgen receptor binding                         & +                          & +            &              & +            & +            & +             \\
Aromatase binding                                 & +                          & +            &              & +            & +            & +            \\
Avian toxicity                                    & --                          & --          &                & --            & --            & --            \\
Blood brain barrier                               & +                          & +            &              & +            & +            & +            \\
BRCP inhibitior                                   & --                          & --          &                & --            & --            & --            \\
Biodegradation                                    & --                          & --          &                & --            & --            & --           \\
BSEP inhibitior            & +                          & +            &              & +            & +            & +        \\
Caco-2                                            & +                          & +            &              & --            & --            & --            \\
\rowcolor[HTML]{D2EAD9}Carcinogenicity (binary)                          & --                          & --             &             & --            & --            & --          \\
\rowcolor[HTML]{D2EAD9}Carcinogenicity (trinary)                         & Non-required               & Non-required   &            & Non-required & Non-required & Non-required  \\
Crustacea aquatic toxicity & --                          & --            &              & --            & --            & --            \\
 CYP1A2 inhibition                                 & +                          & +            &              & --            & --            & +             \\
CYP2C19 inhibition                                & --                          & +            &              & +            & --            & +            \\
CYP2C8 inhibition                                 & --                          & --           &               & +            & +            & +            \\
CYP2C9 inhibition                                 & --                          & --           &               & --            & --            & +             \\
CYP2C9 substrate                                  & --                          & --           &               & --            & --            & --            \\
CYP2D6 inhibition                                 & --                          & --           &               & --            & --            & --            \\
CYP2D6 substrate                                  & --                          & --           &               & --            & --            & --            \\
CYP3A4 inhibition                                 & --                          & +            &              & --            & --            & --            \\
CYP3A4 substrate                                  & +                          & --            &              & +            & +            & +            \\
\rowcolor[HTML]{D2EAD9}CYP inhibitory promiscuity                        & +                          & +                    &      & +            & --            & +            \\
Eye corrosion                                     & --                          & --           &               & --            & --            & --            \\
Eye irritation                                    & --                          & --           &               & --            & --            & --             \\
\rowcolor[HTML]{D8E7FF}Estrogen receptor binding                         & +                          & +                    &      & +            & +            & +            \\
Fish aquatic toxicity                             & --                          & +            &              & +            & --            & --            \\
Glucocorticoid receptor   binding                 & +                          & +             &             & +            & +            & +            \\
Honey bee toxicity                                & --                          & --           &               & --            & --            & --            \\
\rowcolor[HTML]{D2EAD9}Hepatotoxicity                                    & +                          & +            &              & +            & +            & +             \\
Human ether-a-go-go-related gene inhibition     & +                          & +               &           & +            & --            & --           \\
\rowcolor[HTML]{D8E7FF}Human intestinal absorption                       & +                          & +             &             & +            & +            & +    \\
\rowcolor[HTML]{D8E7FF}Human oral bioavailability                        & +                          & +              &            & +            & +            & +     \\
\rowcolor[HTML]{D2EAD9}MATE1 inhibitior                                  & --                          & --              &            & --            & --            & --    \\
\rowcolor[HTML]{D2EAD9}Mitochondrial toxicity                            & +                          & +                &          & +            & +            & +    \\
Micronuclear                                      & +                          & +                          & +            & +            & +           \\
\rowcolor[HTML]{D2EAD9}Nephrotoxicity                                    & --                          & --             &             & --            & --            & --             \\
Acute oral toxicity                               & 2.325                      & 1.874    &     & 1.870        & 3.072        & 3.138        \\
\rowcolor[HTML]{D8E7FF}OATP1B1 inhibitior                                & +                          & +              &            & +            & +            & +             \\
\rowcolor[HTML]{D8E7FF}OATP1B3 inhibitior                                & +                          & +              &            & +            & +            & +             \\
\rowcolor[HTML]{D2EAD9}OATP2B1 inhibitior                                & --                          & --             &             & --            & --            & --             \\
OCT1 inhibitior                                   & --                          & --        &                  & +            & --            & +             \\
OCT2 inhibitior                                   & --                          & --        &                  & --            & --            & +             \\
P-glycoprotein inhibitior                         & --                          & --        &                  & +            & +            & +     \\
P-glycoprotein substrate                          & --                          & --        &                  & +            & +            & +     \\
PPAR gamma                                        & +                          & +          &                & +            & +            & +      \\
\rowcolor[HTML]{D8E7FF}Plasma protein binding                            & 0.359        & 0.745     &    & 0.865        & 0.872        & 0.636       \\
Reproductive toxicity                             & +                          & +          &                & +            & +            & +           \\
Respiratory toxicity                              & +                          & +          &                & +            & +            & +         \\
Skin corrosion                                    & --                          & --        &                  & --            & --            & --           \\
Skin irritation                                   & --                          & --        &                  & --            & --            & --         \\
Skin sensitisation                                & --                          & --        &                  & --            & --            & --          \\
Subcellular localzation                           & Mitochondria               & Mitochondria  &             & Lysosomes    & Mitochondria & Mitochondria \\
Tetrahymena pyriformis                            & 0.398                      & 0.903         &             & 1.033        & 1.958        & 1.606         \\
Thyroid receptor binding                          & +                          & +             &             & +            & +            & +           \\
UGT catelyzed                                     & --                          & --           &               & --            & --            & --           \\
\rowcolor[HTML]{D8E7FF}Water solubility                                  & -3.050                     & -3.078              &       & -3.942       & -3.288       & -2.673     \\
		\bottomrule
	\end{tabular}%
	\begin{tablenotes}[normal,flushleft]
		\begin{footnotesize}
	\item Blue cells highlight crucial properties where a negative outcome (``--'') is desired; for acute oral toxicity (c), a higher category (e.g., ``III'') is desired; and for carcinogenicity (trinary), ``Non-required'' is desired.
	Green cells highlight crucial properties where a positive result (``+'') is desired; for plasma protein binding, a lower value is desired; and for water solubility, values higher than -4 are desired~\cite{logs}.
\!\! \par
		\par
		\end{footnotesize}
	\end{tablenotes}
\end{threeparttable}
\end{scriptsize}
  \vspace{--10pt}    
\end{table*}

%% file: tables/admet_property_nep.tex
\begin{table*}
	\centering
		\caption{Comparison on ADMET Profiles among Generated Molecule Targeting NEP and Approved Drugs for Alzhimer's Disease}
	\label{tbl:admet_nep}
	\begin{scriptsize}
\begin{threeparttable}
	\begin{tabular}{
		@{\hspace{6pt}}l@{\hspace{5pt}}
		@{\hspace{6pt}}r@{\hspace{6pt}}
		@{\hspace{6pt}}r@{\hspace{6pt}}
		@{\hspace{6pt}}r@{\hspace{6pt}}
		@{\hspace{6pt}}r@{\hspace{6pt}}
		@{\hspace{6pt}}r@{\hspace{6pt}}
		%
		%
		%
		}
		\toprule
		\multirow{2}{*}{Property name} & Generated molecule & & \multicolumn{3}{c}{FDA-approved drugs} \\
\cmidrule{2-2}\cmidrule{4-6}
			& NL--003 & & Donepezil	& Galantamine & Rivastigmine \\
		\midrule
\rowcolor[HTML]{D2EAD9} 
Ames   mutagenesis                            & --                      &              & --                                    & --                                 & --                     \\
\rowcolor[HTML]{D2EAD9}Acute oral toxicity (c)                       & III           &                       & III                                  & III                               & II                      \\
Androgen receptor binding                     & +      &      & +            & --         & --         \\
Aromatase binding                             & --     &       & +            & --         & --        \\
Avian toxicity                                & --     &                               & --                                    & --                                 & --                        \\
\rowcolor[HTML]{D8E7FF} 
Blood brain barrier                           & +      &                              & +                                    & +                                 & +                        \\
BRCP inhibitior                               & --     &       & --            & --         & --         \\
Biodegradation                                & --     &                               & --                                    & --                                 & --                        \\
BSEP inhibitior                               & +      &      & +            & --         & --         \\
Caco-2                                        & +      &      & +            & +         & +         \\
\rowcolor[HTML]{D2EAD9} 
Carcinogenicity (binary)                      & --     &                               & --                                    & --                                 & --                        \\
\rowcolor[HTML]{D2EAD9} 
Carcinogenicity (trinary)                     & Non-required    &                     & Non-required                         & Non-required                      & Non-required             \\
Crustacea aquatic toxicity                    & +               &                     & +                                    & +                                 & --                        \\
CYP1A2 inhibition                             & +               &                     & +                                    & --                                 & --                        \\
CYP2C19 inhibition                            & +               &                     & --                                    & --                                 & --                        \\
CYP2C8 inhibition                             & +               &                     & --                                    & --                                 & --                        \\
CYP2C9 inhibition                             & --              &                      & --                                    & --                                 & --                        \\
CYP2C9 substrate                              & --              &                      & --                                    & --                                 & --                        \\
CYP2D6 inhibition                             & --              &                      & +                                    & --                                 & --                        \\
CYP2D6 substrate                              & --              &                      & +                                    & +                                 & +                        \\
CYP3A4 inhibition                             & --              &                      & --                                    & --                                 & --                        \\
CYP3A4 substrate                              & +               &                     & +                                    & +                                 & --                        \\
\rowcolor[HTML]{D2EAD9} 
CYP inhibitory promiscuity                    & +               &                     & +                                    & --                                 & --                        \\
Eye corrosion                                 & --     &       & --            & --         & --         \\
Eye irritation                                & --     &       & --            & --         & --         \\
Estrogen receptor binding                     & +      &      & +            & --         & --         \\
Fish aquatic toxicity                         & --     &                               & +                                    & +                                 & +                        \\
Glucocorticoid receptor binding             & --      &      & +            & --         & --         \\
Honey bee toxicity                            & --    &                                & --                                    & --                                 & --                        \\
\rowcolor[HTML]{D2EAD9} 
Hepatotoxicity                                & +     &                               & +                                    & --                                 & --                        \\
Human ether-a-go-go-related gene inhibition & +       &     & +            & --         & --         \\
\rowcolor[HTML]{D8E7FF} 
Human intestinal absorption                   & +     &                               & +                                    & +                                 & +                        \\
\rowcolor[HTML]{D8E7FF} 
Human oral bioavailability                    & --    &                                & +                                    & +                                 & +                        \\
\rowcolor[HTML]{D2EAD9} 
MATE1 inhibitior                              & --    &                                & --                                    & --                                 & --                        \\
\rowcolor[HTML]{D2EAD9} 
Mitochondrial toxicity                        & +     &                               & +                                    & +                                 & +                        \\
Micronuclear                                  & +     &       & --            & --         & +         \\
\rowcolor[HTML]{D2EAD9} 
Nephrotoxicity                                & --    &                                & --                                    & --                                 & --                        \\
Acute oral toxicity                           & 2.704  &      & 2.098        & 2.767     & 2.726     \\
\rowcolor[HTML]{D8E7FF} 
OATP1B1 inhibitior                            & +      &                              & +                                    & +                                 & +                        \\
\rowcolor[HTML]{D8E7FF} 
OATP1B3 inhibitior                            & +      &                              & +                                    & +                                 & +                        \\
\rowcolor[HTML]{D2EAD9} 
OATP2B1 inhibitior                            & --     &                               & --                                    & --                                 & --                        \\
OCT1 inhibitior                               & +      &      & +            & --         & --         \\
OCT2 inhibitior                               & --     &       & +            & --         & --         \\
P-glycoprotein inhibitior                     & +      &      & +            & --         & --         \\
\rowcolor[HTML]{D8E7FF} 
P-glycoprotein substrate                      & +      &                              & +                                    & +                                 & --                        \\
PPAR gamma                                    & +      &      & --            & --         & --         \\
\rowcolor[HTML]{D8E7FF} 
Plasma protein binding                        & 0.227   &                             & 0.883                                & 0.230                             & 0.606                    \\
Reproductive toxicity                         & +       &     & +            & +         & +         \\
Respiratory toxicity                          & +       &     & +            & +         & +         \\
Skin corrosion                                & --      &      & --            & --         & --         \\
Skin irritation                               & --      &      & --            & --         & --         \\
Skin sensitisation                            & --      &      & --            & --         & --         \\
Subcellular localzation                       & Mitochondria & &Mitochondria & Lysosomes & Mitochondria  \\
Tetrahymena pyriformis                        & 0.053           &                     & 0.979                                & 0.563                             & 0.702                        \\
Thyroid receptor binding                      & +       &     & +            & +         & --             \\
UGT catelyzed                                 & --      &      & --            & +         & --             \\
\rowcolor[HTML]{D8E7FF} 
Water solubility                              & -3.586   &                            & -2.425                               & -2.530                            & -3.062                       \\
		\bottomrule
	\end{tabular}%
	\begin{tablenotes}[normal,flushleft]
		\begin{footnotesize}
	\item Blue cells highlight crucial properties where a negative outcome (``--'') is desired; for acute oral toxicity (c), a higher category (e.g., ``III'') is desired; and for carcinogenicity (trinary), ``Non-required'' is desired.
	Green cells highlight crucial properties where a positive result (``+'') is desired; for plasma protein binding, a lower value is desired; and for water solubility, values higher than -4 are desired~\cite{logs}.
\!\! \par
		\par
		\end{footnotesize}
	\end{tablenotes}
\end{threeparttable}
\end{scriptsize}
  \vspace{--10pt}    
\end{table*}

%% file: algorithms/shapemol.tex
\begin{algorithm}[!h]
    \caption{\method}
    \label{alg:shapemol}
	\textbf{Required Input: $\molx$} \\
	\textbf{Optional Input: $\pocket$} 
    \begin{algorithmic}[1]
        \FullLineComment{calculate a shape embedding with Algorithm~\ref{alg:see_shaperep}}
        \State $\shapehiddenmat$, $\pc$ = $\SEE(\molx)$
        \FullLineComment{generate a molecule conditioned on the shape embedding with Algorithm~\ref{alg:diffgen}}
         \If{\pocket is not available}
        \State $\moly = \diffgenerative(\shapehiddenmat, \molx)$
        \Else
        \State $\moly = \diffgenerative(\shapehiddenmat, \molx, \pocket)$
        \EndIf
        \State \Return \moly
    \end{algorithmic}
\end{algorithm}

%% file: algorithms/shaperep.tex
\begin{algorithm}[!h]
    \caption{\SEE for shape embedding calculation}
    \label{alg:see_shaperep}
    \textbf{Required Input: $\molx$}
    \begin{algorithmic}[1]
        \FullLineComment{sample a point cloud over the molecule surface shape}
        \State $\pc$ = $\text{samplePointCloud}(\molx)$
        \FullLineComment{encode the point cloud into a latent embedding (Equation~\ref{eqn:shape_embed})}
        \State $\shapehiddenmat = \SEE(\pc)$
        \FullLineComment{move the center of \pc to zero}
        \State $\pc = \pc - \text{center}(\pc)$
        \State \Return \shapehiddenmat, \pc
    \end{algorithmic}
\end{algorithm}

%% file: algorithms/diffgen.tex
\begin{algorithm}[!h]
    \caption{\diffgenerative for molecule generation}
    \label{alg:diffgen}
    	\textbf{Required Input: $\molx$, \shapehiddenmat} \\
	\textbf{Optional Input: $\pocket$} 
    \begin{algorithmic}[1]
        \FullLineComment{sample the number of atoms in the generated molecule}
        \State $n = \text{sampleAtomNum}(\molx)$
        \FullLineComment{sample initial positions and types of $n$ atoms}
        \State $\{\pos_T\}^n = \mathcal{N}(0, I)$
        \State $\{\atomfeat_T\}^n = C(K, \frac{1}{K})$
        \FullLineComment{generate a molecule by denoising $\{(\pos_T, \atomfeat_T)\}^n$ to $\{(\pos_0, \atomfeat_0)\}^n$}
        \For{$t = T$ to $1$}
            \IndentLineComment{predict the molecule without noise using the shape-conditioned molecule prediction module \molpred}{1.5}
            \State $(\tilde{\pos}_{0,t}, \tilde{\atomfeat}_{0,t}) = \molpred(\pos_t, \atomfeat_t, \shapehiddenmat)$
            \If{use shape guidance and $t > s$}
                \State $\tilde{\pos}_{0,t} = \shapeguide(\tilde{\pos}_{0,t}, \molx)$
            \EndIf
            \IndentLineComment{sample $(\pos_{t-1}, \atomfeat_{t-1})$ from $(\pos_t, \atomfeat_t)$ and $(\tilde{\pos}_{0,t}, \tilde{\atomfeat}_{0,t})$}{1.5}
            \State $\pos_{t-1} = P(\pos_{t-1}|\pos_t, \tilde{\pos}_{o,t})$
            \State $\atomfeat_{t-1} = P(\atomfeat_{t-1}|\atomfeat_t, \tilde{\atomfeat}_{o,t})$
            \If{use pocket guidance and $\pocket$ is available}
                \State $\pos_{t-1} = \pocketguide(\pos_{t-1}, \pocket)$
            \EndIf  
        \EndFor
        \State \Return $\moly = (\pos_0, \atomfeat_0)$
    \end{algorithmic}
\end{algorithm}

%% file: paper.bbl
\begin{thebibliography}{10}
\expandafter\ifx\csname url\endcsname\relax
  \def\url#1{\texttt{#1}}\fi
\expandafter\ifx\csname urlprefix\endcsname\relax\def\urlprefix{URL }\fi
\providecommand{\bibinfo}[2]{#2}
\providecommand{\eprint}[2][]{\url{#2}}

\bibitem{Sun2022}
\bibinfo{author}{Sun, D.}, \bibinfo{author}{Gao, W.}, \bibinfo{author}{Hu, H.}
  \& \bibinfo{author}{Zhou, S.}
\newblock \bibinfo{title}{Why 90\% of clinical drug development fails and how
  to improve it?}
\newblock \emph{\bibinfo{journal}{Acta Pharm. Sin. B.}}
  \textbf{\bibinfo{volume}{12}}, \bibinfo{pages}{3049--3062}
  (\bibinfo{year}{2022}).

\bibitem{Wouters2020}
\bibinfo{author}{Wouters, O.~J.}, \bibinfo{author}{McKee, M.} \&
  \bibinfo{author}{Luyten, J.}
\newblock \bibinfo{title}{Estimated research and development investment needed
  to bring a new medicine to market, 2009-2018}.
\newblock \emph{\bibinfo{journal}{JAMA}} \textbf{\bibinfo{volume}{323}},
  \bibinfo{pages}{844} (\bibinfo{year}{2020}).

\bibitem{Yu2016}
\bibinfo{author}{Yu, W.} \& \bibinfo{author}{MacKerell, A.~D.}
\newblock \emph{\bibinfo{title}{Computer-Aided Drug Design Methods}},
  \bibinfo{pages}{85--106} (\bibinfo{publisher}{Springer New York},
  \bibinfo{year}{2016}).

\bibitem{Acharya2011}
\bibinfo{author}{Acharya, C.}, \bibinfo{author}{Coop, A.},
  \bibinfo{author}{Polli, J.~E.} \& \bibinfo{author}{MacKerell, A.~D.}
\newblock \bibinfo{title}{Recent advances in ligand-based drug design:
  Relevance and utility of the conformationally sampled pharmacophore
  approach}.
\newblock \emph{\bibinfo{journal}{Curr. Comput. Aided Drug Des.}}
  \textbf{\bibinfo{volume}{7}}, \bibinfo{pages}{10--22} (\bibinfo{year}{2011}).

\bibitem{Anderson2003}
\bibinfo{author}{Anderson, A.~C.}
\newblock \bibinfo{title}{The process of structure-based drug design}.
\newblock \emph{\bibinfo{journal}{Chem. Biol.}} \textbf{\bibinfo{volume}{10}},
  \bibinfo{pages}{787--797} (\bibinfo{year}{2003}).

\bibitem{Gimeno2019}
\bibinfo{author}{Gimeno, A.} \emph{et~al.}
\newblock \bibinfo{title}{The light and dark sides of virtual screening: What
  is there to know?}
\newblock \emph{\bibinfo{journal}{Int. J. Mol. Sci.}}
  \textbf{\bibinfo{volume}{20}}, \bibinfo{pages}{1375} (\bibinfo{year}{2019}).

\bibitem{kingma2013auto}
\bibinfo{author}{Kingma, D.~P.} \& \bibinfo{author}{Welling, M.}
\newblock \bibinfo{title}{Auto-encoding variational bayes}.
\newblock \emph{\bibinfo{journal}{arXiv:1312.6114}}  (\bibinfo{year}{2013}).

\bibitem{song2021denoising}
\bibinfo{author}{Song, J.}, \bibinfo{author}{Meng, C.} \&
  \bibinfo{author}{Ermon, S.}
\newblock \bibinfo{title}{Denoising diffusion implicit models}.
\newblock In \emph{\bibinfo{booktitle}{9th International Conference on Learning
  Representations}} (\bibinfo{year}{2021}).

\bibitem{OpenAI2023}
\bibinfo{author}{OpenAI} \emph{et~al.}
\newblock \bibinfo{title}{{GPT-4} technical report}.
\newblock \emph{\bibinfo{journal}{arXiv:2303.08774}}  (\bibinfo{year}{2023}).

\bibitem{Yu2024}
\bibinfo{author}{Yu, B.}, \bibinfo{author}{Baker, F.~N.},
  \bibinfo{author}{Chen, Z.}, \bibinfo{author}{Ning, X.} \&
  \bibinfo{author}{Sun, H.}
\newblock \bibinfo{title}{Llasmol: Advancing large language models for
  chemistry with a large-scale, comprehensive, high-quality instruction tuning
  dataset}.
\newblock \emph{\bibinfo{journal}{arXiv:2402.09391}}  (\bibinfo{year}{2024}).

\bibitem{jin18jtvae}
\bibinfo{author}{Jin, W.}, \bibinfo{author}{Barzilay, R.} \&
  \bibinfo{author}{Jaakkola, T.}
\newblock \bibinfo{title}{Junction tree variational autoencoder for molecular
  graph generation}.
\newblock In \bibinfo{editor}{Dy, J.} \& \bibinfo{editor}{Krause, A.} (eds.)
  \emph{\bibinfo{booktitle}{Proceedings of the 35th International Conference on
  Machine Learning}}, vol.~\bibinfo{volume}{80} of
  \emph{\bibinfo{series}{Proceedings of Machine Learning Research}},
  \bibinfo{pages}{2323--2332} (\bibinfo{publisher}{PMLR},
  \bibinfo{year}{2018}).

\bibitem{schneuing2022structure}
\bibinfo{author}{Schneuing, A.} \emph{et~al.}
\newblock \bibinfo{title}{Structure-based drug design with equivariant
  diffusion models}.
\newblock \emph{\bibinfo{journal}{arXiv:2210.13695}}  (\bibinfo{year}{2022}).

\bibitem{liu2024conversational}
\bibinfo{author}{Liu, S.} \emph{et~al.}
\newblock \bibinfo{title}{Conversational drug editing using retrieval and
  domain feedback}.
\newblock In \emph{\bibinfo{booktitle}{12th International Conference on
  Learning Representations}} (\bibinfo{year}{2024}).

\bibitem{Bostroem2006}
\bibinfo{author}{Boström, J.}, \bibinfo{author}{Hogner, A.} \&
  \bibinfo{author}{Schmitt, S.}
\newblock \bibinfo{title}{Do structurally similar ligands bind in a similar
  fashion?}
\newblock \emph{\bibinfo{journal}{J. Med. Chem.}}
  \textbf{\bibinfo{volume}{49}}, \bibinfo{pages}{6716--6725}
  (\bibinfo{year}{2006}).

\bibitem{Eberhardt2021}
\bibinfo{author}{Eberhardt, J.}, \bibinfo{author}{Santos-Martins, D.},
  \bibinfo{author}{Tillack, A.~F.} \& \bibinfo{author}{Forli, S.}
\newblock \bibinfo{title}{Autodock vina 1.2.0: New docking methods, expanded
  force field, and python bindings}.
\newblock \emph{\bibinfo{journal}{J. Chem. Inf. Model.}}
  \textbf{\bibinfo{volume}{61}}, \bibinfo{pages}{3891--3898}
  (\bibinfo{year}{2021}).

\bibitem{Bickerton2012}
\bibinfo{author}{Bickerton, G.~R.}, \bibinfo{author}{Paolini, G.~V.},
  \bibinfo{author}{Besnard, J.}, \bibinfo{author}{Muresan, S.} \&
  \bibinfo{author}{Hopkins, A.~L.}
\newblock \bibinfo{title}{Quantifying the chemical beauty of drugs}.
\newblock \emph{\bibinfo{journal}{Nat. Chem.}} \textbf{\bibinfo{volume}{4}},
  \bibinfo{pages}{90--98} (\bibinfo{year}{2012}).

\bibitem{Lipinski1997}
\bibinfo{author}{Lipinski, C.~A.}, \bibinfo{author}{Lombardo, F.},
  \bibinfo{author}{Dominy, B.~W.} \& \bibinfo{author}{Feeney, P.~J.}
\newblock \bibinfo{title}{Experimental and computational approaches to estimate
  solubility and permeability in drug discovery and development settings}.
\newblock \emph{\bibinfo{journal}{Adv. Drug Deliv. Rev.}}
  \textbf{\bibinfo{volume}{23}}, \bibinfo{pages}{3--25} (\bibinfo{year}{1997}).

\bibitem{GmezBombarelli2018}
\bibinfo{author}{G{\'{o}}mez-Bombarelli, R.} \emph{et~al.}
\newblock \bibinfo{title}{Automatic chemical design using a data-driven
  continuous representation of molecules}.
\newblock \emph{\bibinfo{journal}{{ACS} Cent. Sci.}}
  \textbf{\bibinfo{volume}{4}}, \bibinfo{pages}{268--276}
  (\bibinfo{year}{2018}).

\bibitem{Chen2021modof}
\bibinfo{author}{Chen, Z.}, \bibinfo{author}{Min, M.~R.},
  \bibinfo{author}{Parthasarathy, S.} \& \bibinfo{author}{Ning, X.}
\newblock \bibinfo{title}{A deep generative model for molecule optimization via
  one fragment modification}.
\newblock \emph{\bibinfo{journal}{Nat. Mach. Intell.}}
  \textbf{\bibinfo{volume}{3}}, \bibinfo{pages}{1040--1049}
  (\bibinfo{year}{2021}).

\bibitem{hoogeboom22diff}
\bibinfo{author}{Hoogeboom, E.}, \bibinfo{author}{Satorras, V.~G.},
  \bibinfo{author}{Vignac, C.} \& \bibinfo{author}{Welling, M.}
\newblock \bibinfo{title}{Equivariant diffusion for molecule generation in
  3{D}}.
\newblock In \bibinfo{editor}{Chaudhuri, K.} \emph{et~al.} (eds.)
  \emph{\bibinfo{booktitle}{Proceedings of the 39th International Conference on
  Machine Learning}}, vol. \bibinfo{volume}{162} of
  \emph{\bibinfo{series}{Proceedings of Machine Learning Research}},
  \bibinfo{pages}{8867--8887} (\bibinfo{publisher}{PMLR},
  \bibinfo{year}{2022}).

\bibitem{long2022zero}
\bibinfo{author}{Long, S.}, \bibinfo{author}{Zhou, Y.}, \bibinfo{author}{Dai,
  X.} \& \bibinfo{author}{Zhou, H.}
\newblock \bibinfo{title}{Zero-shot 3d drug design by sketching and
  generating}.
\newblock In \bibinfo{editor}{Oh, A.~H.}, \bibinfo{editor}{Agarwal, A.},
  \bibinfo{editor}{Belgrave, D.} \& \bibinfo{editor}{Cho, K.} (eds.)
  \emph{\bibinfo{booktitle}{Advances in Neural Information Processing Systems}}
  (\bibinfo{year}{2022}).

\bibitem{adams2023equivariant}
\bibinfo{author}{Adams, K.} \& \bibinfo{author}{Coley, C.~W.}
\newblock \bibinfo{title}{Equivariant shape-conditioned generation of 3d
  molecules for ligand-based drug design}.
\newblock In \emph{\bibinfo{booktitle}{11th International Conference on
  Learning Representations}} (\bibinfo{year}{2023}).

\bibitem{Chen2023ShapeMol}
\bibinfo{author}{Chen, Z.}, \bibinfo{author}{Peng, B.},
  \bibinfo{author}{Parthasarathy, S.} \& \bibinfo{author}{Ning, X.}
\newblock \bibinfo{title}{Shape-conditioned 3d molecule generation via
  equivariant diffusion models}.
\newblock \emph{\bibinfo{journal}{arXiv:2403.12987}}  (\bibinfo{year}{2023}).

\bibitem{Jonas20a}
\bibinfo{author}{K{\"o}hler, J.}, \bibinfo{author}{Klein, L.} \&
  \bibinfo{author}{Noe, F.}
\newblock \bibinfo{title}{Equivariant flows: Exact likelihood generative
  learning for symmetric densities}.
\newblock In \bibinfo{editor}{III, H.~D.} \& \bibinfo{editor}{Singh, A.} (eds.)
  \emph{\bibinfo{booktitle}{Proceedings of the 37th International Conference on
  Machine Learning}}, vol. \bibinfo{volume}{119} of
  \emph{\bibinfo{series}{Proceedings of Machine Learning Research}},
  \bibinfo{pages}{5361--5370} (\bibinfo{publisher}{PMLR},
  \bibinfo{year}{2020}).

\bibitem{luo2021sbdd}
\bibinfo{author}{Luo, S.}, \bibinfo{author}{Guan, J.}, \bibinfo{author}{Ma, J.}
  \& \bibinfo{author}{Peng, J.}
\newblock \bibinfo{title}{A 3d generative model for structure-based drug
  design}.
\newblock In \bibinfo{editor}{Beygelzimer, A.}, \bibinfo{editor}{Dauphin, Y.},
  \bibinfo{editor}{Liang, P.} \& \bibinfo{editor}{Vaughan, J.~W.} (eds.)
  \emph{\bibinfo{booktitle}{Advances in Neural Information Processing Systems}}
  (\bibinfo{year}{2021}).

\bibitem{peng22pocket2mol}
\bibinfo{author}{Peng, X.} \emph{et~al.}
\newblock \bibinfo{title}{{P}ocket2{M}ol: Efficient molecular sampling based on
  3{D} protein pockets}.
\newblock In \bibinfo{editor}{Chaudhuri, K.} \emph{et~al.} (eds.)
  \emph{\bibinfo{booktitle}{Proceedings of the 39th International Conference on
  Machine Learning}}, vol. \bibinfo{volume}{162} of
  \emph{\bibinfo{series}{Proceedings of Machine Learning Research}},
  \bibinfo{pages}{17644--17655} (\bibinfo{publisher}{PMLR},
  \bibinfo{year}{2022}).

\bibitem{guan2023targetdiff}
\bibinfo{author}{Guan, J.} \emph{et~al.}
\newblock \bibinfo{title}{3d equivariant diffusion for target-aware molecule
  generation and affinity prediction}.
\newblock In \emph{\bibinfo{booktitle}{11th International Conference on
  Learning Representations}} (\bibinfo{year}{2023}).

\bibitem{guan2023decompdiff}
\bibinfo{author}{Guan, J.} \emph{et~al.}
\newblock \bibinfo{title}{{D}ecomp{D}iff: Diffusion models with decomposed
  priors for structure-based drug design}.
\newblock In \bibinfo{editor}{Krause, A.} \emph{et~al.} (eds.)
  \emph{\bibinfo{booktitle}{Proceedings of the 40th International Conference on
  Machine Learning}}, vol. \bibinfo{volume}{202} of
  \emph{\bibinfo{series}{Proceedings of Machine Learning Research}},
  \bibinfo{pages}{11827--11846} (\bibinfo{publisher}{PMLR},
  \bibinfo{year}{2023}).

\bibitem{ragoza2022chemsci}
\bibinfo{author}{Ragoza, M.}, \bibinfo{author}{Masuda, T.} \&
  \bibinfo{author}{Koes, D.~R.}
\newblock \bibinfo{title}{{Generating 3D molecules conditional on receptor
  binding sites with deep generative models}}.
\newblock \emph{\bibinfo{journal}{Chem. Sci.}} \textbf{\bibinfo{volume}{13}},
  \bibinfo{pages}{2701--2713} (\bibinfo{year}{2022}).

\bibitem{liu2022}
\bibinfo{author}{Liu, M.}, \bibinfo{author}{Luo, Y.}, \bibinfo{author}{Uchino,
  K.}, \bibinfo{author}{Maruhashi, K.} \& \bibinfo{author}{Ji, S.}
\newblock \bibinfo{title}{Generating 3{D} molecules for target protein
  binding}.
\newblock In \bibinfo{editor}{Chaudhuri, K.} \emph{et~al.} (eds.)
  \emph{\bibinfo{booktitle}{Proceedings of the 39th International Conference on
  Machine Learning}}, vol. \bibinfo{volume}{162} of
  \emph{\bibinfo{series}{Proceedings of Machine Learning Research}},
  \bibinfo{pages}{13912--13924} (\bibinfo{publisher}{PMLR},
  \bibinfo{year}{2022}).

\bibitem{mose2020}
\bibinfo{author}{Polykovskiy, D.} \emph{et~al.}
\newblock \bibinfo{title}{Molecular sets (moses): A benchmarking platform for
  molecular generation models}.
\newblock \emph{\bibinfo{journal}{Front. Pharmacol.}}
  \textbf{\bibinfo{volume}{11}} (\bibinfo{year}{2020}).

\bibitem{rdkit}
\bibinfo{author}{Landrum, G.} \emph{et~al.}
\newblock \bibinfo{title}{rdkit/rdkit: 2023\_03\_2 (q1 2023) release}
  (\bibinfo{year}{2023}).

\bibitem{Francoeur2020}
\bibinfo{author}{Francoeur, P.~G.} \emph{et~al.}
\newblock \bibinfo{title}{Three-dimensional convolutional neural networks and a
  cross-docked data set for structure-based drug design}.
\newblock \emph{\bibinfo{journal}{J. Chem. Inf. Model.}}
  \textbf{\bibinfo{volume}{60}}, \bibinfo{pages}{4200--4215}
  (\bibinfo{year}{2020}).

\bibitem{Hawkins2006}
\bibinfo{author}{Hawkins, P. C.~D.}, \bibinfo{author}{Skillman, A.~G.} \&
  \bibinfo{author}{Nicholls, A.}
\newblock \bibinfo{title}{Comparison of shape-matching and docking as virtual
  screening tools}.
\newblock \emph{\bibinfo{journal}{J. Med. Chem.}}
  \textbf{\bibinfo{volume}{50}}, \bibinfo{pages}{74--82}
  (\bibinfo{year}{2006}).

\bibitem{Ertl2009}
\bibinfo{author}{Ertl, P.} \& \bibinfo{author}{Schuffenhauer, A.}
\newblock \bibinfo{title}{Estimation of synthetic accessibility score of
  drug-like molecules based on molecular complexity and fragment
  contributions}.
\newblock \emph{\bibinfo{journal}{J. Cheminform.}}
  \textbf{\bibinfo{volume}{1}}, \bibinfo{pages}{8} (\bibinfo{year}{2009}).

\bibitem{peng2023moldiff}
\bibinfo{author}{Peng, X.}, \bibinfo{author}{Guan, J.}, \bibinfo{author}{Liu,
  Q.} \& \bibinfo{author}{Ma, J.}
\newblock \bibinfo{title}{{M}ol{D}iff: Addressing the atom-bond inconsistency
  problem in 3{D} molecule diffusion generation}.
\newblock In \bibinfo{editor}{Krause, A.} \emph{et~al.} (eds.)
  \emph{\bibinfo{booktitle}{Proceedings of the 40th International Conference on
  Machine Learning}}, vol. \bibinfo{volume}{202} of
  \emph{\bibinfo{series}{Proceedings of Machine Learning Research}},
  \bibinfo{pages}{27611--27629} (\bibinfo{publisher}{PMLR},
  \bibinfo{year}{2023}).

\bibitem{zinc22}
\bibinfo{author}{Tingle, B.~I.} \emph{et~al.}
\newblock \bibinfo{title}{Zinc-22-a free multi-billion-scale database of
  tangible compounds for ligand discovery}.
\newblock \emph{\bibinfo{journal}{J. Chem. Inf. Model.}}
  \textbf{\bibinfo{volume}{63}}, \bibinfo{pages}{1166--1176}
  (\bibinfo{year}{2023}).

\bibitem{Ferreira2015}
\bibinfo{author}{Ferreira, L.}, \bibinfo{author}{dos Santos, R.},
  \bibinfo{author}{Oliva, G.} \& \bibinfo{author}{Andricopulo, A.}
\newblock \bibinfo{title}{Molecular docking and structure-based drug design
  strategies}.
\newblock \emph{\bibinfo{journal}{Molecules}} \textbf{\bibinfo{volume}{20}},
  \bibinfo{pages}{13384--13421} (\bibinfo{year}{2015}).

\bibitem{Tadesse2015}
\bibinfo{author}{Tadesse, S.}, \bibinfo{author}{Yu, M.},
  \bibinfo{author}{Kumarasiri, M.}, \bibinfo{author}{Le, B.~T.} \&
  \bibinfo{author}{Wang, S.}
\newblock \bibinfo{title}{Targeting cdk6 in cancer: State of the art and new
  insights}.
\newblock \emph{\bibinfo{journal}{Cell Cycle}} \textbf{\bibinfo{volume}{14}},
  \bibinfo{pages}{3220--3230} (\bibinfo{year}{2015}).

\bibitem{ElAmouri2008}
\bibinfo{author}{El-Amouri, S.~S.} \emph{et~al.}
\newblock \bibinfo{title}{Neprilysin: An enzyme candidate to slow the
  progression of alzheimer’s disease}.
\newblock \emph{\bibinfo{journal}{Am. J. Pathol.}}
  \textbf{\bibinfo{volume}{172}}, \bibinfo{pages}{1342--1354}
  (\bibinfo{year}{2008}).

\bibitem{Burley2022}
\bibinfo{author}{Burley, S.~K.} \emph{et~al.}
\newblock \bibinfo{title}{Rcsb protein data bank (rcsb.org): delivery of
  experimentally-determined pdb structures alongside one million computed
  structure models of proteins from artificial intelligence/machine learning}.
\newblock \emph{\bibinfo{journal}{Nucleic Acids Res.}}
  \textbf{\bibinfo{volume}{51}}, \bibinfo{pages}{D488--D508}
  (\bibinfo{year}{2022}).

\bibitem{Neves2012}
\bibinfo{author}{Neves, M. A.~C.}, \bibinfo{author}{Totrov, M.} \&
  \bibinfo{author}{Abagyan, R.}
\newblock \bibinfo{title}{Docking and scoring with icm: the benchmarking
  results and strategies for improvement}.
\newblock \emph{\bibinfo{journal}{J. Comput. Aided Mol. Des.}}
  \textbf{\bibinfo{volume}{26}}, \bibinfo{pages}{675--686}
  (\bibinfo{year}{2012}).

\bibitem{Yang2018admetsar}
\bibinfo{author}{Yang, H.} \emph{et~al.}
\newblock \bibinfo{title}{admetsar 2.0: web-service for prediction and
  optimization of chemical admet properties}.
\newblock \emph{\bibinfo{journal}{Bioinform.}} \textbf{\bibinfo{volume}{35}},
  \bibinfo{pages}{1067--1069} (\bibinfo{year}{2018}).

\bibitem{Halgren1996}
\bibinfo{author}{Halgren, T.~A.}
\newblock \bibinfo{title}{Merck molecular force field. i. basis, form, scope,
  parameterization, and performance of mmff94}.
\newblock \emph{\bibinfo{journal}{J. Comput. Chem.}}
  \textbf{\bibinfo{volume}{17}}, \bibinfo{pages}{490--519}
  (\bibinfo{year}{1996}).

\bibitem{Patnaik2016}
\bibinfo{author}{Patnaik, A.} \emph{et~al.}
\newblock \bibinfo{title}{Efficacy and safety of abemaciclib, an inhibitor of
  cdk4 and cdk6, for patients with breast cancer, non–small cell lung cancer,
  and other solid tumors}.
\newblock \emph{\bibinfo{journal}{Cancer Discov.}}
  \textbf{\bibinfo{volume}{6}}, \bibinfo{pages}{740--753}
  (\bibinfo{year}{2016}).

\bibitem{Lu2015}
\bibinfo{author}{Lu, J.}
\newblock \bibinfo{title}{Palbociclib: a first-in-class cdk4/cdk6 inhibitor for
  the treatment of hormone-receptor positive advanced breast cancer}.
\newblock \emph{\bibinfo{journal}{J. Hematol. Oncol.}}
  \textbf{\bibinfo{volume}{8}}, \bibinfo{pages}{98} (\bibinfo{year}{2015}).

\bibitem{Tripathy2017}
\bibinfo{author}{Tripathy, D.}, \bibinfo{author}{Bardia, A.} \&
  \bibinfo{author}{Sellers, W.~R.}
\newblock \bibinfo{title}{Ribociclib (lee011): Mechanism of action and clinical
  impact of this selective cyclin-dependent kinase 4/6 inhibitor in various
  solid tumors}.
\newblock \emph{\bibinfo{journal}{Clin. Cancer Res.}}
  \textbf{\bibinfo{volume}{23}}, \bibinfo{pages}{3251--3262}
  (\bibinfo{year}{2017}).

\bibitem{Benigni2010}
\bibinfo{author}{Benigni, R.}, \bibinfo{author}{Bossa, C.},
  \bibinfo{author}{Tcheremenskaia, O.} \& \bibinfo{author}{Giuliani, A.}
\newblock \bibinfo{title}{Alternatives to the carcinogenicity bioassay:in
  silicomethods, and thein vitroandin vivomutagenicity assays}.
\newblock \emph{\bibinfo{journal}{Expert Opin. Drug Metab. Toxicol.}}
  \textbf{\bibinfo{volume}{6}}, \bibinfo{pages}{809--819}
  (\bibinfo{year}{2010}).

\bibitem{Soo2018}
\bibinfo{author}{Soo, J. Y.-C.}, \bibinfo{author}{Jansen, J.},
  \bibinfo{author}{Masereeuw, R.} \& \bibinfo{author}{Little, M.~H.}
\newblock \bibinfo{title}{Advances in predictive in vitro models of
  drug-induced nephrotoxicity}.
\newblock \emph{\bibinfo{journal}{Nat. Rev. Nephrol.}}
  \textbf{\bibinfo{volume}{14}}, \bibinfo{pages}{378--393}
  (\bibinfo{year}{2018}).

\bibitem{Hansen2008}
\bibinfo{author}{Hansen, R.~A.} \emph{et~al.}
\newblock \bibinfo{title}{Efficacy and safety of donepezil, galantamine, and
  rivastigmine for the treatment of alzheimer’s disease: a systematic review
  and meta-analysis}.
\newblock \emph{\bibinfo{journal}{Clin. Interv. Aging}}
  \textbf{\bibinfo{volume}{3}}, \bibinfo{pages}{211--225}
  (\bibinfo{year}{2008}).

\bibitem{Deane2007}
\bibinfo{author}{Deane, R.} \& \bibinfo{author}{Zlokovic, B.}
\newblock \bibinfo{title}{Role of the blood-brain barrier in the pathogenesis
  of alzheimers disease}.
\newblock \emph{\bibinfo{journal}{Curr. Alzheimer Res.}}
  \textbf{\bibinfo{volume}{4}}, \bibinfo{pages}{191--197}
  (\bibinfo{year}{2007}).

\bibitem{Du2016}
\bibinfo{author}{Du, X.} \emph{et~al.}
\newblock \bibinfo{title}{Insights into protein–ligand interactions:
  Mechanisms, models, and methods}.
\newblock \emph{\bibinfo{journal}{Int. J. Mol. Sci.}}
  \textbf{\bibinfo{volume}{17}}, \bibinfo{pages}{144} (\bibinfo{year}{2016}).

\bibitem{Lin2003}
\bibinfo{author}{Lin, J.} \emph{et~al.}
\newblock \bibinfo{title}{The role of absorption, distribution, metabolism,
  excretion and toxicity in drug discovery}.
\newblock \emph{\bibinfo{journal}{Curr. Top. Med. Chem.}}
  \textbf{\bibinfo{volume}{3}}, \bibinfo{pages}{1125--1154}
  (\bibinfo{year}{2003}).

\bibitem{Chen2023}
\bibinfo{author}{Chen, Z.}, \bibinfo{author}{Ayinde, O.~R.},
  \bibinfo{author}{Fuchs, J.~R.}, \bibinfo{author}{Sun, H.} \&
  \bibinfo{author}{Ning, X.}
\newblock \bibinfo{title}{G2retro as a two-step graph generative models for
  retrosynthesis prediction}.
\newblock \emph{\bibinfo{journal}{Commun. Chem.}} \textbf{\bibinfo{volume}{6}},
  \bibinfo{pages}{102} (\bibinfo{year}{2023}).

\bibitem{Gao2020}
\bibinfo{author}{Gao, W.} \& \bibinfo{author}{Coley, C.~W.}
\newblock \bibinfo{title}{The synthesizability of molecules proposed by
  generative models}.
\newblock \emph{\bibinfo{journal}{J. Chem. Inf. Model.}}
  \textbf{\bibinfo{volume}{60}}, \bibinfo{pages}{5714--5723}
  (\bibinfo{year}{2020}).

\bibitem{Atz2021}
\bibinfo{author}{Atz, K.}, \bibinfo{author}{Grisoni, F.} \&
  \bibinfo{author}{Schneider, G.}
\newblock \bibinfo{title}{Geometric deep learning on molecular
  representations}.
\newblock \emph{\bibinfo{journal}{Nat. Mach. Intell.}}
  \textbf{\bibinfo{volume}{3}}, \bibinfo{pages}{1023--1032}
  (\bibinfo{year}{2021}).

\bibitem{park2019sdf}
\bibinfo{author}{Park, J.~J.}, \bibinfo{author}{Florence, P.},
  \bibinfo{author}{Straub, J.}, \bibinfo{author}{Newcombe, R.} \&
  \bibinfo{author}{Lovegrove, S.}
\newblock \bibinfo{title}{Deepsdf: Learning continuous signed distance
  functions for shape representation}.
\newblock In \emph{\bibinfo{booktitle}{Proceedings of the IEEE/CVF Conference
  on Computer Vision and Pattern Recognition (CVPR)}} (\bibinfo{year}{2019}).

\bibitem{deng2021vn}
\bibinfo{author}{Deng, C.} \emph{et~al.}
\newblock \bibinfo{title}{Vector neurons: A general framework for
  so(3)-equivariant networks}.
\newblock In \emph{\bibinfo{booktitle}{Proceedings of the IEEE/CVF
  International Conference on Computer Vision (ICCV)}},
  \bibinfo{pages}{12200--12209} (\bibinfo{year}{2021}).

\bibitem{wang2019dynamic}
\bibinfo{author}{Wang, Y.} \emph{et~al.}
\newblock \bibinfo{title}{Dynamic graph cnn for learning on point clouds}.
\newblock \emph{\bibinfo{journal}{ACM Trans. Graph.}}
  \textbf{\bibinfo{volume}{38}}, \bibinfo{pages}{1--12} (\bibinfo{year}{2019}).

\bibitem{ho2020ddpm}
\bibinfo{author}{Ho, J.}, \bibinfo{author}{Jain, A.} \&
  \bibinfo{author}{Abbeel, P.}
\newblock \bibinfo{title}{Denoising diffusion probabilistic models}.
\newblock In \bibinfo{editor}{Larochelle, H.}, \bibinfo{editor}{Ranzato, M.},
  \bibinfo{editor}{Hadsell, R.}, \bibinfo{editor}{Balcan, M.} \&
  \bibinfo{editor}{Lin, H.} (eds.) \emph{\bibinfo{booktitle}{Advances in Neural
  Information Processing Systems}}, vol.~\bibinfo{volume}{33},
  \bibinfo{pages}{6840--6851} (\bibinfo{publisher}{Curran Associates, Inc.},
  \bibinfo{year}{2020}).

\bibitem{hoogeboom2021catdiff}
\bibinfo{author}{Hoogeboom, E.}, \bibinfo{author}{Nielsen, D.},
  \bibinfo{author}{Jaini, P.}, \bibinfo{author}{Forr\'{e}, P.} \&
  \bibinfo{author}{Welling, M.}
\newblock \bibinfo{title}{Argmax flows and multinomial diffusion: Learning
  categorical distributions}.
\newblock In \bibinfo{editor}{Ranzato, M.}, \bibinfo{editor}{Beygelzimer, A.},
  \bibinfo{editor}{Dauphin, Y.}, \bibinfo{editor}{Liang, P.} \&
  \bibinfo{editor}{Vaughan, J.~W.} (eds.) \emph{\bibinfo{booktitle}{Advances in
  Neural Information Processing Systems}}, vol.~\bibinfo{volume}{34},
  \bibinfo{pages}{12454--12465} (\bibinfo{publisher}{Curran Associates, Inc.},
  \bibinfo{year}{2021}).

\bibitem{kullback1951information}
\bibinfo{author}{Kullback, S.} \& \bibinfo{author}{Leibler, R.~A.}
\newblock \bibinfo{title}{On information and sufficiency}.
\newblock \emph{\bibinfo{journal}{The Annals of Mathematical Statistics}}
  \textbf{\bibinfo{volume}{22}}, \bibinfo{pages}{79--86}
  (\bibinfo{year}{1951}).

\bibitem{Jumper2021}
\bibinfo{author}{Jumper, J.} \emph{et~al.}
\newblock \bibinfo{title}{Highly accurate protein structure prediction with
  alphafold}.
\newblock \emph{\bibinfo{journal}{Nat.}} \textbf{\bibinfo{volume}{596}},
  \bibinfo{pages}{583--589} (\bibinfo{year}{2021}).

\bibitem{jing2021learning}
\bibinfo{author}{Jing, B.}, \bibinfo{author}{Eismann, S.},
  \bibinfo{author}{Suriana, P.}, \bibinfo{author}{Townshend, R. J.~L.} \&
  \bibinfo{author}{Dror, R.}
\newblock \bibinfo{title}{Learning from protein structure with geometric vector
  perceptrons}.
\newblock In \emph{\bibinfo{booktitle}{11th International Conference on
  Learning Representations}} (\bibinfo{year}{2021}).

\bibitem{satorras2021}
\bibinfo{author}{Garcia~Satorras, V.}, \bibinfo{author}{Hoogeboom, E.},
  \bibinfo{author}{Fuchs, F.}, \bibinfo{author}{Posner, I.} \&
  \bibinfo{author}{Welling, M.}
\newblock \bibinfo{title}{E(n) equivariant normalizing flows}.
\newblock In \bibinfo{editor}{Ranzato, M.}, \bibinfo{editor}{Beygelzimer, A.},
  \bibinfo{editor}{Dauphin, Y.}, \bibinfo{editor}{Liang, P.} \&
  \bibinfo{editor}{Vaughan, J.~W.} (eds.) \emph{\bibinfo{booktitle}{Advances in
  Neural Information Processing Systems}}, vol.~\bibinfo{volume}{34},
  \bibinfo{pages}{4181--4192} (\bibinfo{publisher}{Curran Associates, Inc.},
  \bibinfo{year}{2021}).

\bibitem{torge2023diffhopp}
\bibinfo{author}{Torge, J.}, \bibinfo{author}{Harris, C.},
  \bibinfo{author}{Mathis, S.~V.} \& \bibinfo{author}{Lio, P.}
\newblock \bibinfo{title}{Diffhopp: A graph diffusion model for novel drug
  design via scaffold hopping}.
\newblock \emph{\bibinfo{journal}{arXiv:2308.07416}}  (\bibinfo{year}{2023}).

\bibitem{dhariwal2021diffusion}
\bibinfo{author}{Dhariwal, P.} \& \bibinfo{author}{Nichol, A.~Q.}
\newblock \bibinfo{title}{Diffusion models beat {GAN}s on image synthesis}.
\newblock In \bibinfo{editor}{Beygelzimer, A.}, \bibinfo{editor}{Dauphin, Y.},
  \bibinfo{editor}{Liang, P.} \& \bibinfo{editor}{Vaughan, J.~W.} (eds.)
  \emph{\bibinfo{booktitle}{Advances in Neural Information Processing Systems}}
  (\bibinfo{year}{2021}).

\bibitem{adam}
\bibinfo{author}{Kingma, D.~P.} \& \bibinfo{author}{Ba, J.}
\newblock \bibinfo{title}{Adam: {A} method for stochastic optimization}.
\newblock In \bibinfo{editor}{Bengio, Y.} \& \bibinfo{editor}{LeCun, Y.} (eds.)
  \emph{\bibinfo{booktitle}{3rd International Conference on Learning
  Representations, {ICLR} 2015, San Diego, CA, USA, 2015}}
  (\bibinfo{year}{2015}).

\bibitem{logs}
\bibinfo{author}{Portal, O.~C.}
\newblock \bibinfo{title}{logs calculation}.
\newblock \urlprefix\url{https://www.organic-chemistry.org/prog/peo/logS.html}.

\bibitem{Wjcikowski2015oddt}
\bibinfo{author}{W{\'{o}}jcikowski, M.}, \bibinfo{author}{Zielenkiewicz, P.} \&
  \bibinfo{author}{Siedlecki, P.}
\newblock \bibinfo{title}{Open drug discovery toolkit ({ODDT}): a new
  open-source player in the drug discovery field}.
\newblock \emph{\bibinfo{journal}{J. Cheminform.}} \textbf{\bibinfo{volume}{7}}
  (\bibinfo{year}{2015}).

\bibitem{ravi2020pytorch3d}
\bibinfo{author}{Ravi, N.} \emph{et~al.}
\newblock \bibinfo{title}{Accelerating 3d deep learning with pytorch3d}.
\newblock \emph{\bibinfo{journal}{arXiv:2007.08501}}  (\bibinfo{year}{2020}).

\bibitem{nichol2021}
\bibinfo{author}{Nichol, A.~Q.} \& \bibinfo{author}{Dhariwal, P.}
\newblock \bibinfo{title}{Improved denoising diffusion probabilistic models}.
\newblock In \bibinfo{editor}{Meila, M.} \& \bibinfo{editor}{Zhang, T.} (eds.)
  \emph{\bibinfo{booktitle}{Proceedings of the 38th International Conference on
  Machine Learning}}, vol. \bibinfo{volume}{139} of
  \emph{\bibinfo{series}{Proceedings of Machine Learning Research}},
  \bibinfo{pages}{8162--8171} (\bibinfo{publisher}{PMLR},
  \bibinfo{year}{2021}).

\bibitem{kong2021diffwave}
\bibinfo{author}{Kong, Z.}, \bibinfo{author}{Ping, W.}, \bibinfo{author}{Huang,
  J.}, \bibinfo{author}{Zhao, K.} \& \bibinfo{author}{Catanzaro, B.}
\newblock \bibinfo{title}{Diffwave: A versatile diffusion model for audio
  synthesis}.
\newblock In \emph{\bibinfo{booktitle}{International Conference on Learning
  Representations}} (\bibinfo{year}{2021}).

\end{thebibliography}


\begin{thebibliography}{10}
\expandafter\ifx\csname url\endcsname\relax
  \def\url#1{\texttt{#1}}\fi
\expandafter\ifx\csname urlprefix\endcsname\relax\def\urlprefix{URL }\fi
\providecommand{\bibinfo}[2]{#2}
\providecommand{\eprint}[2][]{\url{#2}}

\bibitem{adam}
\bibinfo{author}{Kingma, D.~P.} \& \bibinfo{author}{Ba, J.}
\newblock \bibinfo{title}{Adam: {A} method for stochastic optimization}.
\newblock In \bibinfo{editor}{Bengio, Y.} \& \bibinfo{editor}{LeCun, Y.} (eds.)
  \emph{\bibinfo{booktitle}{3rd International Conference on Learning
  Representations, {ICLR} 2015, San Diego, CA, USA, 2015}}
  (\bibinfo{year}{2015}).

\bibitem{adams2023equivariant}
\bibinfo{author}{Adams, K.} \& \bibinfo{author}{Coley, C.~W.}
\newblock \bibinfo{title}{Equivariant shape-conditioned generation of 3d
  molecules for ligand-based drug design}.
\newblock In \emph{\bibinfo{booktitle}{11th International Conference on
  Learning Representations}} (\bibinfo{year}{2023}).

\bibitem{Eberhardt2021}
\bibinfo{author}{Eberhardt, J.}, \bibinfo{author}{Santos-Martins, D.},
  \bibinfo{author}{Tillack, A.~F.} \& \bibinfo{author}{Forli, S.}
\newblock \bibinfo{title}{Autodock vina 1.2.0: New docking methods, expanded
  force field, and python bindings}.
\newblock \emph{\bibinfo{journal}{J. Chem. Inf. Model.}}
  \textbf{\bibinfo{volume}{61}}, \bibinfo{pages}{3891--3898}
  (\bibinfo{year}{2021}).

\bibitem{Lipinski1997}
\bibinfo{author}{Lipinski, C.~A.}, \bibinfo{author}{Lombardo, F.},
  \bibinfo{author}{Dominy, B.~W.} \& \bibinfo{author}{Feeney, P.~J.}
\newblock \bibinfo{title}{Experimental and computational approaches to estimate
  solubility and permeability in drug discovery and development settings}.
\newblock \emph{\bibinfo{journal}{Adv. Drug Deliv. Rev.}}
  \textbf{\bibinfo{volume}{23}}, \bibinfo{pages}{3--25} (\bibinfo{year}{1997}).

\bibitem{Neves2012}
\bibinfo{author}{Neves, M. A.~C.}, \bibinfo{author}{Totrov, M.} \&
  \bibinfo{author}{Abagyan, R.}
\newblock \bibinfo{title}{Docking and scoring with icm: the benchmarking
  results and strategies for improvement}.
\newblock \emph{\bibinfo{journal}{J. Comput. Aided Mol. Des.}}
  \textbf{\bibinfo{volume}{26}}, \bibinfo{pages}{675--686}
  (\bibinfo{year}{2012}).

\bibitem{Patnaik2016}
\bibinfo{author}{Patnaik, A.} \emph{et~al.}
\newblock \bibinfo{title}{Efficacy and safety of abemaciclib, an inhibitor of
  cdk4 and cdk6, for patients with breast cancer, non–small cell lung cancer,
  and other solid tumors}.
\newblock \emph{\bibinfo{journal}{Cancer Discov.}}
  \textbf{\bibinfo{volume}{6}}, \bibinfo{pages}{740--753}
  (\bibinfo{year}{2016}).

\bibitem{Lu2015}
\bibinfo{author}{Lu, J.}
\newblock \bibinfo{title}{Palbociclib: a first-in-class cdk4/cdk6 inhibitor for
  the treatment of hormone-receptor positive advanced breast cancer}.
\newblock \emph{\bibinfo{journal}{J. Hematol. Oncol.}}
  \textbf{\bibinfo{volume}{8}}, \bibinfo{pages}{98} (\bibinfo{year}{2015}).

\bibitem{Tripathy2017}
\bibinfo{author}{Tripathy, D.}, \bibinfo{author}{Bardia, A.} \&
  \bibinfo{author}{Sellers, W.~R.}
\newblock \bibinfo{title}{Ribociclib (lee011): Mechanism of action and clinical
  impact of this selective cyclin-dependent kinase 4/6 inhibitor in various
  solid tumors}.
\newblock \emph{\bibinfo{journal}{Clin. Cancer Res.}}
  \textbf{\bibinfo{volume}{23}}, \bibinfo{pages}{3251--3262}
  (\bibinfo{year}{2017}).

\bibitem{logs}
\bibinfo{author}{Portal, O.~C.}
\newblock \bibinfo{title}{logs calculation}.
\newblock \urlprefix\url{https://www.organic-chemistry.org/prog/peo/logS.html}.

\bibitem{Hansen2008}
\bibinfo{author}{Hansen, R.~A.} \emph{et~al.}
\newblock \bibinfo{title}{Efficacy and safety of donepezil, galantamine, and
  rivastigmine for the treatment of alzheimer’s disease: a systematic review
  and meta-analysis}.
\newblock \emph{\bibinfo{journal}{Clin. Interv. Aging}}
  \textbf{\bibinfo{volume}{3}}, \bibinfo{pages}{211--225}
  (\bibinfo{year}{2008}).

\bibitem{Atz2021}
\bibinfo{author}{Atz, K.}, \bibinfo{author}{Grisoni, F.} \&
  \bibinfo{author}{Schneider, G.}
\newblock \bibinfo{title}{Geometric deep learning on molecular
  representations}.
\newblock \emph{\bibinfo{journal}{Nat. Mach. Intell.}}
  \textbf{\bibinfo{volume}{3}}, \bibinfo{pages}{1023--1032}
  (\bibinfo{year}{2021}).

\bibitem{Wjcikowski2015oddt}
\bibinfo{author}{W{\'{o}}jcikowski, M.}, \bibinfo{author}{Zielenkiewicz, P.} \&
  \bibinfo{author}{Siedlecki, P.}
\newblock \bibinfo{title}{Open drug discovery toolkit ({ODDT}): a new
  open-source player in the drug discovery field}.
\newblock \emph{\bibinfo{journal}{J. Cheminform.}} \textbf{\bibinfo{volume}{7}}
  (\bibinfo{year}{2015}).

\bibitem{ravi2020pytorch3d}
\bibinfo{author}{Ravi, N.} \emph{et~al.}
\newblock \bibinfo{title}{Accelerating 3d deep learning with pytorch3d}.
\newblock \emph{\bibinfo{journal}{arXiv:2007.08501}}  (\bibinfo{year}{2020}).

\bibitem{hoogeboom2021catdiff}
\bibinfo{author}{Hoogeboom, E.}, \bibinfo{author}{Nielsen, D.},
  \bibinfo{author}{Jaini, P.}, \bibinfo{author}{Forr\'{e}, P.} \&
  \bibinfo{author}{Welling, M.}
\newblock \bibinfo{title}{Argmax flows and multinomial diffusion: Learning
  categorical distributions}.
\newblock In \bibinfo{editor}{Ranzato, M.}, \bibinfo{editor}{Beygelzimer, A.},
  \bibinfo{editor}{Dauphin, Y.}, \bibinfo{editor}{Liang, P.} \&
  \bibinfo{editor}{Vaughan, J.~W.} (eds.) \emph{\bibinfo{booktitle}{Advances in
  Neural Information Processing Systems}}, vol.~\bibinfo{volume}{34},
  \bibinfo{pages}{12454--12465} (\bibinfo{publisher}{Curran Associates, Inc.},
  \bibinfo{year}{2021}).

\bibitem{ho2020ddpm}
\bibinfo{author}{Ho, J.}, \bibinfo{author}{Jain, A.} \&
  \bibinfo{author}{Abbeel, P.}
\newblock \bibinfo{title}{Denoising diffusion probabilistic models}.
\newblock In \bibinfo{editor}{Larochelle, H.}, \bibinfo{editor}{Ranzato, M.},
  \bibinfo{editor}{Hadsell, R.}, \bibinfo{editor}{Balcan, M.} \&
  \bibinfo{editor}{Lin, H.} (eds.) \emph{\bibinfo{booktitle}{Advances in Neural
  Information Processing Systems}}, vol.~\bibinfo{volume}{33},
  \bibinfo{pages}{6840--6851} (\bibinfo{publisher}{Curran Associates, Inc.},
  \bibinfo{year}{2020}).

\bibitem{guan2023targetdiff}
\bibinfo{author}{Guan, J.} \emph{et~al.}
\newblock \bibinfo{title}{3d equivariant diffusion for target-aware molecule
  generation and affinity prediction}.
\newblock In \emph{\bibinfo{booktitle}{11th International Conference on
  Learning Representations}} (\bibinfo{year}{2023}).

\bibitem{nichol2021}
\bibinfo{author}{Nichol, A.~Q.} \& \bibinfo{author}{Dhariwal, P.}
\newblock \bibinfo{title}{Improved denoising diffusion probabilistic models}.
\newblock In \bibinfo{editor}{Meila, M.} \& \bibinfo{editor}{Zhang, T.} (eds.)
  \emph{\bibinfo{booktitle}{Proceedings of the 38th International Conference on
  Machine Learning}}, vol. \bibinfo{volume}{139} of
  \emph{\bibinfo{series}{Proceedings of Machine Learning Research}},
  \bibinfo{pages}{8162--8171} (\bibinfo{publisher}{PMLR},
  \bibinfo{year}{2021}).

\bibitem{kong2021diffwave}
\bibinfo{author}{Kong, Z.}, \bibinfo{author}{Ping, W.}, \bibinfo{author}{Huang,
  J.}, \bibinfo{author}{Zhao, K.} \& \bibinfo{author}{Catanzaro, B.}
\newblock \bibinfo{title}{Diffwave: A versatile diffusion model for audio
  synthesis}.
\newblock In \emph{\bibinfo{booktitle}{International Conference on Learning
  Representations}} (\bibinfo{year}{2021}).

\end{thebibliography}
